\documentclass[10pt]{mystyle}
\usepackage{float}
\usepackage{fontawesome5}
\usepackage{wrapfig}
\usepackage{multirow}
\usepackage{array}
\usepackage{tikz}
\usetikzlibrary{arrows.meta,positioning,fit,calc,shapes.geometric}
\usepackage{booktabs,longtable,pdflscape,pifont}
\usepackage{makecell}
\usepackage{pgfplots}
\pgfplotsset{compat=1.18}
\usepackage[round,authoryear]{natbib}
\usepackage[nameinlink,noabbrev]{cleveref}
\usepackage{forest}
\usepackage{xcolor}
\usepackage{hyperref}
\usepackage{adjustbox}
\newcommand{\BenchmarkCount}{82}
\newcommand{\ArchitectureCount}{297}

\newcommand{\TtwoCount}{204}
\newcommand{\TthreeCount}{17}
\newcommand{\AdaptiveCount}{221}
\newcommand{\AdaptivePercent}{74.4}
\newcommand{\TtwoPercent}{68.7}
\newcommand{\TthreePercent}{5.7}
\newcommand{\HybridPercent}{46.8}

\newcommand{\CellCrossTThree}{13}

\newcommand{\CellHybridTThree}{4}
\newcommand{\CellHybridTotal}{139}

\definecolor{tierone}{HTML}{D98C1D}
\definecolor{tiertwo}{HTML}{158E95}
\definecolor{tierthree}{HTML}{C82E66}
\definecolor{nativeblue}{HTML}{2E6FBB}
\definecolor{crosspurple}{HTML}{7651A8}
\definecolor{softgray}{HTML}{F3F5F7}

\tcbuselibrary{skins,breakable}  
\definecolor{mmagent_dark}{HTML}{E8D5BE} 
\definecolor{mmagent_light}{HTML}{E8D5BE} 
\definecolor{mmagent_lighter}{HTML}{FFF7E8}
\newtcolorbox{keybox}[2][]{
  colback=mmagent_light!20,
  colframe=mmagent_light!80,
  colbacktitle=mmagent_light!80,
  coltitle=black,
  title={\bfseries #2},
  boxrule=2pt,
  arc=5pt,
  drop shadow,
  parbox=false,
  breakable,
  before skip=8pt,
  after skip=8pt,
  left=5pt,
  right=5pt,
  #1
}

\title{Seeing, Maintaining, and Learning:\\The Evolution of Multimodal Agent Memory}
\runningtitle{Seeing, Maintaining, and Learning: The Evolution of Multimodal Agent Memory}
\newcommand{\authorrow}[1]{\noindent\makebox[\linewidth][c]{#1}\par}
\author{%
  \parbox{\linewidth}{\centering
    \normalfont\fontsize{9.3}{14}\selectfont
    \authorrow{%
      \mbox{\textbf{Minghao Guo}\textsuperscript{1,*,\textsection}}\hspace{8pt}%
      \mbox{\textbf{Qingyue Jiao}\textsuperscript{2,*,\textsection}}\hspace{8pt}%
      \mbox{\textbf{Xiujin Liu}\textsuperscript{3,*}}\hspace{8pt}%
      \mbox{\textbf{Jianhong Pang}\textsuperscript{4,*}}\hspace{8pt}%
      \mbox{\textbf{Ruimeng Yang}\textsuperscript{4,*}}\hspace{8pt}%
      \mbox{\textbf{Naihao Xu}\textsuperscript{2,\textdaggerdbl}}}%
    \authorrow{%
      \mbox{\textbf{Liwei Che}\textsuperscript{1,\textdaggerdbl}}\hspace{8pt}%
      \mbox{\textbf{Haochen Xue}\textsuperscript{5,\textdaggerdbl}}\hspace{8pt}%
      \mbox{\textbf{Ziqi Wang}\textsuperscript{1,\textdaggerdbl}}\hspace{8pt}%
      \mbox{\textbf{Xinying Cai}\textsuperscript{1,\textdaggerdbl}}\hspace{8pt}%
      \mbox{\textbf{Jiaxuan Luo}\textsuperscript{6,\textdaggerdbl}}\hspace{8pt}%
      \mbox{\textbf{Jiaxin Deng}\textsuperscript{7,\textdaggerdbl}}\hspace{8pt}%
      \mbox{\textbf{Xuesong Ye}\textsuperscript{9,\textdaggerdbl}}}%
    \authorrow{%
      \mbox{\textbf{Jintao Qu}\textsuperscript{11,\textdaggerdbl}}\hspace{8pt}%
      \mbox{\textbf{Wenyao Gao}\textsuperscript{10,\textdaggerdbl}}\hspace{8pt}%
      \mbox{\textbf{Jingwen Gu}\textsuperscript{8,\textdaggerdbl}}\hspace{8pt}%
      \mbox{\textbf{Chenyang Zhao}\textsuperscript{12}}\hspace{8pt}%
      \mbox{\textbf{Shichun Liu}\textsuperscript{4}}\hspace{8pt}%
      \mbox{\textbf{Guibin Zhang}\textsuperscript{13}}}%
    \authorrow{%
      \mbox{\textbf{Zefang Liu}\textsuperscript{14}}\hspace{8pt}%
      \mbox{\textbf{Siyin Wang}\textsuperscript{4}}\hspace{8pt}%
      \mbox{\textbf{Wujiang Xu}\textsuperscript{15}}\hspace{8pt}%
      \mbox{\textbf{Danrui Li}\textsuperscript{1}}\hspace{8pt}%
      \mbox{\textbf{Zeru Shi}\textsuperscript{1}}\hspace{8pt}%
      \mbox{\textbf{Woongyeong Yeo}\textsuperscript{18}}\hspace{8pt}%
      \mbox{\textbf{Shuang Liu}\textsuperscript{16}}}%
    \authorrow{%
      \mbox{\textbf{Wei-Chieh Huang}\textsuperscript{17}}\hspace{8pt}%
      \mbox{\textbf{Weizhi Zhang}\textsuperscript{17}}\hspace{8pt}%
      \mbox{\textbf{Tianyu Yang}\textsuperscript{2}}\hspace{8pt}%
      \mbox{\textbf{Kangsan Kim}\textsuperscript{18}}\hspace{8pt}%
      \mbox{\textbf{Yebowen Hu}\textsuperscript{19}}\hspace{8pt}%
      \mbox{\textbf{Bowen Wen}\textsuperscript{20}}}%
    \authorrow{%
      \mbox{\textbf{Bowen Jiang}\textsuperscript{21}}\hspace{8pt}%
      \mbox{\textbf{Haohao Su}\textsuperscript{22}}\hspace{8pt}%
      \mbox{\textbf{Ming Yin}\textsuperscript{23}}\hspace{8pt}%
      \mbox{\textbf{Sheldon Yu}\textsuperscript{8}}\hspace{8pt}%
      \mbox{\textbf{Mian Zhang}\textsuperscript{24}}\hspace{8pt}%
      \mbox{\textbf{Yuyan Chen}\textsuperscript{25,26}}\hspace{8pt}%
      \mbox{\textbf{Hanrong Zhang}\textsuperscript{17}}}%
    \authorrow{%
      \mbox{\textbf{Kai Mei}\textsuperscript{15}}\hspace{8pt}%
      \mbox{\textbf{Zihao Zhang}\textsuperscript{15}}\hspace{8pt}%
      \mbox{\textbf{Mingyu Jin}\textsuperscript{1}}\hspace{8pt}%
      \mbox{\textbf{Xi Zhu}\textsuperscript{1}}\hspace{8pt}%
      \mbox{\textbf{Hengyi Wang}\textsuperscript{1}}\hspace{8pt}%
      \mbox{\textbf{Tianxin Wei}\textsuperscript{34}}\hspace{8pt}%
      \mbox{\textbf{Linyong Nan}\textsuperscript{19}}}%
    \authorrow{%
      \mbox{\textbf{Zhenwen Liang}\textsuperscript{2}}\hspace{8pt}%
      \mbox{\textbf{Zhendong Chu}\textsuperscript{29}}\hspace{8pt}%
      \mbox{\textbf{Ziqi Huang}\textsuperscript{30}}\hspace{8pt}%
      \mbox{\textbf{Yongye Su}\textsuperscript{31}}\hspace{8pt}%
      \mbox{\textbf{Sixun Dong}\textsuperscript{28}}\hspace{8pt}%
      \mbox{\textbf{Shujian Liu}\textsuperscript{19}}\hspace{8pt}%
      \mbox{\textbf{Kaiqu Liang}\textsuperscript{27}}}%
    \authorrow{%
      \mbox{\textbf{Jiaojiao Han}\textsuperscript{15}}\hspace{8pt}%
      \mbox{\textbf{Zhaoye Fei}\textsuperscript{4}}\hspace{8pt}%
      \mbox{\textbf{Kun Wang}\textsuperscript{30}}\hspace{8pt}%
      \mbox{\textbf{Shihan Dou}\textsuperscript{4}}\hspace{8pt}%
      \mbox{\textbf{Zhichun Guo}\textsuperscript{32}}\hspace{8pt}%
      \mbox{\textbf{Xingjian Diao}\textsuperscript{33}}\hspace{8pt}%
      \mbox{\textbf{Zeqi Gu}\textsuperscript{37}}}%
    \authorrow{%
      \mbox{\textbf{Heng Wang}\textsuperscript{34}}\hspace{8pt}%
      \mbox{\textbf{Tao Feng}\textsuperscript{34}}\hspace{8pt}%
      \mbox{\textbf{Cheng Qian}\textsuperscript{34}}\hspace{8pt}%
      \mbox{\textbf{Hamid Palangi}\textsuperscript{35}}\hspace{8pt}%
      \mbox{\textbf{Silei Xu}\textsuperscript{19}}\hspace{8pt}%
      \mbox{\textbf{Xi Yun}\textsuperscript{19}}\hspace{8pt}%
      \mbox{\textbf{Yi Zhu}\textsuperscript{36}}\hspace{8pt}%
      \mbox{\textbf{Shilong Liu}\textsuperscript{27}}}%
    \authorrow{%
      \mbox{\textbf{Qingsong Wen}\textsuperscript{29}}\hspace{8pt}%
      \mbox{\textbf{Sung Ju Hwang}\textsuperscript{18}}\hspace{8pt}%
      \mbox{\textbf{Zuxuan Wu}\textsuperscript{4}}\hspace{8pt}%
      \mbox{\textbf{Yu-Gang Jiang}\textsuperscript{4}}\hspace{8pt}%
      \mbox{\textbf{Yiyu Shi}\textsuperscript{2}}\hspace{8pt}%
      \mbox{\textbf{Dimitris N. Metaxas}\textsuperscript{1}}}%
    \authorrow{%
      \mbox{\textbf{Jiaxuan You}\textsuperscript{34}}\hspace{8pt}%
      \mbox{\textbf{Julian McAuley}\textsuperscript{8}}\hspace{8pt}%
      \mbox{\textbf{Mengdi Wang}\textsuperscript{27}}\hspace{8pt}%
      \mbox{\textbf{Jiawei Han}\textsuperscript{34}}\hspace{8pt}%
      \mbox{\textbf{Philip S. Yu}\textsuperscript{17}}\hspace{8pt}%
      \mbox{\textbf{Ziyi Ye}\textsuperscript{4,\textdagger,\textsection}}}%
  }%
}

\renewcommand{\authorcontributions}{%
  \textsuperscript{*} Equal first authors;\quad
  \textsuperscript{\textdaggerdbl} Core authors;\quad
  \textsuperscript{\textdagger} Corresponding author;\quad
  \textsuperscript{\textsection} Project organizers.}

\newcommand{\resourcegithuburl}{https://github.com/Seeing-Maintaining-Learning/multimodal-agent-memory-evolution}
\newcommand{\resourcegithublabel}{Seeing-Maintaining-Learning/multimodal-agent-memory-evolution}
\renewcommand{\paperresources}{%
  \begingroup
  \par\vspace{8pt}%
  \normalfont\fontsize{9}{12}\selectfont
  \noindent\makebox[\linewidth][c]{%
  \begin{tabular}{@{}c@{}}
    \faEnvelope%
    \hspace{5pt}%
    \href{mailto:minghao.guo@rutgers.edu}{minghao.guo@rutgers.edu},
    \href{mailto:qjiao@nd.edu}{qjiao@nd.edu},
    \href{mailto:zyye@fudan.edu.cn}{zyye@fudan.edu.cn}%
  \if\relax\expandafter\detokenize\expandafter{\resourcegithuburl}\relax\else
    \\[4pt]
      \href{\resourcegithuburl}{%
        \raisebox{-1.4pt}{\includegraphics[height=10pt]{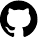}}%
        \hspace{5pt}\textcolor{TinaCrimson}{\resourcegithublabel}}%
  \fi
  \end{tabular}}\par
  \endgroup
}

\newcommand{\institutionmark}[4][12pt]{%
  \mbox{\textsuperscript{#2}\hspace{1pt}%
    \raisebox{\dimexpr 0.5ex-0.5\height\relax}{%
      \includegraphics[width=#1,height=12pt,keepaspectratio]{content/figures/logos/#3.pdf}}%
    \hspace{2.5pt}#4}%
}
\newcommand{\institutiontext}[2]{\mbox{\textsuperscript{#1}\hspace{1pt}#2}}
\newsavebox{\institutionrowbox}
\newcommand{\institutionrow}[1]{%
  \sbox{\institutionrowbox}{#1}%
  \ifdim\wd\institutionrowbox>\linewidth
    \PackageWarning{institutionlayout}{Institution row exceeds text width}%
  \fi
  \noindent\makebox[\linewidth][c]{\usebox{\institutionrowbox}}\par
}
\renewcommand{\institutiondisplay}{%
  \begingroup\centering\normalfont\fontsize{8.2}{17.5}\selectfont
  \setlength{\parskip}{0pt}%
  \institutionrow{%
    \institutionmark{1}{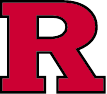}{Rutgers}\hspace{4pt}%
    \institutionmark{2}{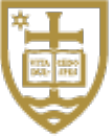}{Notre Dame}\hspace{4pt}%
    \institutionmark{3}{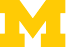}{Michigan}\hspace{4pt}%
    \institutionmark{4}{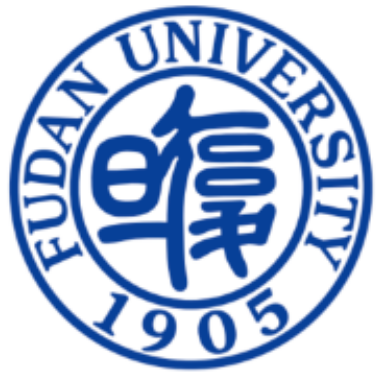}{Fudan}\hspace{4pt}%
    \institutionmark{5}{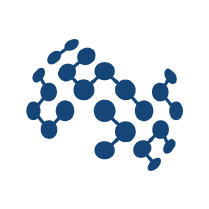}{MBZUAI}\hspace{4pt}%
    \institutionmark{6}{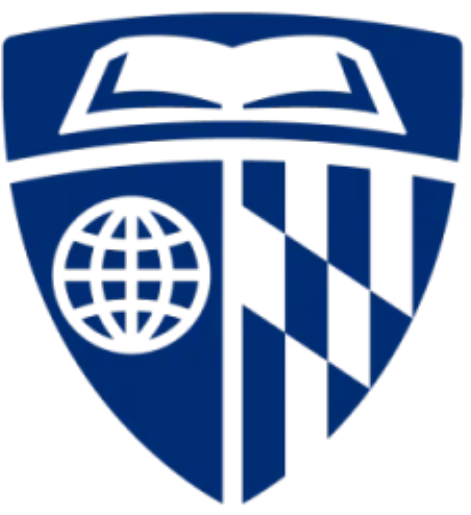}{JHU}\hspace{4pt}%
    \institutionmark{7}{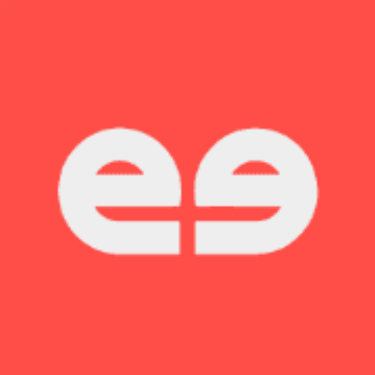}{Peer AI}\hspace{4pt}%
    \institutionmark{8}{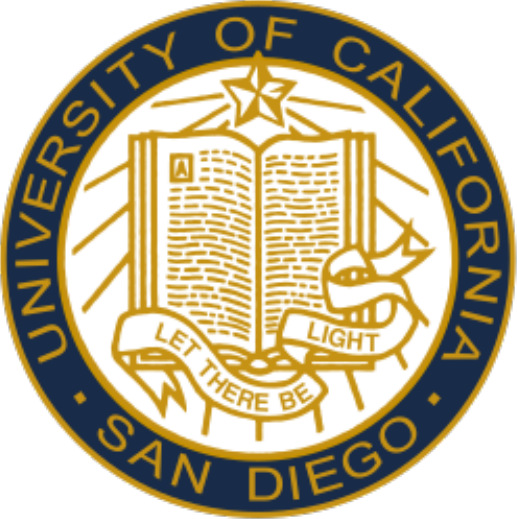}{UC San Diego}}%
  \institutionrow{%
    \institutionmark[17pt]{9}{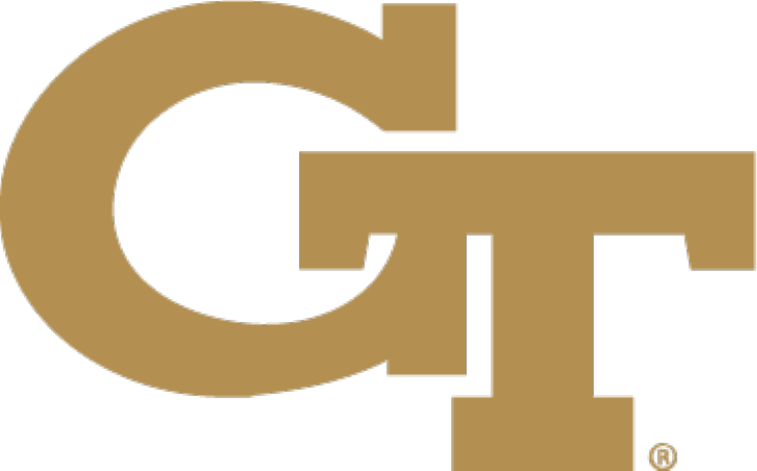}{Georgia Tech}\hspace{4pt}%
    \institutionmark{10}{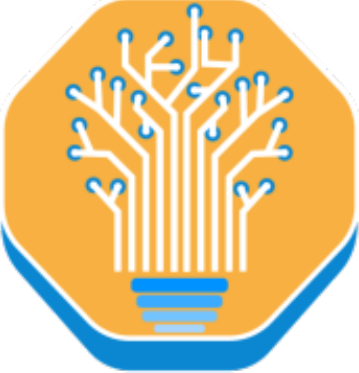}{SemiAnalysis}\hspace{4pt}%
    \institutionmark{11}{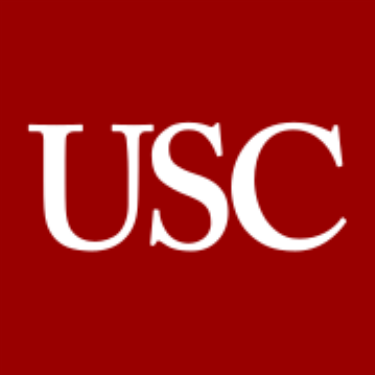}{USC}\hspace{4pt}%
    \institutionmark[21pt]{12}{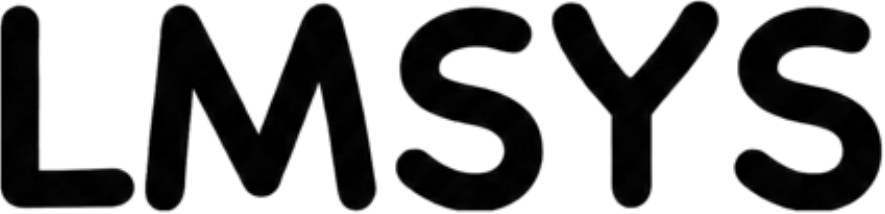}{LMSYS Org}\hspace{4pt}%
    \institutionmark{13}{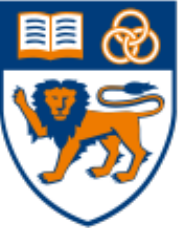}{NUS}\hspace{4pt}%
    \institutionmark{14}{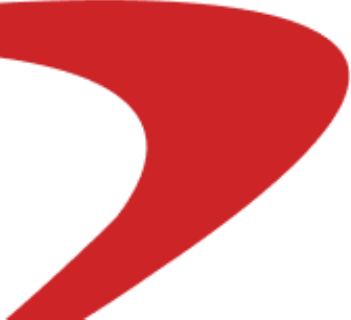}{Capital One}\hspace{4pt}%
    \institutiontext{15}{Independent Researcher}}%
  \institutionrow{%
    \institutionmark{16}{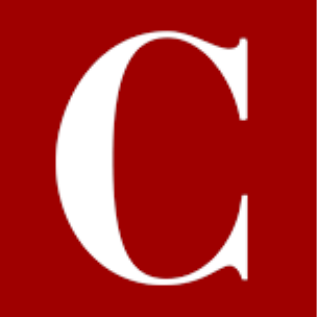}{CMU}\hspace{4pt}%
    \institutionmark{17}{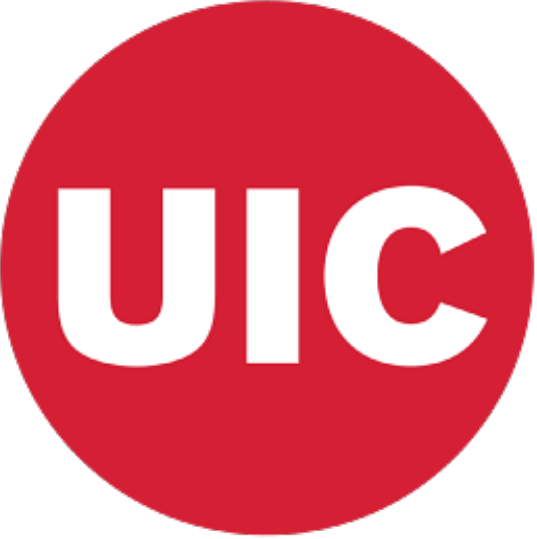}{UIC}\hspace{4pt}%
    \institutionmark{18}{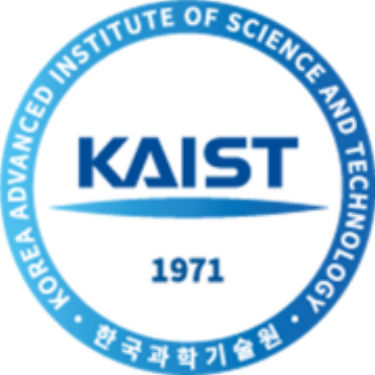}{KAIST}\hspace{4pt}%
    \institutionmark{19}{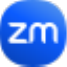}{Zoom}\hspace{4pt}%
    \institutionmark[16pt]{20}{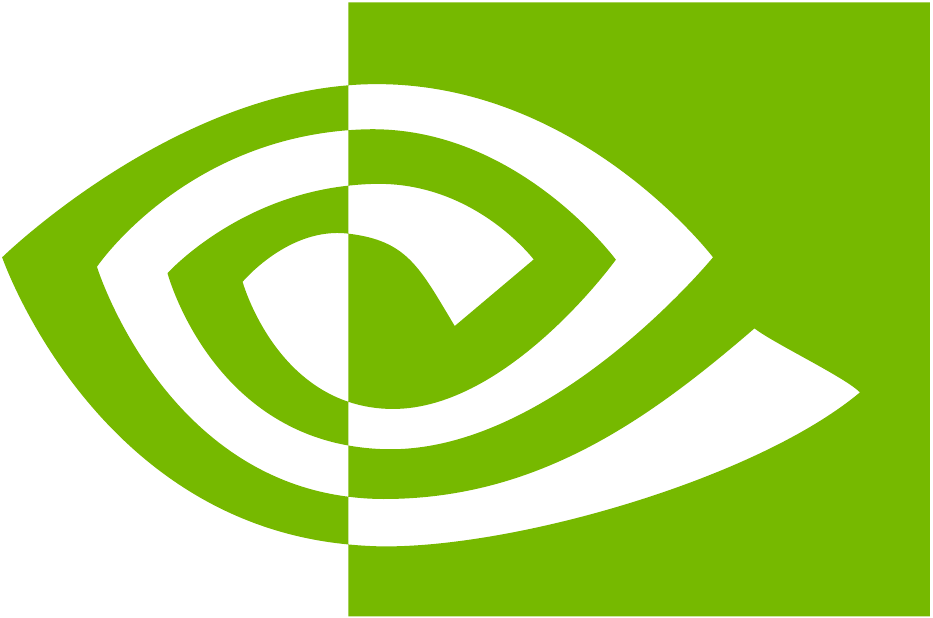}{NVIDIA}\hspace{4pt}%
    \institutionmark{21}{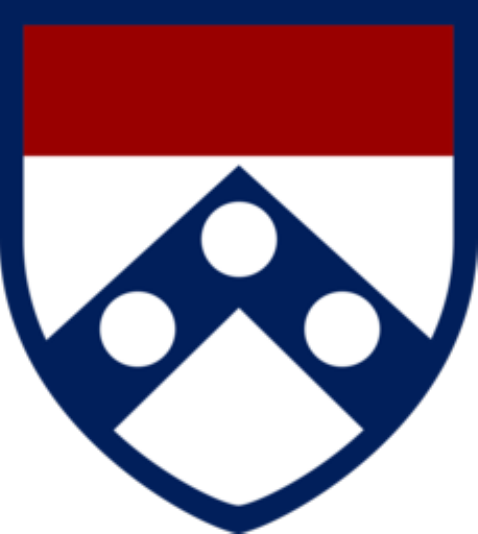}{Penn}\hspace{4pt}%
    \institutionmark{22}{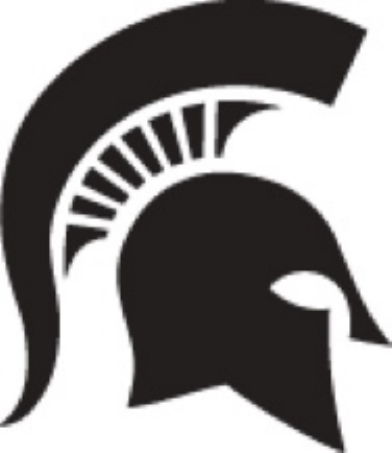}{MSU}\hspace{4pt}%
    \institutionmark{23}{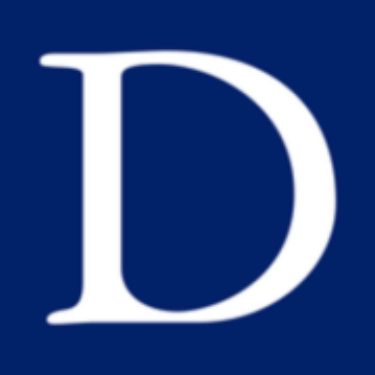}{Duke}\hspace{4pt}%
    \institutionmark{24}{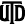}{UT Dallas}}%
  \institutionrow{%
    \institutionmark{25}{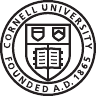}{Cornell}\hspace{4pt}%
    \institutionmark{26}{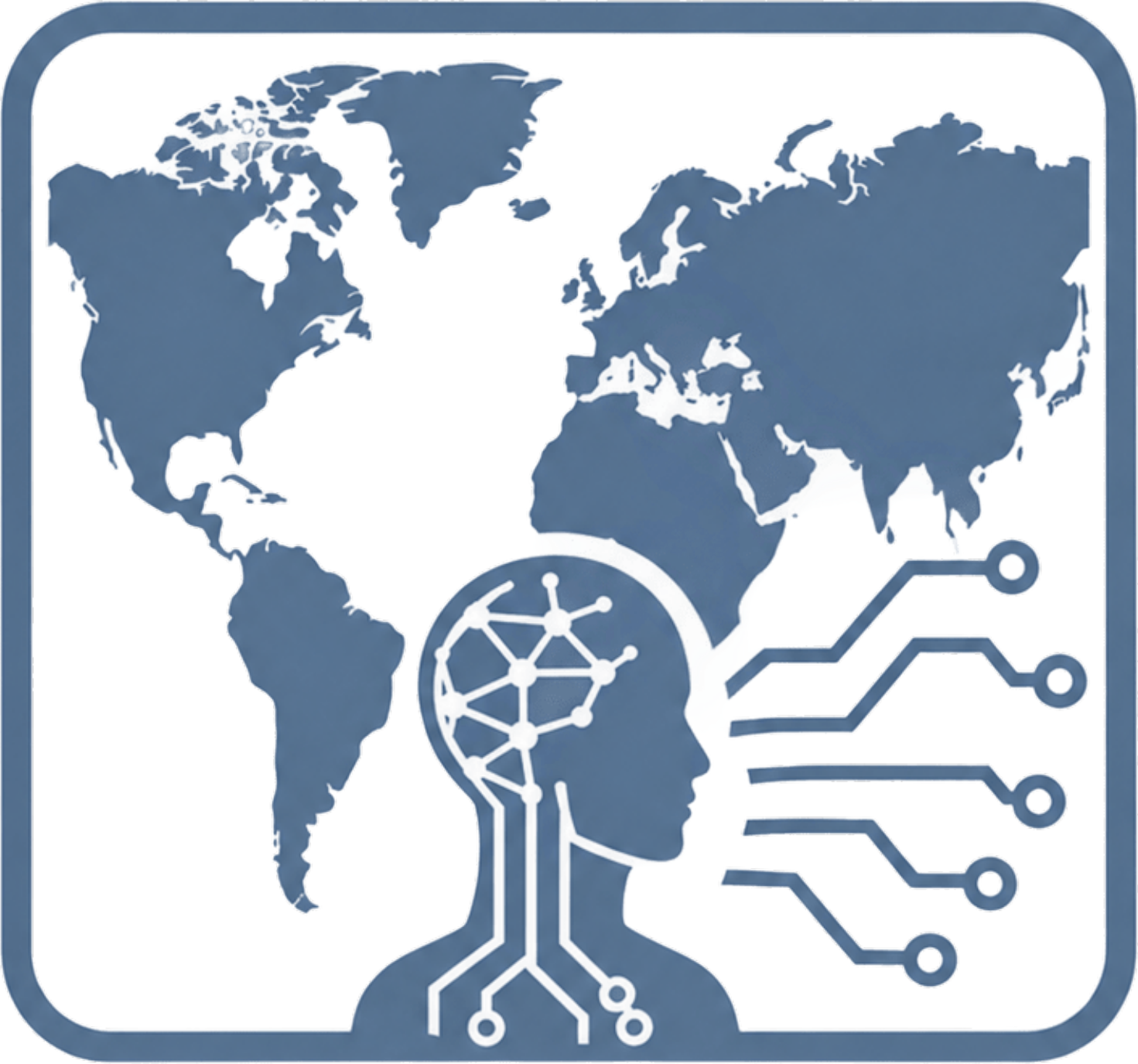}{ModelsLive}\hspace{4pt}%
    \institutionmark{27}{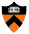}{Princeton}\hspace{4pt}%
    \institutionmark[16pt]{28}{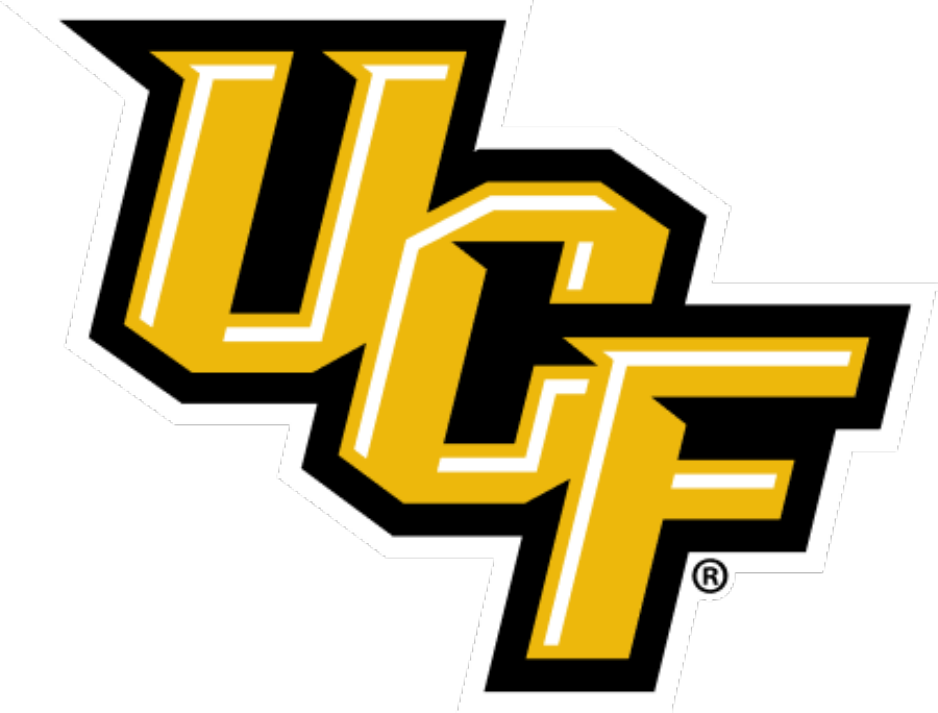}{UCF}\hspace{4pt}%
    \institutionmark{29}{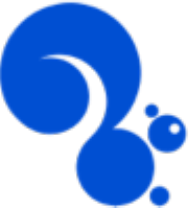}{Squirrel AI}\hspace{4pt}%
    \institutionmark{30}{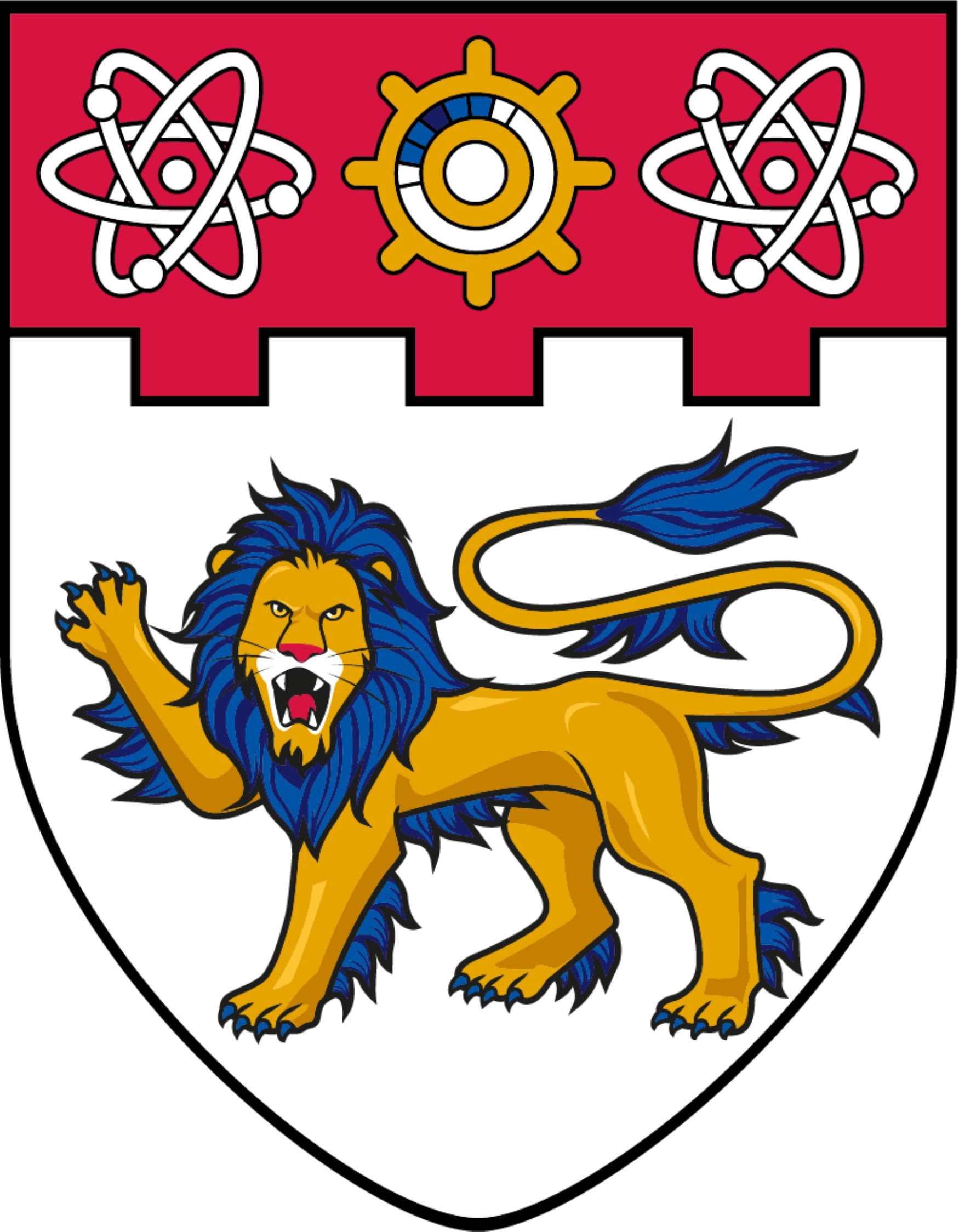}{NTU}\hspace{4pt}%
    \institutionmark{31}{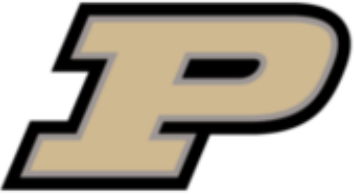}{Purdue}}%
  \institutionrow{%
    \institutionmark{32}{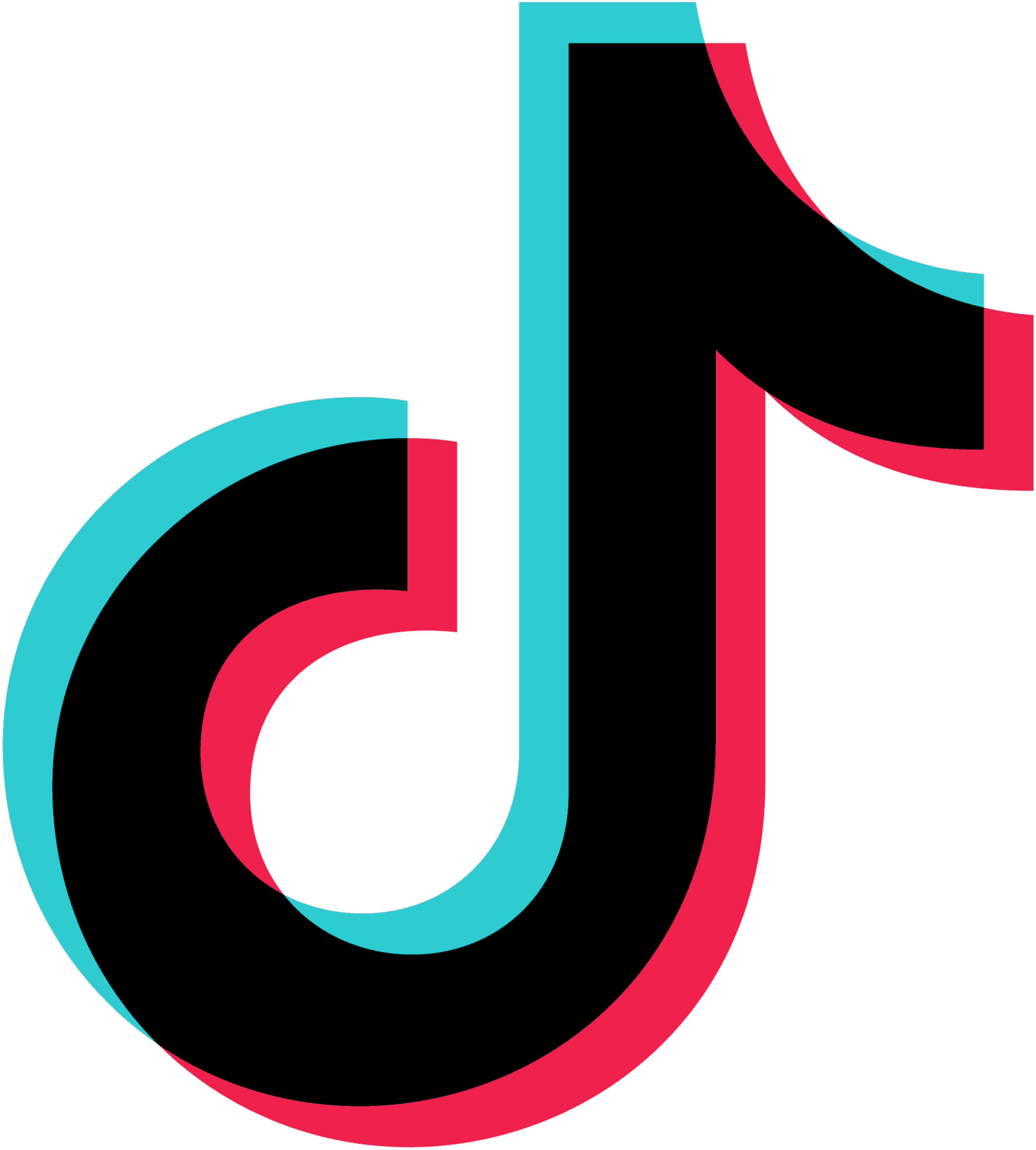}{TikTok}\hspace{4pt}%
    \institutionmark{33}{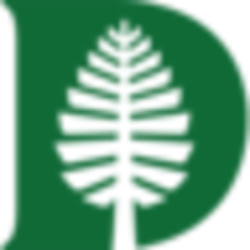}{Dartmouth}\hspace{4pt}%
    \institutionmark{34}{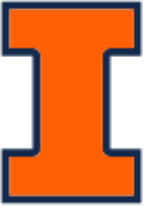}{UIUC}\hspace{4pt}%
    \institutionmark{35}{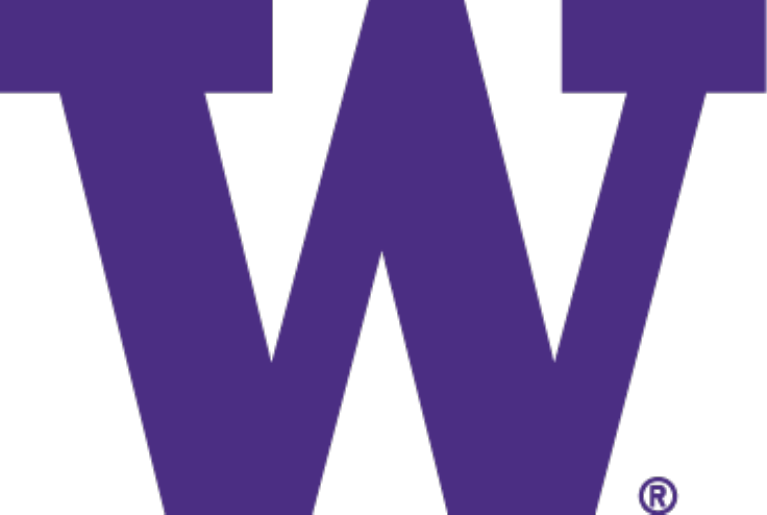}{Washington}\hspace{4pt}%
    \institutionmark{36}{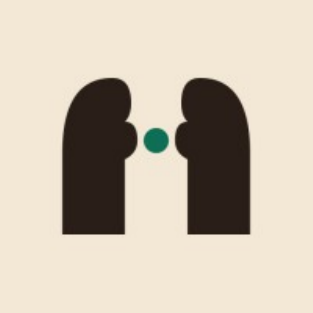}{Humani Labs}\hspace{4pt}%
    \institutionmark[17pt]{37}{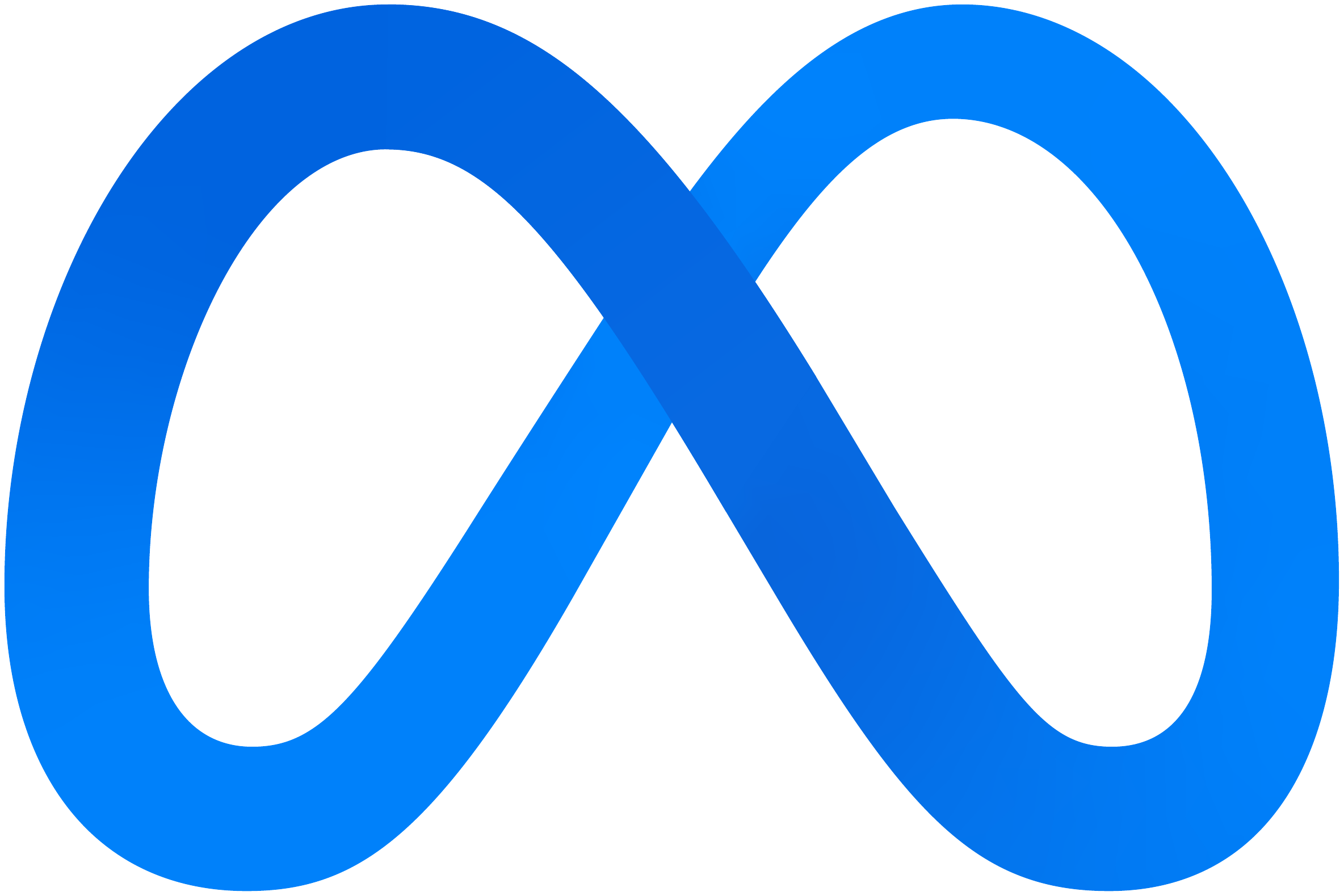}{Meta}}%
  \endgroup
}

\date{}

\begin{document}
\begin{abstract}
Multimodal models and agents have emerged to enable AI systems to learn from and interact with the real world more like humans by integrating information across multiple modalities.
However, enabling multimodal agents to continually acquire useful information from complex, heterogeneous multimodal data remains a significant challenge as they move toward long-horizon tasks and lifelong settings.
Addressing this challenge requires effective memory mechanisms that can maintain continuity across sessions, adapt to changing users and environments, and refine memory-management policies through accumulated experience. 
We introduce a two-axis analytical framework for multimodal agent memory
structured around two key questions: how multimodal information is formed and represented~(Axis~I, \emph{Formation and Representation}), and how its management adapts with accumulated experience~(Axis~II, \emph{Memory Management}).
Axis~I, \emph{Formation and Representation}, characterizes what information is retained and how it is encoded~(e.g., modality-specific embeddings, textual and structured records, latent and parametric representations) to support subsequent reasoning and action.
Axis~II, \emph{Memory Management}, distinguishes three tiers: T1 Static Memory reuses fixed memory; T2 Dynamic Memory updates memory through a stable policy; and T3 Self-Evolving Memory updates memory through an evolving policy.
Based on this framework, 
we synthesize recurring design patterns and derive principles for balancing source fidelity with abstraction, maintaining temporal validity and provenance, grounding memory use in recoverable evidence, and evolving memory policies over time. 
Furthermore, we
discuss the challenges and potential future directions for the memory mechanism of multimodal agents.
\end{abstract}
\maketitle
\clearpage

\begingroup
  \setcounter{tocdepth}{2}
  \setlength{\parskip}{0pt}
  \small
  \tableofcontents
\endgroup
\clearpage

\section{Introduction}
\label{sec:introduction}

Large foundation models~\citep{gpt56,claudeopus5,2026deepseekv4, googledeepmind_gemini37flash}, such as LLMs and MLLMs, have become the reasoning and perception backbones of AI agents that interpret goals, plan, use tools, and act through interaction with users and environments. 
An agent places the model within a continuing loop of observation, reasoning, and action~\citep{xi2025rise, wei2026agenticreasoning, plaat2025jair-agentic, codexharness}, increasingly spanning multimodal digital and physical environments.
As this interaction unfolds over time, later decisions often depend on information and experience accumulated beyond the immediately available context. \textit{Memory} provides the persistent state that bridges these interactions, carrying relevant information from prior observations and experience into subsequent reasoning and action~\citep{hu2026memoryageaiagents,zhang2025memorymechanisms,huang2026agentmemory}.

Memory is therefore a core component of AI agents~\citep{amem2025,hu2026memoryageaiagents}. 
At the most basic level, it enables continuity by preserving task state and relevant experience beyond the current interaction, allowing agents to operate coherently over long horizons~\citep{packer2023memgpt} and across repeated interactions.
Beyond continuity, persistent memory allows agents to accumulate information over time, supporting personalization through user-specific experience~\citep{hong2026tame} and maintaining consistent behavior as users, tasks, and environments evolve~\citep{zhong2024memorybank}.
By retaining task outcomes, feedback~\citep{reflexion2023}, failure traces~\citep{visualclaw2026}, and successful strategies~\citep{fan2026rrm}, memory enables agents to refine future decisions and accumulate reusable knowledge~\citep{optimus12024} and skills~\citep{voyager2023}. 

The growing demand for agents that can interpret visual information~\citep{visualmem2026,visrag2024}, process temporally structured video and speech~\citep{hippomm2025,someki2026planragaudio}, reason about and navigate 3D spaces~\citep{conceptgraphs2024}, and interact with graphical and physical environments~\citep{agashe2024agentsopenagentic,liu2025dynamem} has made multimodal capabilities increasingly essential. 
To operate in these settings, an agent must integrate heterogeneous, high-dimensional observations, capture their temporal and spatial structure, and produce appropriate actions or multimodal outputs~\citep{aesopagent2024}.
Unlike textual-only inputs, these multimodal observations cannot always be compactly or losslessly reduced to a single representation for later use. This shift does not merely increase the volume of experience an agent may encounter; it changes the nature of what must be remembered and organized. 
Consequently, multimodal memory must determine both what aspects of experience should persist and how those retained memories should remain useful as new observations accumulate and the underlying world evolves.

To investigate these two requirements, we characterize multimodal agent memory through a lifecycle with two coupled stages: \textit{formation} determines what experience is retained and in what representation, while \textit{management} governs how the resulting memory persists, changes, is accessed, and guides later reasoning and action. These stages are tightly coupled: decisions made during formation determine what evidence can later be recovered, revised, or verified, while subsequent use and new observations may require memories to be updated, reconciled, consolidated, or forgotten. Multimodality makes both stages substantially more challenging.

The first challenge arises during \textbf{formation and representation}: 
A multimodal agent must decide what experience to retain, how to divide continuous or heterogeneous observations into memory units, and in what form those memories should persist. 
Unlike text-centered interactions, which can be retained in a compact textual form that is directly consumable and editable~\citep{packer2023memgpt, zhong2024memorybank, mem02025} by LLMs, multimodal experience may contain perceptual, temporal, spatial, and acoustic information that is expensive to preserve exhaustively and difficult to compress without loss.
A memory unit may correspond to an image~\citep{visualmem2026}, image region~\citep{pixelrag2026}, video frame or clip~\citep{song2024moviechat}, audiovisual event~\citep{hippomm2025}, object track~\citep{fan2024videoagent}, interface state~\citep{mirix2025}, action--observation transition~\citep{worldlines2026}, trajectory~\citep{memoir2026}, or complete episode~\citep{ella2025}. 
Each unit may be stored as source media~\citep{visrag2024}, encoded as modality-specific features~\citep{malmm2024} or geometric state~\citep{conceptgraphs2024}, abstracted into textual~\citep{feng2026m2a} or structured records~\citep{long2025m3agent}, compressed into a latent state~\citep{qian2024videostreaming}, or internalized in model parameters~\citep{video2lora2026}.
Each choice preserves some aspects of experience while discarding or obscuring others, creating trade-offs among fidelity, efficiency, editability, source recoverability, and compatibility with downstream reasoning and action.

The second challenge arises during \textbf{memory management}: the memory state formed from prior experience must remain valid as new observations arrive, tasks progress, and environments change. The central question of multimodal memory management is not only how the memory system responds as its environment changes and experience accumulates, but also how to reconcile heterogeneous and partial observations into a consistent, evolving state. New evidence may extend an existing record~\citep{long2025m3agent}, correct an
earlier belief~\citep{ma2026reflectworld}, invalidate stale state~\citep{robomemory2026}, connect previously separate events~\citep{pyravid2026}, or provide feedback about the effectiveness of prior memory decisions~\citep{barmann2026h2emv}.
Multimodal memory management is particularly challenging because evidence is often partial and viewpoint-dependent, while the same entity or event may be represented across source media, modality-specific features, textual summaries, and structured world states~\citep{fan2024videoagent}.
Effective management must reconcile new evidence across these linked
representations~\citep{feng2026m2a}, preserve connections to the supporting
observations~\citep{hippomm2025}, and prevent stale or conflicting state from
guiding later actions~\citep{worldmm2026}.
Memory systems differ in where persistent adaptation occurs.
Their responses may range from reusing formed memory without persistent
change~\citep{scriptviz2026}, to updating memory content or organization under a
stable policy~\citep{canvas2026}, to using feedback to improve and evolve the policy
governing future memory writing~\citep{zou2026taskmem}, maintenance~\citep{barmann2026h2emv}, and
retrieval~\citep{fan2026rrm}.

\begin{figure}[H]
    \centering
    \includegraphics[width=\textwidth]
    {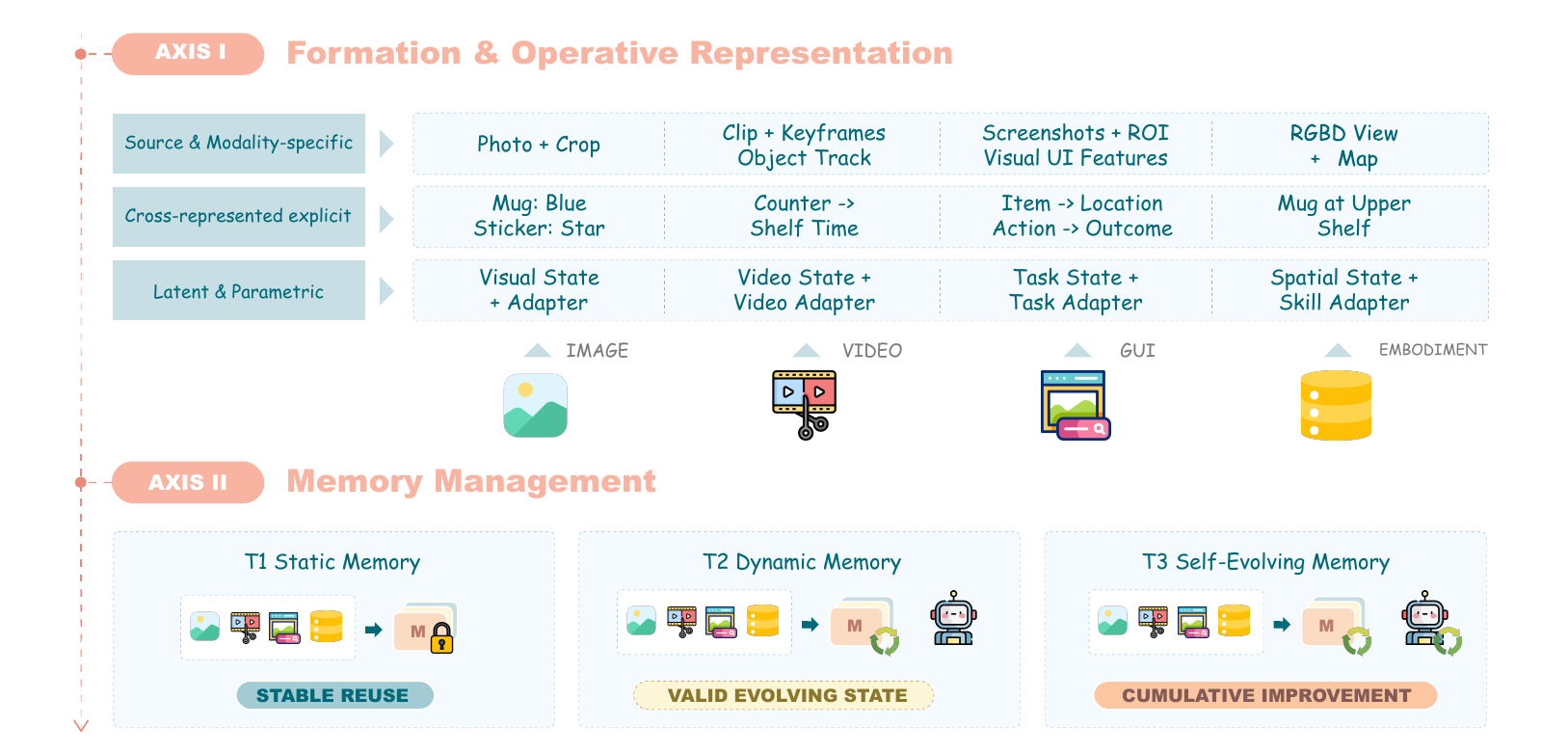}
    \caption{Two complementary axes of multimodal agent memory. Axis~I, Formation and Representation, characterizes the memory representation supplied to downstream computation. Axis~II, Memory Management, characterizes the scope of persistent change in memory state and management policy.}
    \label{fig:multimodal_memory_axes}
\end{figure}

Motivated by these lifecycle challenges, we present a unified analytical framework for multimodal agent memory spanning images and documents, video and audiovisual streams, audio and speech, GUI interaction, and embodied or 3D environments. Our framework organizes the field along two analytically distinct but operationally coupled axes, illustrated in
\Cref{fig:multimodal_memory_axes}. Axis~I, \emph{Formation and Representation}, characterizes what experience is retained, how it is organized into memory units, and the representation ultimately supplied to downstream reasoning or action. We distinguish source- and modality-specific representations, textual and structured representations, and latent and parametric representations, while separately examining hybrid architectures that combine multiple paths and often link compact abstractions to recoverable source evidence. Axis~II, \emph{Memory Management}, characterizes where persistent change occurs as experience accumulates: T1 Static Memory systems reuse formed memory without persistent modification; T2 Dynamic Memory systems update the memory state through a stable management policy; and T3 Self-Evolving Memory systems use feedback to update the policy governing future memory writing, maintenance, or retrieval. Together, these axes capture both what multimodal evidence survives into future computation and how the memory system evolves through continued experience.

Beyond providing a unified taxonomy, this study offers three broader contributions. First, we provide comprehensive coverage of multimodal agent memory across modalities, tasks, architectures, and evaluation settings, including a corpus of \ArchitectureCount{} memory architectures and \BenchmarkCount{} benchmark records in the current version of our taxonomy. Second, by jointly analyzing memory representation and management, we identify cross-cutting patterns that are difficult to observe from either dimension alone, including the prevalence of dynamic memory updates, the widespread use of hybrid representations, and the relative scarcity of deployment-time memory-policy evolution. Third, we synthesize the limitations of current systems and derive future directions around recoverable and provenance-aware memory, temporal validity and cross-representation consistency, self-improving memory management, bounded lifelong memory, governed forgetting, and lifecycle-aware evaluation.

This study focuses on
memory mechanisms in multimodal agents. 
We define memory as experience-derived state that persists beyond the step in which it is formed and later affects reasoning, generation, planning, or action.
At least one non-text modality must contribute to its formation or use; transient activations, one-pass reasoning traces, and incidental prompt context are excluded unless explicitly persisted and reused.
Existing studies of agent memory classify memory by source~\citep{zhang2025memorymechanisms}, representation~\citep{hu2026memoryageaiagents,zhang2025memorymechanisms,du2025rethinkingmemory}, function~\citep{hu2026memoryageaiagents,wu2025humanmemory,du2025rethinkingmemory}, temporal scope~\citep{wu2025humanmemory,du2025rethinkingmemory}, and operation~\citep{zhang2025memorymechanisms,du2025rethinkingmemory}, while others trace its evolution from storage to reusable experience~\citep{hu2026memoryageaiagents,huang2026agentmemory,luo2026storageexperience}.
Multimodal extensions cover LLM or MLLM memory types~\citep{jia2025aihippocampus}, VLM taxonomies~\citep{xia2026memoryvlm}, and multimodal-agent architectures and storage~\citep{xie2024largemultimodalagents,mokaria2026multimodalagentic}, benchmarks~\citep{xie2024largemultimodalagents}, and applications~\citep{xie2024largemultimodalagents,mokaria2026multimodalagentic}.
In contrast, we shift the unit of analysis from memory components to the operational lifecycle: how multimodal experience becomes persistent memory, how representation constrains subsequent retrieval, revision, and verification, and where adaptation occurs as experience accumulates.

The remainder of this paper is organized as follows. \Cref{sec:background} introduces the background, formal scope, and multimodal memory lifecycle. \Cref{sec:formation-and-representation} studies memory formation and representation, from source-preserving and modality-native memory to explicit textual and structured representations, latent and parametric memory, and hybrid source-linked designs. \Cref{sec:memory-management} examines memory management across T1 Static Memory, T2 Dynamic Memory, and T3 Self-Evolving Memory. \Cref{sec:cross-axis-analysis} analyzes how representation choices constrain memory management and synthesizes corpus-level patterns across the two axes. \Cref{sec:memory-infrastructure} examines the systems mechanisms that realize these logical memory designs under resource and coordination constraints. \Cref{sec:evaluation} examines evaluation benchmarks and applications, while \Cref{sec:agenda} discusses open challenges and future directions. \Cref{sec:conclusion} concludes the paper.

\section{Background}
\label{sec:background}

\begin{figure}[!t]
  \centering
  \includegraphics[width=\textwidth]{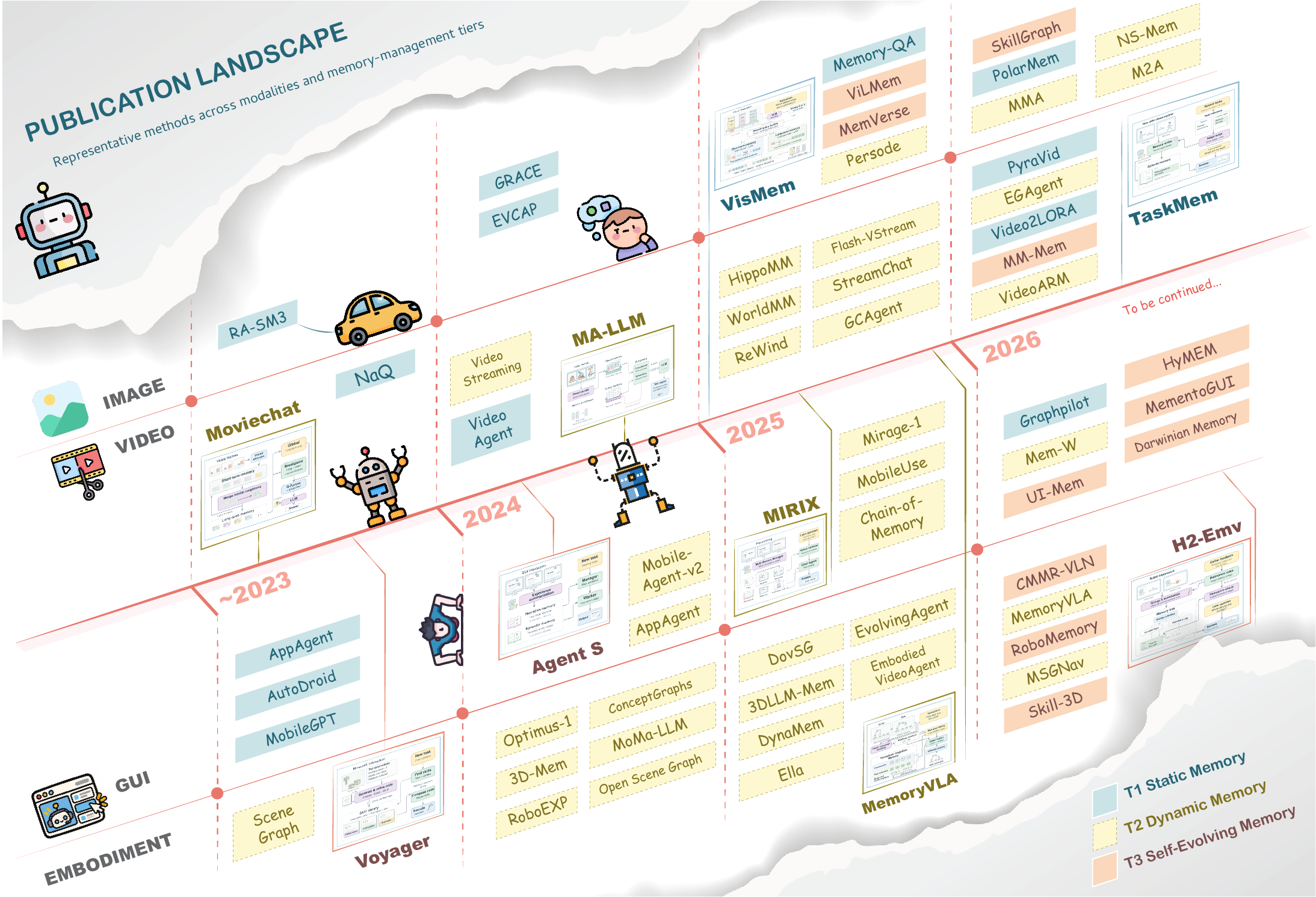}
  \caption{\textbf{Publication landscape of multimodal agent memory.}
  Representative methods across image, video, GUI, and embodied settings from 2023 to 2026, organized by memory-management tier: T1 Static Memory, T2 Dynamic Memory, and T3 Self-Evolving Memory.}
  \label{fig:memory_timeline}
\end{figure}

\begin{figure}[!htbp]
    \centering
    \includegraphics[width=0.68\textwidth]{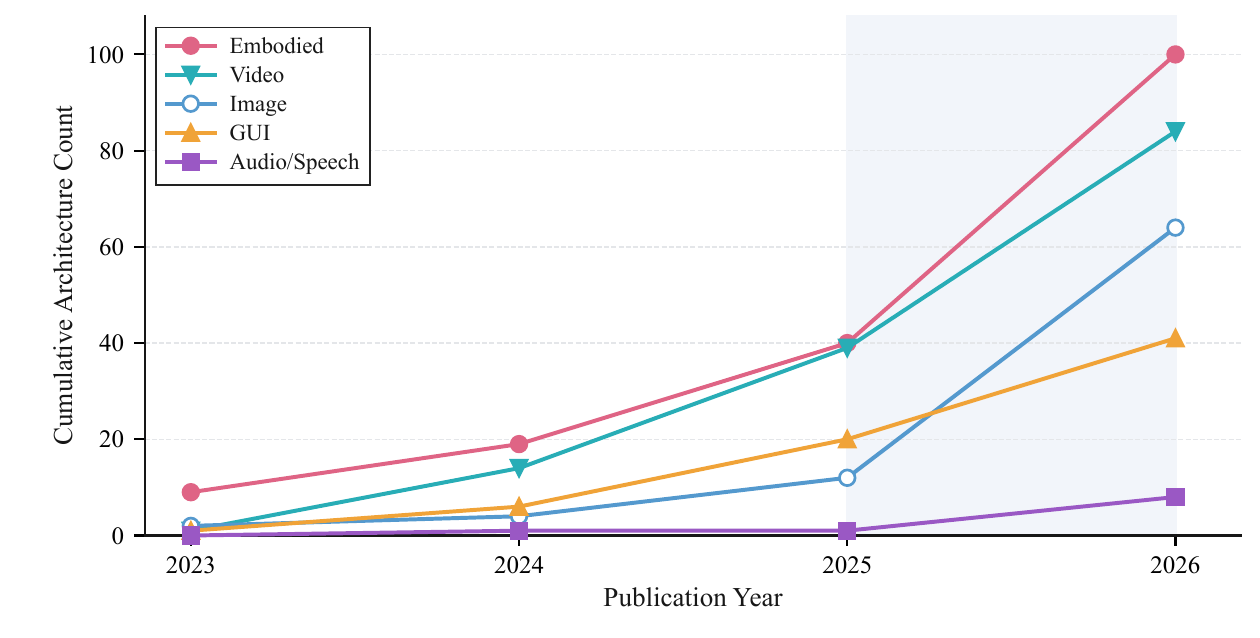}
    \caption{\textbf{Cumulative growth of multimodal-agent memory architectures differs across modalities.}
    Cumulative architecture counts in the taxonomy-current corpus are plotted by publication year from 2023 to 2026 for embodied, video, image, GUI, and audio/speech modalities. Embodied and video memory architectures show the steepest growth by 2026, followed by image and GUI systems, whereas audio/speech memory remains comparatively sparse. The shaded interval highlights the recent growth period after 2025.}
    \label{fig:modality_cumulative_growth}
\end{figure}

Multimodal agents extend foundation vision-language models from one-shot
prediction to an interaction loop in which the system observes an environment,
reasons about a goal, selects an action, and interprets the resulting
state~\citep{xie2024largemultimodalagents, mokaria2026multimodalagentic}. Their
observations may include text, images~\citep{deepimagesearch2026},
video~\citep{egagent2026}, audio~\citep{agenticasr2026}, graphical
interfaces~\citep{agmem2026}, robot poses~\citep{dovsg2025}, or other sensor
streams~\citep{echovla2025}. Because these inputs accumulate across steps and
often describe a changing world, the current context alone is rarely sufficient
for long-horizon reasoning. Memory provides continuity by retaining selected
information from earlier interaction and making it available to later decisions. \Cref{fig:memory_timeline,fig:modality_cumulative_growth} summarize representative multimodal agent memory methods across modalities, publication years, memory-management tiers, and their growth over time.

\begin{keybox}{Definition of Agent Memory}
In this study, \emph{agent memory} refers to state derived from prior observation or experience that persists beyond the instant at which it is produced and later affects reasoning, generation, or action.
\end{keybox}

This definition includes external evidence stores~\citep{sun2026pmmc}, image and
video banks~\citep{fan2024videoagent}, maps~\citep{soma2026}, scene
graphs~\citep{conceptgraphs2024}, bounded visual
buffers~\citep{wen2026eventmemagent}, recurrent states explicitly designed as
memory~\citep{videollamb2025}, experience libraries~\citep{museagent2026},
executable skills~\citep{voyager2023}, and parameters generated or adapted for
subsequent reuse~\citep{video2lora2026}. It excludes ordinary feed-forward activations, one-pass chains of thought, and incidental prompt tokens when retained state and later reuse are not part of the paper's memory contribution.

This distinction separates memory from context. Context is the information currently exposed to a model, whereas memory is the mechanism that determines what information can survive, how it is organized, and when it can re-enter the context or policy. A retrieved image can therefore be both a stored memory item and part of the current context, while an image seen once and immediately discarded is only an observation. The distinction is especially important in multimodal systems, where storing every perceptual input is computationally impractical and often unnecessary.

In this study, we use source material to denote the external stream, document, interaction, or environment from which an agent obtains information. For each architecture, an observation is a bounded input item made available to its memory-formation process at a particular interaction step. We use evidence in a functional sense to describe an observation or identifiable memory item that is used or made available to support, ground, verify, or revise a claim, decision, action, or memory state.

Text-based-agent research established many of the designs now reused by multimodal systems: MemGPT \citep{packer2023memgpt} organizes working, recall, and archival memory; Generative Agents \citep{park2023generative} transform observation streams into higher-level reflections; MemoryBank \citep{zhong2024memorybank} combines dialogue history, user modeling, and forgetting; Reflexion \citep{reflexion2023} stores feedback as reusable linguistic lessons; and Voyager \citep{voyager2023} preserves executable programs that can be retrieved and composed across tasks. Together, these works show that memory may serve as a context manager, a structured knowledge store, a reflective record, or a procedural skill library.

Multimodal memory inherits these functions but introduces a more difficult
representation problem. Images contain spatial detail that may not be captured
by a caption~\citep{visrag2024}; video and audio unfold continuously and must be
segmented over time~\citep{zhang2025dvd}; interface observations bind appearance
to actionable screen coordinates~\citep{agmem2026}; and embodied experience must
be associated across viewpoints, object identities, and changing
locations~\citep{embodiedvideoagent2025,wen2021bundletrack,wen2023bundlesdf,xie2026cari4d}. As a result, the decision to retain a
frame~\citep{contextmemory2025}, caption an event~\citep{videoem2025}, fuse a 3D
map~\citep{hiocc2026}, cache visual tokens~\citep{moviechatplus2026}, or
internalize experience in parameters~\citep{memverse2025} changes what the agent can
later recover.

Two broad strategies have emerged. Some systems preserve modality-specific
evidence or states so that later computation can return to
visual~\citep{toolmerge2026}, auditory~\citep{someki2026planragaudio}, or
geometric information~\citep{liu2025dynamem}. Others translate experience into
language~\citep{feng2026m2a}, structured symbols~\citep{long2025m3agent}, shared
latent spaces~\citep{l2vmas2026}, or parameters~\citep{video2lora2026} that are
easier to search and reuse. Native preservation keeps a wider range of
perceptual detail but increases storage and retrieval cost. Translation produces
compact and editable memory units but can remove information that was not
anticipated when the memory was constructed. This trade-off motivates our memory formation and representation axis.

\subsection{Classical Memory in AI \& Cognitive Science}

Memory has long been viewed as a requirement for intelligent behavior. In
cognitive science, models often distinguish between short-lived perceptual
storage, limited-capacity working memory~\citep{baddeley1974working}, and more
durable long-term memory~\citep{atkinson1968human}. This view suggests that
memory is not only a place to keep information, but also a mechanism for
selecting, organizing, and using information during reasoning and action.

Cognitive theories also distinguish memory by function. Episodic memory stores
experiences tied to particular events, while semantic memory stores general
knowledge that can be reused across contexts~\citep{tulving1972episodic}.
Procedural memory supports skills and action routines~\citep{squire1992memory}.
These distinctions are useful for agent systems because memory may preserve
specific past experience~\citep{himm2026new}, provide reusable
abstractions~\citep{echomc2026}, or support future behavior~\citep{eam2026}.

Classical AI studied memory from a computational perspective. Symbolic systems
represented knowledge as facts, rules, frames, scripts, or production rules,
while cognitive architectures such as SOAR~\citep{newell1994unified} and
ACT-R~\citep{anderson2004integrated} placed memory inside a control loop for
problem solving and decision making. Case-based reasoning made experience reuse
explicit by solving new problems through the retrieval and adaptation of past
cases~\citep{aamodt1994case}. These ideas anticipate many concerns in modern
agent memory, including indexing~\citep{hymem2026},
retrieval~\citep{agentcvr2026}, consolidation~\citep{flexmem2026},
forgetting~\citep{darwinianmem2026}, and reuse~\citep{gems2026}.

Neural models changed the implementation of memory. Recurrent networks store
information in hidden states~\citep{hochreiter1997long}, while attention-based
models can use the context as temporary memory~\citep{vaswani2017attention}.
Memory-augmented neural networks further introduce external modules that can be
read and written during computation~\citep{graves2014neural, weston2014memory}.
For modern agents, the central issue is therefore not only how information is
represented internally, but also what experience persists beyond the current
computation and how that retained information is later accessed or changed. This
perspective connects classical memory research to the two functional stages
considered in this study: formation and representation, which
determine what experience is retained and in what form it reaches downstream
computation, and memory management, which determines how retained memory is
accessed, maintained, and adapted over time.

\subsection{Foundations of Modern Agentic Memory}

Retrieval-augmented generation (RAG) provides an important conceptual foundation for modern agentic memory by separating model parameters from an external, queryable information source \citep{lewis2020retrieval}.
At inference time, retrieved evidence is introduced into the model context, enabling access to information beyond parametric knowledge and the current input. 
In its standard form, however, RAG primarily addresses how stored information is accessed. It does not by itself specify how an agent should form memory from its own experience, maintain that memory over time, or revise it through later interactions.

Modern agent memory broadens this retrieval-centered view by treating prior interaction and experience as persistent state that can be retained, transformed, and reused across tasks or sessions.
Systems such as MemGPT \citep{packer2023memgpt}, Generative Agents \citep{park2023generative}, and MemoryBank \citep{zhong2024memorybank} helped establish this shift from external knowledge retrieval toward persistent experience management. 
Taken together, these systems move memory beyond factual retrieval toward persistence, transformation, and experience reuse across interactions.
They provide conceptual foundations for multimodal agent memory, where the same problems arise over richer forms of experience and a broader range of representations.

When considering how long-horizon agents retain and use information accumulated over extended interactions, it is important to distinguish three closely related concepts: \emph{long-context processing}, \emph{context management}, and \emph{memory}. Long-context processing refers to a model's capacity to capture and use information distributed across extended input sequences at manageable computational cost. Context management determines which information is included in the active context at each inference step and how it is presented to the model. Memory, as defined in this study, denotes experience-derived state that persists beyond the current inference step and subsequently influences reasoning, generation, planning, or action. Depending on its representation and integration, memory can support long-context processing by carrying historical information across input segments and support context management by selecting relevant memory content as context for the model. Long-context processing and memory augmentation serve complementary aspects for building long-horizon and lifelong agents: memory extends access to information beyond the active context, while long-context models integrate retained information from memory into current computation~\citep{wu2022memorizing,he2025hmt}. Beyond these functions, persistent memory enables agents to accumulate experience across interactions, tasks, and sessions, providing a substrate for inference-time adaptation and longer-term agent evolution.

\subsection{The Multimodal Memory Lifecycle}

\begin{figure}[tbp]
    \centering
    \includegraphics[width=0.90\textwidth]
    {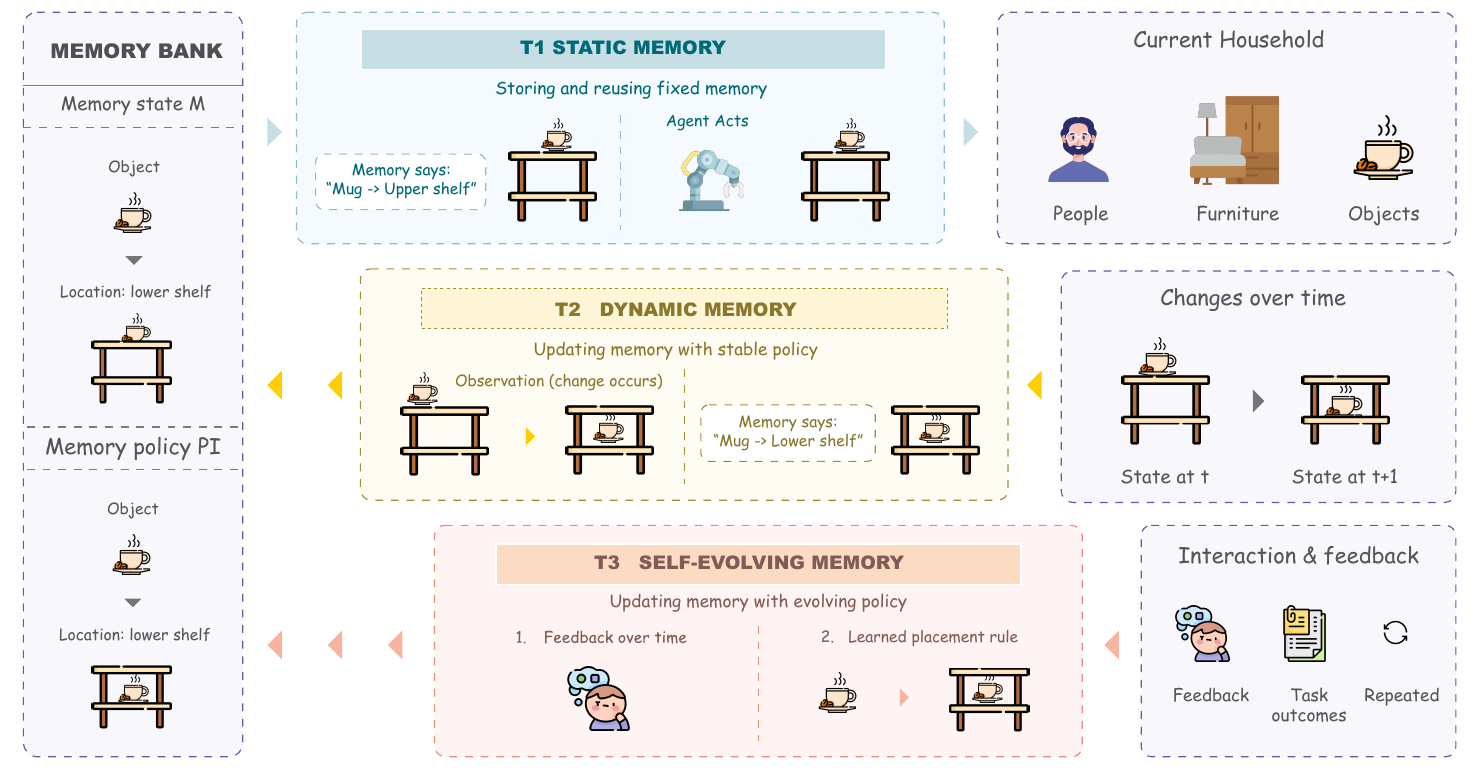}
    \caption{The multimodal memory lifecycle illustrated through a household‑robot example. The same experience forms alternative memory representations, managed as Static, Dynamic, or Self-Evolving Memory, and later accessed and used for action.}
    \label{fig:embodied_memory_lifecycle}
\end{figure}

\newcommand{\MMTaxoFont}{\fontsize{6.5}{8.5}\selectfont}

\definecolor{MMLinkBlue}{HTML}{3B35C8}
\definecolor{MMStageFill}{HTML}{ECEEF4}
\definecolor{MMCategoryFill}{HTML}{FFFFFF}
\definecolor{MMLine}{HTML}{4D4D4D}
\definecolor{MMProgress}{HTML}{6B66C7}

\definecolor{MMEdgeA}{HTML}{35701F} \definecolor{MMMidA}{HTML}{B4D9A2} \definecolor{MMSoftA}{HTML}{E7F3E0}
\definecolor{MMEdgeB}{HTML}{1F5490} \definecolor{MMMidB}{HTML}{A9CAEE} \definecolor{MMSoftB}{HTML}{E2EDF9}
\definecolor{MMEdgeC}{HTML}{52389C} \definecolor{MMMidC}{HTML}{C0B0E6} \definecolor{MMSoftC}{HTML}{E9E3F7}
\definecolor{MMEdgeD}{HTML}{14706E} \definecolor{MMMidD}{HTML}{A2D6D5} \definecolor{MMSoftD}{HTML}{DFF1F0}
\definecolor{MMEdgeE}{HTML}{8A6410} \definecolor{MMMidE}{HTML}{F0DA9E} \definecolor{MMSoftE}{HTML}{FAF0D9}
\definecolor{MMEdgeF}{HTML}{A94E14} \definecolor{MMMidF}{HTML}{F5CBA5} \definecolor{MMSoftF}{HTML}{FCE8D6}
\definecolor{MMEdgeG}{HTML}{99293C} \definecolor{MMMidG}{HTML}{F2BDC3} \definecolor{MMSoftG}{HTML}{FBE0E3}

\definecolor{MMEdgeE}{HTML}{246678} 
\definecolor{MMMidE}{HTML}{CAE4E7}  
\definecolor{MMSoftE}{HTML}{ECF6F7}

\definecolor{MMEdgeF}{HTML}{817220} 
\definecolor{MMMidF}{HTML}{FDF9CF}  
\definecolor{MMSoftF}{HTML}{FEFDEE}

\definecolor{MMEdgeG}{HTML}{7C4E4E} 
\definecolor{MMMidG}{HTML}{FCD8BC}  
\definecolor{MMSoftG}{HTML}{FEF1E8}

\newcommand{\MMSectionRef}[1]{%
  \hyperref[#1]{\textcolor{MMLinkBlue}{(\S\ref*{#1})}}%
}

%
\newlength{\MMColStage}     \setlength{\MMColStage}{22mm}
\newlength{\MMColCategory}  \setlength{\MMColCategory}{30mm}
\newlength{\MMColFunction}  \setlength{\MMColFunction}{32mm}
\newlength{\MMColTechnique} \setlength{\MMColTechnique}{32mm}
\newlength{\MMColPaper}     \setlength{\MMColPaper}{68mm}
\newlength{\MMSpanPaperOne} \setlength{\MMSpanPaperOne}{62mm}

\newcommand{\MMPaperBox}[2]{%
  \parbox[t]{#1}{%
    \MMTaxoFont
    \raggedright\hyphenpenalty=10000\exhyphenpenalty=10000%
    \strut #2\strut\par
  }%
}
\newcommand{\MMAlignedPapers}[1]{\MMPaperBox{\MMSpanPaperOne}{#1}}

\newcommand{\MMNodeText}[2]{%
  \parbox[c]{#1}{%
    \MMTaxoFont
    \centering\hyphenpenalty=10000\exhyphenpenalty=10000%
    \strut #2\strut\par
  }%
}

\newcommand{\MMNamedPaper}[2]{%
  \textcolor{black}{#1}~\textcolor{red}{\citep{#2}}%
}

\newcommand{\MMLinkedPaper}[3]{%
  \textcolor{black}{#1}~\href{#3}{\textcolor{red}{(#2)}}%
}

\newcommand{\MMPapersSourceEvidence}{%
  \MMNamedPaper{PIXELRAG}{pixelrag2026};
  \MMNamedPaper{AGMem}{agmem2026};
  \MMNamedPaper{ToolMerge}{toolmerge2026};
  \MMNamedPaper{M3-Agent}{long2025m3agent};
  \MMNamedPaper{VisRAG}{visrag2024}%
}

\newcommand{\MMPapersModalitySpecific}{%
  \MMNamedPaper{MemoryVLA}{memoryvla2026};
  \MMNamedPaper{Flash-VStream}{flashvstream2025};
  \MMNamedPaper{DynaMem}{liu2025dynamem};
  \MMNamedPaper{MovieChat}{song2024moviechat}%
}

\newcommand{\MMPapersTextual}{%
  \MMNamedPaper{Video-EM}{videoem2025};
  \MMNamedPaper{MIRIX}{mirix2025};
  \MMNamedPaper{VideoAgent}{fan2024videoagent}%
}

\newcommand{\MMPapersStructured}{%
  \MMNamedPaper{PyraVid}{pyravid2026};
  \MMNamedPaper{WorldMM}{worldmm2026};
  \MMNamedPaper{TAME}{hong2026tame};
  \MMNamedPaper{PolarMem}{polarmem2026}%
}

\newcommand{\MMPapersLatent}{%
  \MMNamedPaper{TRAM}{tram2026};
  \MMNamedPaper{Light-Omni}{lightomni2026};
  \MMNamedPaper{L$^{2}$-VMAS}{l2vmas2026};
  \MMNamedPaper{Mem-W}{memw2026}%
}

\newcommand{\MMPapersParametric}{%
  \MMNamedPaper{Frames2LoRA}{video2lora2026};
  \MMNamedPaper{SE-GA}{sega2026};
  \MMNamedPaper{VLA-Pro}{vlapro2026};
  \MMNamedPaper{MemEIC}{memeic2025}%
}

\newcommand{\MMPapersHybrid}{%
  \MMNamedPaper{VimRAG}{vimrag2026};
  \MMNamedPaper{FOLIO}{folio2026};
  \MMNamedPaper{MementoGUI}{mementogui2026};
  \MMNamedPaper{ObsGraph}{obsgraph2026};
  \MMNamedPaper{HiMe}{hime2026};
}

\newcommand{\MMPapersTOneStorage}{%
  \MMNamedPaper{VideoAgent}{fan2024videoagent};
  \MMNamedPaper{PIXELRAG}{pixelrag2026};
  \MMNamedPaper{VisRAG}{visrag2024};
  \MMNamedPaper{AutoDroid}{wen2024autodroid}%
}

\newcommand{\MMPapersTOneRetrieval}{%
  \MMNamedPaper{PIXELRAG}{pixelrag2026};
  \MMNamedPaper{VisRAG}{visrag2024};
  \MMNamedPaper{VideoAgent}{fan2024videoagent};
  \MMNamedPaper{Memory-QA}{memoryqa2025}%
}

\newcommand{\MMPapersTTwoUpdate}{%
  \MMNamedPaper{M2A}{feng2026m2a};
  \MMNamedPaper{M3-Agent}{long2025m3agent};
  \MMNamedPaper{DynaMem}{liu2025dynamem};
  \MMNamedPaper{RoboEXP}{jiang2024roboexp}%
}
\newcommand{\MMPapersTTwoConsolidation}{%
  \MMNamedPaper{MovieChat}{song2024moviechat};
  \MMNamedPaper{StreamChat}{xiong2025streamchat};
  \MMNamedPaper{NS-Mem}{jiang2026nsmem};
  \MMNamedPaper{ReMA}{chen2026mmlifelong};
  \MMNamedPaper{FluxMem}{xie2026fluxmemadaptivehierarchicalmemory}%
}
\newcommand{\MMPapersTTwoRevision}{%
  \MMNamedPaper{RoboEXP}{jiang2024roboexp};
  \MMNamedPaper{ReflectWorld-MM}{ma2026reflectworld};
  \MMNamedPaper{DovSG}{dovsg2025};
  \MMNamedPaper{M2A}{feng2026m2a}%
}
\newcommand{\MMPapersTTwoReorganization}{%
  \MMNamedPaper{MovieChat}{song2024moviechat};
  \MMNamedPaper{M3-Agent}{long2025m3agent};
  \MMNamedPaper{NS-Mem}{jiang2026nsmem};
  \MMNamedPaper{TAME}{hong2026tame}%
}
\newcommand{\MMPapersTTwoForgetting}{%
  \MMNamedPaper{EventMemAgent}{wen2026eventmemagent};
  \MMNamedPaper{StreamChat}{xiong2025streamchat};
  \MMNamedPaper{KARMA}{wang2024karma};
  \MMNamedPaper{OCR-Memory}{li2026ocrmemory}%
}

\newcommand{\MMPapersTThreeWrite}{%
  \MMNamedPaper{TaskMem}{zou2026taskmem};
  \MMNamedPaper{ABot-AgentOS}{tian2026abot}%
}
\newcommand{\MMPapersTThreeMaintenance}{%
  \MMNamedPaper{H$^{2}$-EMV}{barmann2026h2emv}%
}
\newcommand{\MMPapersTThreeRetrieval}{%
  \MMNamedPaper{RRM}{fan2026rrm};
  \MMNamedPaper{ABot-AgentOS}{tian2026abot};
  \MMNamedPaper{SkeMex}{sun2026skemex}%
}

\begin{figure}[p]
  \centering
  \begin{adjustbox}{max width=\textwidth, max totalheight=0.88\textheight, keepaspectratio}
  \begin{forest}
    for tree={
      grow=east,
      parent anchor=east,
      child anchor=west,
      anchor=west,
      draw=MMLine,
      line width=0.4pt,
      rounded corners=2pt,
      edge={draw=MMLine, line width=0.4pt, -},
      edge path={
        \noexpand\path[\forestoption{edge}]
          (!u.parent anchor) -- +(1.5mm,0) |- (.child anchor)
          \forestoption{edge label};
      },
      font=\MMTaxoFont,
      align=center,
      inner xsep=3pt,
      inner ysep=2.5pt,
      l sep=3mm,
      s sep=1.6mm,
      fit=band,
    },
    before drawing tree={
      where={n_children()==1}{
        minimum height/.wrap pgfmath arg={#1pt}{%
          max(max_y("!1")-min_y("!1"), max_y()-min_y())},
        for children={
          minimum height/.wrap pgfmath arg={#1pt}{%
            max(max_y("!u")-min_y("!u"), max_y()-min_y())},
        },
      }{},
    },
    root/.style={
      wrappable=22mm,
      fill=white,
      font=\MMTaxoFont\bfseries,
      anchor=center,
      calign=last,
      parent anchor=east,
      child anchor=west,
      inner xsep=3pt,
      inner ysep=3pt,
    },
    wrappable/.style n args={1}{
      content format={%
        \noexpand\MMNodeText{#1}{\forestoption{content}}%
      },
    },
    stage/.style={
      wrappable=\MMColStage,
      fill=MMStageFill,
      s sep=4mm,
      font=\MMTaxoFont\bfseries,
      l sep=3mm,
    },
    category/.style={
      wrappable=\MMColCategory,
      font=\MMTaxoFont\bfseries,
      s sep=1.6mm,
    },
    tier/.style={
      wrappable=\MMColCategory,
      font=\MMTaxoFont\bfseries,
      s sep=1.6mm,
    },
    function/.style={
      wrappable=\MMColFunction,
      font=\MMTaxoFont,
      minimum height=11mm,
      s sep=1.6mm,
    },
    technique/.style={
      wrappable=\MMColTechnique,
      font=\MMTaxoFont,
      minimum height=11mm,
      s sep=1.6mm,
    },
    subcategory/.style={
      wrappable=\MMColFunction,
      font=\MMTaxoFont,
      minimum height=11mm,
      s sep=1.6mm,
    },
    paper/.style={
      minimum height=11mm,
      font=\MMTaxoFont,
      align=left,
      inner ysep=2.5pt,
    },
    paperspan/.style={paper},
    tint/.style n args={3}{
      for tree={draw=#1, edge={draw=#1, line width=0.4pt}},
      fill=#2,
      for descendants={fill=#3},
      for children={fill=#2!55!white},
      for descendants={if={n_children()==0}{fill=white}{}},
    },
    tintA/.style={tint={MMEdgeA}{MMMidA}{MMSoftA}},
    tintB/.style={tint={MMEdgeB}{MMMidB}{MMSoftB}},
    tintC/.style={tint={MMEdgeC}{MMMidC}{MMSoftC}},
    tintD/.style={tint={MMEdgeD}{MMMidD}{MMSoftD}},
    tintE/.style={tint={MMEdgeE}{MMMidE}{MMSoftE}},
    tintF/.style={tint={MMEdgeF}{MMMidF}{MMSoftF}},
    tintG/.style={tint={MMEdgeG}{MMMidG}{MMSoftG}},
    [Multimodal Memory Lifecycle, root
      [Memory Management \MMSectionRef{sec:memory-management}, stage
        [T3: Self-Evolving Memory \MMSectionRef{sec:t3-feedback-evolution}, tier, tintG, name=TThreeTier
          [Retrieval-Policy Evolution \MMSectionRef{sec:t3-retrieval-policy-evolution}, function
            [\MMAlignedPapers{\MMPapersTThreeRetrieval}, paperspan]
          ]
          [Maintenance-Policy Evolution \MMSectionRef{sec:t3-maintenance-policy-evolution}, function
            [\MMAlignedPapers{\MMPapersTThreeMaintenance}, paperspan]
          ]
          [Write-Policy Evolution \MMSectionRef{sec:t3-write-policy-evolution}, function
            [\MMAlignedPapers{\MMPapersTThreeWrite}, paperspan]
          ]
        ]
        [T2: Dynamic Memory \MMSectionRef{sec:t2-fixed-policy}, tier, tintF, name=TTwoTier
          [Memory Forgetting \MMSectionRef{sec:t2-memory-forgetting}, function
            [\MMAlignedPapers{\MMPapersTTwoForgetting}, paperspan]
          ]
          [Memory Reorganization \MMSectionRef{sec:t2-memory-reorganization}, function
            [\MMAlignedPapers{\MMPapersTTwoReorganization}, paperspan]
          ]
          [Memory Revision \MMSectionRef{sec:t2-memory-revision}, function
            [\MMAlignedPapers{\MMPapersTTwoRevision}, paperspan]
          ]
          [Memory Consolidation \MMSectionRef{sec:t2-memory-consolidation}, function
            [\MMAlignedPapers{\MMPapersTTwoConsolidation}, paperspan]
          ]
          [Memory Update \MMSectionRef{sec:t2-memory-update}, function
            [\MMAlignedPapers{\MMPapersTTwoUpdate}, paperspan]
          ]
        ]
        [T1: Static Memory \MMSectionRef{sec:t1-fixed-memory}, tier, tintE, name=TOneTier
          [Retrieval \MMSectionRef{sec:t1-fixed-memory}, function
            [\MMAlignedPapers{\MMPapersTOneRetrieval}, paperspan]
          ]
          [Storage and Indexing \MMSectionRef{sec:t1-fixed-memory}, function
            [\MMAlignedPapers{\MMPapersTOneStorage}, paperspan]
          ]
        ]
      ]
      [Formation and Representation \MMSectionRef{sec:formation-and-representation}, stage
        [Hybrid Representations \MMSectionRef{sec:hybrid_representations}, category, tintD
          [Cross-Representation Composition \MMSectionRef{sec:hybrid_representations}, subcategory
            [\MMAlignedPapers{\MMPapersHybrid}, paperspan]
          ]
        ]
        [Internalized Latent and Parametric Memory \MMSectionRef{sec:latent-and-parametric-representations}, category, tintC
          [Experience-Conditioned Parametric Memory \MMSectionRef{sec:parametric_internalization}, subcategory
            [\MMAlignedPapers{\MMPapersParametric}, paperspan]
          ]
          [Persistent Multimodal Latent State \MMSectionRef{sec:shared-and-recurrent-latent_state}, subcategory
            [\MMAlignedPapers{\MMPapersLatent}, paperspan]
          ]
        ]
        [Cross-Represented Explicit Memory \MMSectionRef{sec:textual-structured-representations}, category, tintB
          [{Spatial, Temporal, and Cross-Modal Structured Memory \MMSectionRef{sec:graphs-and-structured-records}}, subcategory
            [\MMAlignedPapers{\MMPapersStructured}, paperspan]
          ]
          [Perception-to-Text Records and Summaries \MMSectionRef{sec:textual-records-and-summaries}, subcategory
            [\MMAlignedPapers{\MMPapersTextual}, paperspan]
          ]
        ]
        [Source- and Modality-Specific Representations \MMSectionRef{sec:representation-source-native}, category, tintA
          [Modality-Specific Encodings and Geometric State \MMSectionRef{sec:modality-specific-encodings}, subcategory
            [\MMAlignedPapers{\MMPapersModalitySpecific}, paperspan]
          ]
          [Retained Perceptual Source Evidence \MMSectionRef{sec:retained-source-evidence}, subcategory
            [\MMAlignedPapers{\MMPapersSourceEvidence}, paperspan]
          ]
        ]
      ]
      [, phantom]
    ]
  \end{forest}
  \end{adjustbox}
  \caption{Taxonomy of multimodal agent memory along two complementary axes: Formation and Representation, and Memory Management. T1 Static Memory stores and reuses fixed memory; T2 Dynamic Memory updates memory through a stable policy; T3 Self-Evolving Memory updates memory through an evolving policy. Representative systems may incorporate multiple mechanisms across memory lifecycle stages and therefore appear in multiple categories.}
  \label{fig:multimodal-memory-lifecycle-tree}
\end{figure}
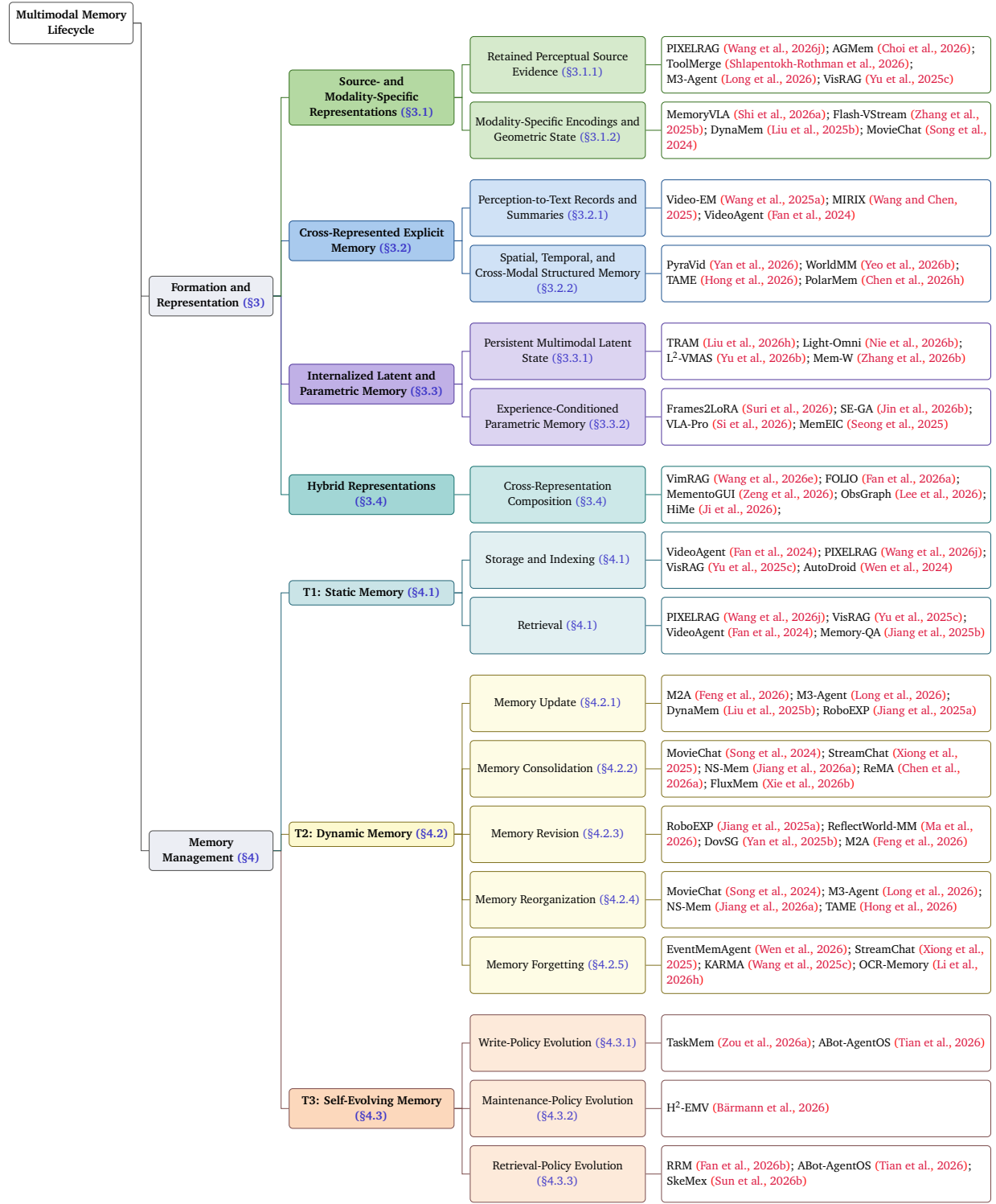

Multimodal agent memory can be understood through two coupled functional stages: (1) formation and representation and (2) memory management. Formation determines what experience is retained, how it is organized, and the representation through which it becomes available to downstream computation. Memory management determines how the resulting state persists, changes, is accessed, and is used during deployment. These stages form a lifecycle rather than independent modules: management is constrained by what formation preserves, while the outcomes of memory use may trigger further formation, revision, consolidation, or forgetting. \Cref{fig:embodied_memory_lifecycle} illustrates the two coupled stages of multimodal agent memory and the three memory-management tiers through a household-robot example.

\paragraph{Formation and Representation.}
The first stage, formation and representation, determines what experience is saved as memory, how it is organized into memory units, and how those units are represented. The definition of a memory unit varies across modalities and levels of granularity. For static images, a unit can be an entire image, region, object, or scene~\citep{visualmem2026, pixelrag2026}. For video and audio, it may correspond to a keyframe, clip, event, object track, or temporally aligned audiovisual segment~\citep{song2024moviechat, hippomm2025}. In embodied settings, memory units may further capture multimodal observations, estimated world and object states, action-observation transitions, trajectories, or complete interaction~\citep{liu2025dynamem, worldlines2026, weng2024digitaltwins}. Once these units are formed, the agent may preserve raw or compressed sensory
evidence~\citep{fan2024videoagent}, encode modality-native features or
tokens~\citep{malmm2024}, translate perceptual observations into textual or
structured symbolic records~\citep{feng2026m2a}, organize spatial and temporal
relations as event graphs, scene graphs, or spatial
maps~\citep{conceptgraphs2024}, encode accumulated experience in recurrent
latent states~\citep{qian2024videostreaming}, or internalize it through updates
to model parameters~\citep{video2lora2026}.

\paragraph{Memory Management.}
The second stage governs how formed memories are retained, updated, accessed,
and used during deployment. Axis~II distinguishes three tiers according to
whether the memory state and the policy that manages it change during use.
\emph{T1: Static Memory} stores and reuses fixed memory without modifying its
content, organization, access indices, or metadata. \emph{T2: Dynamic Memory}
updates the memory state through a stable management policy, including
appending new observations~\citep{liu2025dynamem}, revising existing
state~\citep{feng2026m2a}, merging redundant entries~\citep{song2024moviechat},
consolidating repeated evidence~\citep{digimongpt2026}, invalidating outdated
information~\citep{atmem2026}, and forgetting under capacity
constraints~\citep{mukv2026}. \emph{T3: Self-Evolving Memory} updates the
memory state while also revising the policy that governs memory writing,
maintenance, or access in response to feedback. Access includes the
retrieval~\citep{anomalyagent2026}, routing~\citep{sun2026skemex}, or
activation~\citep{latentspatialmemory2026} mechanisms that make relevant
memory available to current computation. Utilization determines how the
accessed information is incorporated into reasoning, generation, planning,
action, or learning. These processes form a coupled
management loop but remain distinct failure points: an agent may fail to
retrieve relevant evidence, retrieve it but apply it incorrectly, or produce a
correct output from model priors without using
memory~\citep{dmvbench2026, robommeinterference2026}. Management is also
constrained by formation and representation: no access mechanism can recover
information that was discarded during formation or subsequently corrupted,
invalidated, or forgotten.

Axis~I, \emph{Formation and Representation}, classifies the memory representation through which retained experience becomes available to downstream computation. Axis~II, \emph{Memory Management}, classifies systems by the scope of persistent deployment-time change they support in memory state and management policy. The resulting taxonomy is summarized in \Cref{fig:multimodal-memory-lifecycle-tree}.

\section{Axis I: Formation and Representation}
\label{sec:formation-and-representation}

\begin{figure}[H]
    \centering
    \includegraphics[width=\textwidth]{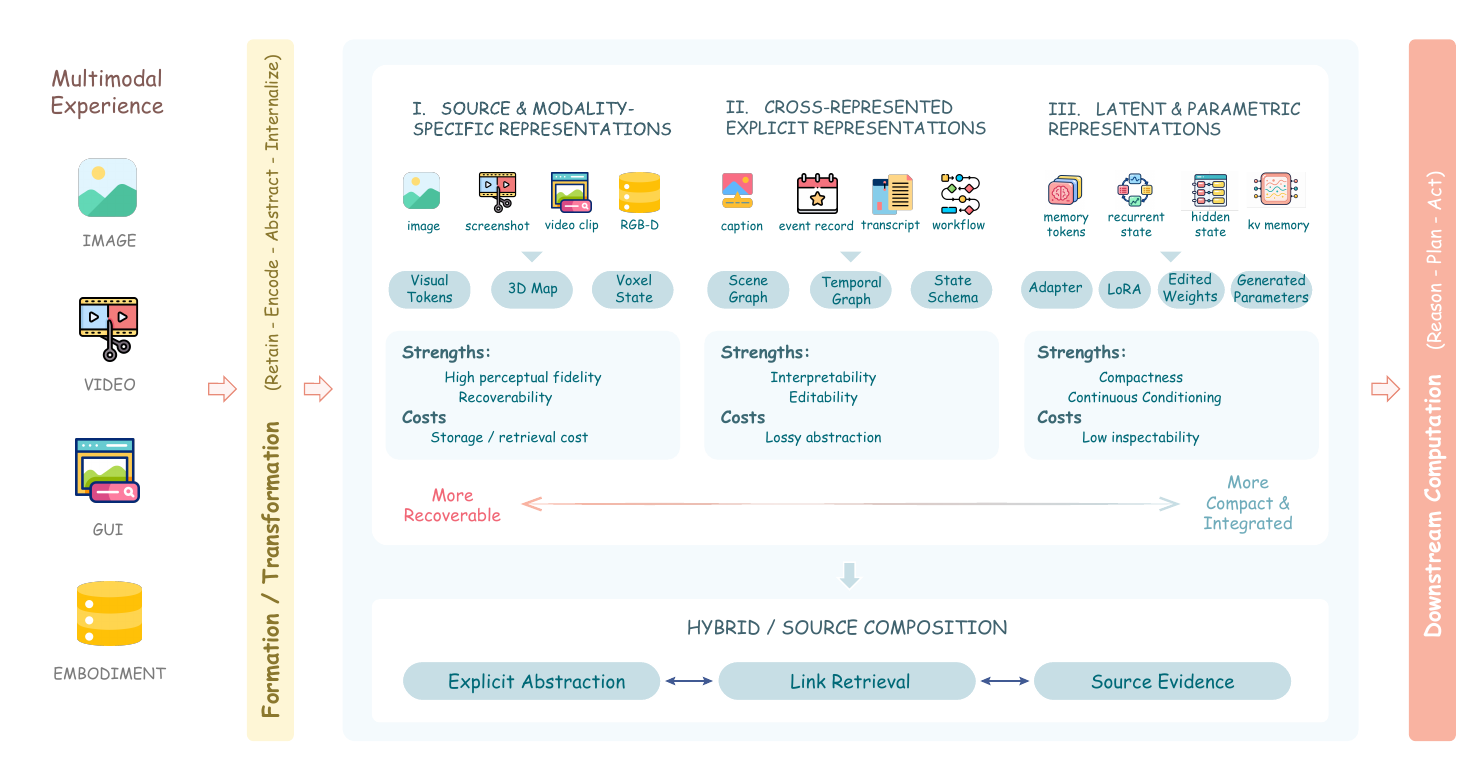}
    \caption{Representations in multimodal agent memory.
Multimodal experience may be retained close to the source modality, transformed into explicit textual or structured records, or internalized as latent states or parametric memory. We classify each memory path by the representation ultimately supplied to downstream computation, rather than by formats used only for storage or indexing. These families expose recurring trade-offs among recoverability, interpretability, compactness, and tight integration with the agent policy. Hybrid architectures combine multiple representations, often linking compact abstractions to recoverable source evidence.}
    \label{fig:operative_representation_taxonomy}
\end{figure}

In Axis~I, we examine how multimodal experience forms persistent memory and the representations that will be retrieved and used in future reasoning or action. As summarized in \Cref{fig:operative_representation_taxonomy}, we classify by the representation ultimately supplied to the reasoner or action policy, rather than merely by the format used as retrieval index. Formation mechanisms determine the representational form in which raw experiences are encoded and retained as memory. This process creates distinct memory units and determines what information remains available for subsequent use. The same image, video, audio stream, or embodied episode may be retained as source evidence~\citep{fan2024videoagent}, transformed into modality-specific visual tokens~\citep{song2024moviechat}, acoustic features~\citep{omarc2026},
temporally organized video features~\citep{malmm2024}, or geometric
state~\citep{conceptgraphs2024}, abstracted into
linguistic~\citep{videoarm2026}, symbolic~\citep{jiang2026nsmem},
relational~\citep{long2025m3agent}, or executable records~\citep{voyager2023}, or
maintained as persistent latent state~\citep{qian2024videostreaming} or
knowledge encoded in model parameters~\citep{video2lora2026}. Accordingly, we organize this axis into three families.

\begin{keybox}{Three Representation Families}
\begin{enumerate}[leftmargin=*,itemsep=3pt,topsep=2pt]
    \item \textbf{Source- and Modality-Specific Representations}
          (\Cref{sec:representation-source-native}): Retained evidence preserves
          rich detail and supports future re-encoding but incurs substantial
          storage and processing costs, whereas modality-specific encodings
          reduce these costs while preserving perceptual structure, but are
          lossy and model-specific.

    \item \textbf{Cross-Represented Explicit Memory}
          (\Cref{sec:textual-structured-representations}): Explicit abstractions
          improve interpretability and editability while discarding perceptual
          detail.

    \item \textbf{Internalized Latent and Parametric Memory}
          (\Cref{sec:latent-and-parametric-representations}): Latent or
          parametric memory integrates tightly with inference but provides less
          direct access, provenance, and control.
\end{enumerate}
\end{keybox}

Many systems combine multiple representations~\citep{dovsg2025, himm2026new, msgnav2026}. For example, a system might use
a semantic graph to locate compressed visual features and then reopen the source
image for verification~\citep{fan2024videoagent, long2025m3agent}. We therefore
discuss \emph{hybrid and evidence-linked architectures} separately as
system-level compositions rather than as a fourth representation type.

\subsection{Source- and Modality-Specific Representations}
\label{sec:representation-source-native}

At the least abstract end of the representation axis, multimodal experiences are
saved in their originating modality. We distinguish two forms. \emph{Retained
source evidence} preserves images~\citep{ptah2026}, video or audio
segments~\citep{longvideoagent2025}, and sensor
observations~\citep{dorbala2026memctrl} as media that the agent can reopen during
inference. \emph{Modality-native encodings} instead retain learned
features~\citep{diko2025rewind}, tokens~\citep{moviedreamer2024}, or geometric
state~\citep{momallm2024} that preserve modality-specific structure without
requiring the original media to be processed again. Both avoid reducing
experience to language or symbolic records, but they serve different priorities:
source evidence remains directly verifiable and interpretable, whereas encoded
state is more compact and computationally efficient but is often lossy and
difficult to interpret.

\subsubsection{Retained Perceptual Source Evidence}
\label{sec:retained-source-evidence}

\begin{keybox}{Definition of Retained Perceptual Source Evidence}
Retained perceptual source evidence preserves observations in their native
modality as addressable memory entries. The defining property is that the stored
observation remains accessible in its source modality and can be retrieved and
supplied to the downstream model as contextual evidence, rather than being
available only through a derived caption, embedding, or symbolic description.
\end{keybox}

Depending on the input, these items may
be complete images~\citep{visualmem2026}, regions or crops~\citep{agmem2026},
document screenshots~\citep{visrag2024}, sampled video
frames~\citep{zhang2025dvd}, audiovisual segments~\citep{hippomm2025}, RGB-D
observations~\citep{threedmem2024}, or observations paired with timestamps and agent
poses~\citep{dovsg2025}.

For static images and graphical user interfaces (GUIs), retaining visual
evidence preserves layout, typography, color, fine-grained appearance, and
spatial organization that textualization may
discard~\citep{visrag2024, pixelrag2026}. PixelRAG~\citep{pixelrag2026}, for
example, renders webpages as screenshot tiles, indexes them using visual
embeddings, and supplies the highest-ranked tiles directly to a VLM as pixel
inputs. In this case, the embeddings serve only as a retrieval index, whereas
the retrieved screenshots constitute the source-level visual evidence made
available during inference. Source-evidence memory can also operate at different
granularities: related systems retain complete screenshots~\citep{palui2025},
action-relevant GUI regions~\citep{agmem2026}, cropped interface
elements~\citep{mementogui2026} , or complete visual interaction histories with frame- and region-level access~\citep{vista2026}, and selectively retrieve the visual observations
most relevant to the current decision.

Video and audio pose an additional access challenge because relevant evidence must first be localized within a continuous temporal stream.
Existing systems address this challenge through different organizational structures~\citep{lian2026mmmem}.
ToolMerge \citep{toolmerge2026} retains sampled video frames and uses an LLM planner to compose complementary visual retrieval tools, including global image-text matching, region-level entity matching, and OCR. 
HippoMM \citep{hippomm2025} adopts an event-centered organization, where audiovisual experience is segmented into event records with semantic summaries, temporal boundaries, and pointers to source spans.
M3-Agent~\citep{long2025m3agent} instead organizes memory around persistent entities, linking text, image, and audio evidence to people, objects, and events over time.
These designs correspond to frame-, event-, and entity-level organization, respectively, showing that retaining source media is only one part of the problem; the evidence must also be indexed at a granularity suited to later retrieval.

Spatial indexing plays a similar role in embodied agents~\citep{cmmrvln2026}.
GraphEQA \citep{grapheqa2025} combines a hierarchical 3D scene graph with a task-relevant memory of images captured during exploration, allowing retrieved images to restore local visual details that may be abstracted away in the global graph.
DovSG \citep{dovsg2025} associates RGB-D observations and camera poses with a dynamic object-level map so that historical observations can support relocalization and local map refinement when the environment changes.
These systems show that source evidence and structured spatial memory serve complementary roles: the structured representation supports compact, global reasoning, while linked observations preserve detailed perceptual evidence for inspection, grounding, and update.

The primary advantage of preserving source-modality evidence is recoverability:
information retained in the original observation can be retrieved at different
levels of granularity for later uses that may not have been anticipated during
memory formation. This preserves fine-grained visual details,
temporal structure~\citep{videolucy2025}, acoustic
cues~\citep{someki2026planragaudio}, and observational
context~\citep{remembr2024}, while also allowing later reasoning and actions to
be grounded in and checked against specific observations.

\subsubsection{Modality-Specific Encodings and Geometric State}
\label{sec:modality-specific-encodings}

\begin{keybox}{Definition of Modality-Specific Encodings and Geometric State}
  Modality-Specific Encodings and Geometric State are memory representations that store past observations as learned perceptual features, modality tokens, or
  world-centered geometric state. They preserve task-relevant structure inherited from the source modality, such as spatial layout, temporal order, acoustic dynamics, or metric geometry.
  \end{keybox}
  
Retaining source observations maximizes recoverability, but storing, retrieving, and repeatedly encoding every image, video frame, audio segment, or sensor observation can be prohibitively expensive for long-running agents. 
A more compact strategy is to persist encoded features or geometric state. 
We assign a memory architecture to this category when the stored value is itself consumed by the reasoner or action policy and preserves task-relevant structure inherited from the originating modality, such as spatial organization, temporal order, acoustic dynamics, or metric geometry.

Video provides the clearest setting for modality-specific encoded memory because long streams require compact representations that preserve both spatial and temporal information. Representative systems differ mainly in the granularity and consolidation strategy of the retained features.
MovieChat \citep{song2024moviechat} accumulates frame-level visual tokens in a fixed-capacity short-term memory and repeatedly merges similar representations from temporally adjacent frames before transferring the consolidated features to long-term memory.
MA-LMM \citep{malmm2024} instead stores patch-level features in a visual memory bank, where they directly serve as keys and values for cross-attention; when capacity is exceeded, temporally adjacent features are merged at corresponding spatial locations.
Flash-VStream~\citep{flashvstream2025} uses a coarse-to-fine design in which low-resolution feature centroids provide broad temporal coverage and high-resolution feature maps preserve detail for representative key frames. At query time, these features are combined according to their temporal positions and supplied to the language model through the visual projector. 
These systems illustrate three forms of encoded video memory: temporally consolidated frame tokens, spatially aligned patch features, and coarse-to-fine representations of a continuing stream. Despite their different structures, all persist visual features as the memory records itself rather than merely as retrieval indices for source frames. Modality-specific encoding can therefore limit memory growth while retaining spatial and temporal structure needed for later inference, although the resulting memory remains tied to a compatible model interface.

Image memory follows a different pattern. Representative agent systems more commonly retain decodable source images or pointers to them and use visual embeddings only for indexing and retrieval~\citep{camrollagent2026, deepimagesearch2026}.
M3-Agent \citep{long2025m3agent}, for example, stores image content alongside embeddings used for similarity-based retrieval, while M$^{2}$A \citep{feng2026m2a} associates stored images with textual captions and cross-modal embeddings that support both text- and image-based search.
Other systems transform visual observations into textual~\citep{chainofmemory2025} or structured records~\citep{wen2024autodroid} and discard the source images.
UniversalRAG~\citep{yeo2026universalrag} extends this indexing pattern across heterogeneous corpora by routing queries to modality- and granularity-specific retrievers.
MIRIX~\citep{mirix2025} follows this approach for screenshot-based memory.
These designs show that image encoders frequently support memory formation and retrieval, but their feature vectors less commonly serve as persistent memory values consumed directly by the downstream model.

Persistent encoded audio memory remains underexplored among the analyzed systems. Unlike speaker embeddings used only as retrieval keys~\citep{egomem2025}, such memory would preserve acoustic information as the stored value and expose it directly to the
  downstream model.
OMAC \citep{omarc2026}, introduced in O-MARC, provides a relevant encoding mechanism. 
It retains high-importance audio tokens as anchors and merges nearby discarded tokens into them through weighted aggregation. The resulting tokens remain temporally ordered and are consumed directly by a compatible omnimodal backbone without reconstructing the original waveform. 
OMAC therefore illustrates how acoustic features could serve as memory unit values rather than retrieval indices. 
However, it compresses the current audiovisual context rather than preserving information across interactions, so it is better viewed as a transferable representation mechanism than as an established example of persistent agent memory. 
This distinction also highlights how little persistent audio-feature memory has been explored relative to its visual counterpart.

Embodied agents extend modality-specific memory from neural feature sequences to world-centered geometric state~\citep{mslmaps2025, hiocc2026}.
DynaMem \citep{liu2025dynamem} 
converts posed RGB-D observations into an evolving sparse voxel map in a global coordinate frame. Its voxels combine metric location, observation history, timestamps, and vision-language features, while new depth observations update or remove inconsistent state. This structure preserves metric relations and occupancy changes that would be difficult to recover from an unordered feature collection.
A different design is to retain perceptual history for action generation rather than to construct an explicit world model. MemoryVLA \citep{memoryvla2026} compresses visual-encoder outputs into perceptual tokens that query a temporally indexed long-term bank. Retrieved history is fused with the current perceptual state and used to condition the action policy, while capacity is controlled by merging similar temporally adjacent entries. The resulting bank therefore stores a bounded history of policy-relevant visual features without reconstructing source observations. 
The key distinction lies in the role of memory: DynaMem organizes memory around persistent world geometry, whereas MemoryVLA organizes it around perceptual history required by the control policy.

Across these systems, the representation is decided by the structure required by the downstream interface. 
Video memories preserve spatial and temporal organization~\citep{malmm2024}, emerging acoustic encodings retain temporally localized sound information~\citep{omarc2026}, geometric maps expose metric layout and occupancy~\citep{liu2025dynamem}, and policy-facing memories retain action-relevant perceptual history~\citep{memoryvla2026}.
Persisting these representations reduces storage, context, and repeated encoding costs while preserving information that may be difficult to express through language or symbolic records.
The resulting compression is generally non-invertible: information removed through
sampling~\citep{cnav2025}, merging~\citep{moviechatplus2026},
clustering~\citep{flashvstream2025}, spatial reduction~\citep{photoflow2026}, or
voxelization~\citep{echovla2025} cannot be recovered from the encoded state alone.
Learned feature memories are also model-specific and normally require a compatible projector, Q-Former, attention module, or policy interface.
Geometric state is more explicitly structured, but discretization and feature aggregation can still remove perceptual detail.
Modality-specific encodings and geometric state therefore trade recoverability and portability for compact, task-aligned access to perceptual and spatial structure.

\subsection{Cross-Represented Explicit Memory}
\label{sec:textual-structured-representations}

Multimodal experience can also be transformed into natural language or structured representations such as knowledge graphs. For example, an image may be stored as a caption~\citep{feng2026m2a}, while an embodied observation may be represented as a scene graph containing nodes and relations between them~\citep{conceptgraphs2024}. Because the relevant content is expressed through readable text or named graph components, it can be directly retrieved and used for reasoning or action. Compared with modality-specific encodings and latent or parametric memory, these representations can usually be inspected, searched, edited, or invalidated at the level of individual entries. However, their construction is also lossy: detailed visual information omitted during captioning, entity extraction, or relation construction usually cannot be recovered from the resulting representation alone.

Within \emph{Cross-Represented Explicit Memory}, \emph{Perception-to-Text Records and Summaries} translate multimodal experience into natural-language entries that are retrieved and supplied to the agent as text. Although these entries may be indexed by timestamps, identifiers, or embedding similarity, the content placed in the backbone LLM's context remains textual. \emph{Spatial, Temporal, and Cross-Modal Structured Memory} organizes information as graph-linked entities and relations or schema-defined fields that can be queried and updated individually. For example, a vector database of captions is textual memory, whereas a scene graph is structured memory when its nodes, relations, or topology support retrieval, reasoning, or action. Because a single architecture may combine multiple representation forms, we discuss each form in its respective section rather than assigning the architecture as a whole to a single category.

\subsubsection{Perception-to-Text Records and Summaries}
\label{sec:textual-records-and-summaries}
\begin{keybox}{Definition of Perception-to-Text Records and Summaries}
  Perception-to-Text Records and Summaries transform multimodal observations into natural language memory units, such as captions, event descriptions, facts,
  profiles, summaries, or procedures. These units are retrieved and supplied directly to an LLM as textual context.
  \end{keybox}

Textual memory expresses past experience as descriptions~\citep{augustus2025}, event records~\citep{streamarena2026}, summaries~\citep{emembench_embodied2026}, facts~\citep{omnisimplemem2026}, profiles~\citep{mirix2025}, or natural-language procedures~\citep{fan2026rrm}.
Its memory unit is a linguistic entry that can be retrieved through semantic similarity, lexical matching, temporal filtering, or structured querying and then inserted directly into an LLM context. 
This representation aligns naturally with language-model reasoning and permits human inspection and correction without requiring a decoder. 
Textual memory is also well suited to evolving state because updates, corrections, and overrides can be directly incorporated into compact, ordered records, whereas modifying or reconciling multimodal evidence and its embeddings is less straightforward 
. It also separates storage from the original perception model: a textual memory produced by one visual or audio encoder can generally be read by another language model.

Long video provides a clear setting for this strategy because retaining every frame is expensive, while many downstream questions concern events, actions, and state changes that can be summarized linguistically.
VideoAgent \citep{fan2024videoagent} constructs a compact semantic memory of temporal event descriptions and allows an LLM agent to query these records while deciding when additional visual evidence is needed.
Video-EM \citep{videoem2025} makes the event abstraction more explicit. It localizes potentially relevant moments, groups them into temporally coherent events, and writes grounded episodic records containing temporal indices and when, where, what, and who cues. A chain-of-thought retrieval process then selects a minimal set of event memories for answering the current question. 
ReMA~\citep{chen2026mmlifelong} processes a lifelong video stream as fixed-duration clips, writes timestamp-aligned descriptions into a vector-indexed memory bank, and merges temporally overlapping observations into updated summaries. Its controller later searches these textual records and adds the results of query-directed inspections back into the same belief state.
Across these systems, linguistic abstraction changes the basic memory unit from a frame to a semantically delimited event or observation, allowing temporally distant evidence to be addressed without repeatedly processing the complete video.

Textual memory is also prominent in GUI agents, where screenshots and interactions can
be converted into persistent facts~\citep{chainofmemory2025}, events~\citep{palui2025},
and workflows~\citep{darwinianmem2026}.
MIRIX \citep{mirix2025} filters visually redundant screenshots and routes new screen and conversational evidence to specialized memory managers.
These managers construct natural-language summaries, event details, semantic records, workflow steps, document excerpts, and user-profile facts across six memory components. 
During retrieval, the system searches these components using embedding-, BM25-, or string-based methods, and the selected entries are ultimately supplied to the chat agent as labeled text.
The resulting representation abstracts transient interface appearance into concepts that remain useful across later interactions, such as user activities, previously accessed resources, or recurring workflows.

Persistent image memory is less commonly represented solely in text. 
A single image is comparatively inexpensive to retain, while future queries may depend on typography, color, layout, object appearance, or fine spatial relations that cannot always be anticipated when a caption is written.
Purely textual image memory is therefore most useful when images occur as part of a longer personal or interaction history, such as a screenshot stream or camera roll \citep{mirix2025, camrollagent2026}, and the relevant memory objects are people, activities, events, or preferences rather than arbitrary visual details. 
In such settings, captions and event summaries provide efficient semantic access, while
source images can be retained separately when later visual verification is
required~\citep{memoryqa2025}. 
This pattern explains why text-only image memory is less dominant than textual event memory for long video and interaction histories.

Pure textual memory is less common for embodied experience as well. 
Text can summarize observed events and completed tasks, but embodied action often depends on precise and current information about object locations, metric distances, containment, connectivity, and other spatial relations that cannot be reliably reconstructed from isolated free-form text records. 
Embodied systems therefore typically associate linguistic descriptions with spatial maps or entity-centered state rather than relying on unstructured text alone \citep{embodiedlgr2026, robomemory2026}. 
Textual records remain valuable for episodic histories, task outcomes, and high-level instructions, while localization and planning generally rely on the structured representations discussed in the next section.

Across modalities, textualization provides a compact and interpretable interface between perception and reasoning. 
It supports semantic retrieval, direct insertion into model context, human editing, and consolidation of observations across modalities. 
Its primary limitation is the semantic bottleneck introduced when memories are written. 
Captions and summaries retain information judged salient by the formation model, but may omit visual details, flatten temporal or spatial relationships, discard acoustic cues, or introduce unsupported interpretations. 
Because retrieval cannot recover information that was never encoded in the textual record, the quality of textual memory depends not only on retrieval accuracy but also on the coverage, grounding, and factual reliability of the textualization process.

\subsubsection{Spatial, Temporal, and Cross-Modal
Structured Memory}
\label{sec:graphs-and-structured-records}

\begin{keybox}{Definition of Spatial, Temporal, and Cross-Modal Structured Memory}
  Spatial, Temporal, and Cross-Modal Structured Memory represents experience as schema-defined records or graphs containing addressable entities, attributes,
  relations, state variables, and spatial, temporal, or cross-modal links. The organization of these elements is part of the operative memory and can be queried, traversed, or updated to support agent reasoning, planning, and action.
  \end{keybox}

Graphs and structured records represent experience through explicitly organized and addressable entities~\citep{ella2025}, attributes~\citep{scenegraphgrounder2026}, relations~\citep{jiang2024roboexp}, state variables~\citep{wang2024karma}, temporal links~\citep{scrapmem2026}, or spatial structures~\citep{momallm2024}. Their meaning derives not only from the content of individual nodes or records but also from the relationships and organization among them. Accordingly, access may involve field-level lookup, graph traversal, temporal filtering, relational propagation, or spatial queries, rather than treating each entry as an independent text passage. Such representations are particularly valuable when an agent must determine not only what occurred, but also how entities are related, which state is currently valid, or where an action should be executed.

Embodied agents provide particularly clear examples of structured memory, as navigation and manipulation often require persistent representations of objects, their spatial and functional relationships, and, in dynamic environments, changes in scene state over time. ConceptGraphs \citep{conceptgraphs2024} lifts open-vocabulary detections from successive RGB-D observations into a 3D coordinate frame and incrementally fuses them into object nodes and relational edges. The resulting scene graph supports language-grounded, object-centric, and spatial access, allowing an agent to retrieve entities and their locations for downstream perception and planning. DovSG \citep{dovsg2025} extends this design to changing environments. It maintains persistent 3D object nodes with geometric and semantic features, connects them through spatial relations, and locally refines regions affected by robot interactions or environmental changes. The task planner reads target coordinates, object features, and graph relations to decompose instructions and support navigation, grasping, and placement. In this case, the graph is not merely an index over observations; it is the current world state on which planning operates. GraphEQA \citep{grapheqa2025} demonstrates how structured spatial memory can guide active information acquisition. During exploration, it incrementally constructs a hierarchical 3D metric-semantic scene graph containing object, room, and frontier information. The graph exposes what has already been observed, how the explored space is organized, and where additional evidence may be obtained. This state is supplied to a VLM planner that decides whether the available memory is sufficient to answer the question or whether the agent should continue exploring. Structured spatial memory therefore supports both retrospective access and prospective action: it preserves previously acquired observations while guiding the agent toward locations where additional evidence can be gathered.

Video agents impose similar structure on temporal experience. PyraVid \citep{pyravid2026} organizes streaming video into a three-level graph comprising fine-grained fact nodes, clip-level summaries, and a global narrative. Hierarchical aggregation edges connect facts to clips and clips to the global memory, while additional logical links connect related evidence across time. Retrieval begins from query-relevant seed nodes, checks whether the evidence is sufficient, and traverses the graph to expand or prune the active evidence set. WorldMM \citep{worldmm2026} constructs multiple temporal knowledge graphs from video segments at different resolutions and an evolving semantic graph from consolidated entity-action-entity triplets. Episodic access uses graph-based Personalized PageRank followed by cross-scale reranking, while semantic access performs edge-focused traversal over the consolidated graph. New triplets can update or remove outdated and conflicting relations. These systems show how explicit temporal and relational structure can preserve event dependencies that would be difficult to recover from an unordered collection of captions.

For persistent image memory, structured representations are particularly useful when an agent must track entities and their attributes across multiple observations. TAME \citep{hong2026tame}, for example, maintains a Dynamic State Memory for transient or changing attributes and a Static Personalized Memory for stable attributes, both indexed by personalized concept identifiers. Attributes that remain stable can migrate from dynamic to static memory, while outdated attributes are removed from the bounded dynamic store. During retrieval, the system identifies the personalized concept in the current image and retrieves its associated attributes from both memories. Separating dynamic from stable attributes helps the agent distinguish transient observations, such as a recent appearance or condition, from persistent user-specific properties.


Structured visual memory can also encode the presence status of remembered concepts. PolarMem \citep{polarmem2026} constructs a polarized graph that links visual memories to concept nodes through HAS and NOT\_HAS edges, thereby distinguishing verified presence from verified absence. At query time, its logic-aware retrieval procedure enforces logical consistency before ranking candidates by semantic similarity. Memories that conflict with the query constraints are therefore suppressed even when they are semantically similar. PolarMem thus demonstrates that graph structure can represent not only associations among memory items but also positive and negative evidence. This capability is particularly valuable when an agent must explicitly rule out concepts that are visually plausible but absent from the observed evidence.

Persistent structured memory is also relatively uncommon in GUI agents. Although a GUI presents structured elements at any given moment, its layout and element hierarchy may change across application versions, window sizes, navigation states, and dynamically generated content. Memories tied to absolute screen coordinates or specific interface graphs may therefore generalize poorly across tasks and sessions. Most GUI agents more commonly retain textual interaction histories~\citep{agashe2024agentsopenagentic} or parameterized workflow templates~\citep{xiao2026uimem} that abstract away from a particular interface configuration. When structured GUI memory is used, it therefore tends to represent task state~\citep{mobileagentv22024}, interface entities~\citep{graphpilot2026}, or reusable procedures~\citep{mobilegpt2023} rather than a persistent geometric map of the interface.

Graphs and structured records support compositional queries, localized updates, temporal reasoning, and relation-aware planning that are difficult to perform over independent, free-form text. These capabilities, however, introduce substantial construction and maintenance costs. Entities must be detected and resolved across observations; attributes and relations must be extracted accurately; schemas must accommodate previously unseen concepts; and updates must maintain consistency as the environment evolves. Errors in entity creation, relation extraction, or cross-observation data association may propagate through retrieval, graph traversal, and downstream planning. Structured representations therefore improve addressability and control but do not inherently guarantee faithful memory content. Their reliability depends on both accurately transforming perceptual evidence into structured semantic content and correctly reconciling new evidence with the existing graph by updating existing entities and relations or introducing new ones.

\subsection{Internalized Latent and Parametric
Memory }
\label{sec:latent-and-parametric-representations}

While external memory preserves information as independently stored and retrievable artifacts, such as raw observations, textual records, structured entries, or modality-specific features, latent and parametric memory integrate experience more directly into the model's computation. Latent memory maintains experience as continuous states that are carried and updated during inference, whereas parametric memory internalizes experience in model weights or dynamically generated adapters that shape subsequent reasoning and action.

We distinguish two forms of memory that integrate experience directly into model computation. Persistent multimodal latent states represent experience as continuous vectors, memory tokens, hidden states, or key–value units that are separate from backbone parameters. These states can be updated during inference and integrated into the model through attention, residual connections, or learned projection and fusion modules. Recent systems apply this approach to retain visual information and intermediate reasoning states~\citep{vismem2025}, coordinate memory across visual agents~\citep{l2vmas2026}, and compress GUI interaction histories into model-compatible continuous context~\citep{memw2026}. On the other hand, parametric internalization encodes experience in model weights or experience-specific adapters. Internalized memory influences subsequent inference implicitly, without requiring retrieval or inclusion of the original multimodal evidence in the input context. For example, Frames2LoRA \citep{video2lora2026} generates a LoRA adapter from a video, enabling the adapted model to use information from that video without processing its visual tokens again. Instead, MemVerse \citep{memverse2025} periodically distills selected knowledge from an explicit long-term memory store into a parametric memory model.

Both forms provide compact access to past experience and integrate memory closely with the model's reasoning and control processes. They can reduce repeated encoding, retrieval costs, and context growth while retaining information that may be difficult to capture in language or symbolic structures. This close integration, however, limits direct control over the stored information. Latent and parametric memories are generally more difficult to inspect, edit, trace to specific observations, or transfer across models with incompatible architectures. Latent compression may discard information that cannot be recovered, while repeated parameter updates may cause interference or forgetting. 

\subsubsection{Persistent Multimodal Latent State}
\label{sec:shared-and-recurrent-latent_state}

\begin{keybox}{Definition of Persistent Multimodal Latent State}
Persistent Multimodal Latent State is one form of Internalized Latent and Parametric Memory. It represents information from past observations as continuous hidden states that remain separate from the backbone model's parameters. The state can be recurrent, shared, or both. A recurrent state is carried forward and updated as new observations arrive, whereas a shared state can be accessed by multiple model components or agents.
\end{keybox}

Common implementations use hidden states~\citep{qian2024videostreaming}, memory tokens~\citep{videollamb2025}, or key–value caches~\citep{mukv2026} and integrate them with the current input through attention, residual connections, or learned projection and fusion layers.

KV caches can also serve as persistent latent memory when they are deliberately retained and reused beyond the computation that produced them.
ReKV, for example, incrementally constructs and retains video-derived KV states and selectively restores query-relevant blocks for later question answering \citep {rekv2025}. 
MuKV further compresses multi-grained KV states to support long-horizon streaming video question answering under bounded memory \citep {mukv2026}.

Shared and recurrent latent states have been widely explored in image-based agents to preserve visual information and intermediate reasoning across inference steps.
VisMem \citep{vismem2025} uses a short-term memory former to generate latent tokens containing fine-grained evidence from the current visual input, and a long-term memory former provides higher-level semantic knowledge relevant to the current reasoning context. The model learns when to invoke each memory type and inserts the resulting latent tokens into the inference sequence. TRAM \citep{tram2026} preserves information from earlier reasoning steps so that it can inform later stages of a multi-step reasoning process. After each reasoning step, the model summarizes its internal activations and uses them to update two recurrent latent states with different decay rates. The fast state emphasizes recently completed reasoning, while the slow state retains information over a longer portion of the trajectory. Their combined representation is then added to selected decoder layers at the beginning of the next reasoning step. These methods address a common limitation of extended multimodal reasoning: visual evidence and intermediate conclusions may lose influence as the generation grows longer \citep{visualneglect2026}.

For long video streams, latent memory can reduce the cost of repeatedly processing accumulated visual context~\citep{flexmem2026}. Light-Omni \citep{lightomni2026} maintains a bounded, non-parametric global state by progressively summarizing and merging episodic scripts. This hierarchical state preserves recent events in greater detail while representing older events through higher-level summaries. Given the global state and the current multimodal input, the model generates latent features that determine whether to speak or search and produce an embedding for retrieving relevant entries from long-term memory. Light-Omni therefore adopts a hybrid design: explicit episodic scripts and their consolidated global state preserve historical information, while the generated latent features support immediate action control and context-aware retrieval. This design is well suited to long-video tasks, which require both a compact representation of global context and access to specific events, temporal details, or original multimodal evidence.

Beyond modality-specific designs for images and video, latent memory can also support information sharing among multiple vision-language agents. L$^{2}$-VMAS \citep{l2vmas2026} maintains two shared memory banks: a perception memory that stores latent representations of visual inputs and a thinking memory that stores hidden states from previous agents' reasoning. Both are organized as key-value memory units that can be retrieved and integrated into subsequent agent computation.

GUI agents use latent memory to connect interaction history with the current interface state~\citep{comem2025}. Mem-W \citep{memw2026} uses a shared trajectory-to-latent compressor to encode two sources of GUI memory: completed trajectories retrieved from an external memory bank and the expired prefix of the current interaction. The resulting memory tokens are combined with the task instruction, current screenshot, and recent observation-action context in a unified continuous embedding sequence. Retrieved trajectories provide procedural evidence from both successful and failed experiences, while the compressed in-session history preserves unfinished task progress. Both forms of experience can therefore directly inform the agent's next action without being converted into textual summaries. This representation is well suited to GUI tasks, where the relevant state is distributed across visual observations, previous actions, and intermediate task progress.

This approach extends naturally from GUI interaction to embodied control, where agents must relate current observations to previous actions and task progress~\citep{trmvla2026}. LaMem-VLA \citep{lamemvla2026} organizes an agent's interaction history into two complementary memory vaults: a short-term vault that retains recent visual information and a long-term vault that preserves task progress and action continuity. Given the current observation and instruction, the model retrieves relevant information from both vaults and compresses it into fixed-length latent memory tokens. These tokens are combined with the current visual, instruction, and action-query tokens in a unified embedding sequence, allowing historical experience to directly guide action generation . This design is well suited to Vision-Language-Action (VLAs) models because it integrates perception, memory, and control within a shared continuous representation space.

Persistent latent memory remains relatively underexplored in audio and speech agents. These systems more commonly preserve semantic content as transcripts~\citep{agenticasr2026} and represent speaker identity, prosody, and other acoustic properties as separately accessible features~\citep{someki2026planragaudio}. Such representations support direct retrieval and speaker attribution while retaining temporal and acoustic evidence that recurrent compression may discard. Consequently, persistent latent states designed to preserve general auditory experience across long interactions or multiple episodes remain less developed than their counterparts in visual reasoning, GUI interaction, and embodied control.

Across modalities, latent memory operates over different persistence horizons. Some latent states are maintained within a single reasoning trajectory \citep{vismem2025,tram2026}, whereas others preserve information across successive agents or interaction steps \citep{l2vmas2026,memw2026}. Latent representations may also be constructed on demand from retrieved long-term experience \citep{memw2026,lamemvla2026}. Despite these differences, these mechanisms share a common property: past experience is represented in a continuous form that directly contributes to subsequent model computation.

\subsubsection{Experience-Conditioned Parametric
Memory}
\label{sec:parametric_internalization}


\begin{keybox}{Definition of Experience-Conditioned Parametric Memory}
Experience-Conditioned Parametric Memory is another form of Internalized Latent and Parametric Memory. It incorporates information from past experience into learned weights or parameter-efficient modules that affect later inference.
\end{keybox}

A system may update shared model parameters through gradient-based learning~\citep{uivoyager2026} or encode experience in instance- or task-specific adapters attached to a frozen backbone~\citep{mavis2025, hylovqa2026}. Once encoded in shared weights or a selected adapter, the information influences model computation without retrieving the original observations or adding them to the input context.

Image-based parametric memory has primarily been explored through multimodal knowledge editing, where the objective is to retain newly introduced or corrected visual facts within a model. MemEIC \citep{memeic2025} adopts a hybrid design that combines explicit evidence with parametric internalization. It maintains separate LoRA adapters for visual and textual knowledge updates while keeping the pretrained backbone frozen. A visual edit updates only the visual adapter, whereas a textual edit updates only the textual adapter. During inference, the system activates the adapter associated with the query modality; when a query requires both modalities, it activates both adapters and uses a learned connector to combine their representations. This design illustrates both an advantage and a challenge of parametric memory. Incorporating an edit into learned parameters allows the updated knowledge to participate directly in subsequent computation, but repeatedly writing heterogeneous information into shared parameters can cause interference and forgetting. MemEIC addresses this problem by separating parameter updates by modality, while retaining explicit edit records externally for reliable retrieval and verification.

Video introduces a different motivation for parametric memory: supporting repeated queries without retaining and reprocessing a long sequence of visual tokens. Frames2LoRA \citep{video2lora2026} first encodes a video using a frozen vision-language model. A Perceiver-based hypernetwork then converts the resulting layer-wise representations into a video-specific LoRA adapter in a single forward pass. This adapter is attached to the frozen model and acts as a reusable parametric representation of the video. For each subsequent question, the model receives only the textual query, while information from the video influences its response through the adapter parameters; neither the original video nor its visual tokens are included in the input context. Unlike conventional per-video fine-tuning, the method does not iteratively optimize a new adapter for each video. Instead, its trained hypernetwork directly generates the adapter from the video representations. Frames2LoRA therefore demonstrates how parametric memory can convert a particular video into a compact and reusable parameter component, reducing the cost of repeated inference while sacrificing direct access to the original visual evidence.

In GUI agents, parametric memory more commonly captures reusable operational knowledge than the appearance of a particular screen~\citep{sega2026, uivoyager2026}. SE-GA \citep{sega2026} illustrates this form of experience internalization through a hybrid memory architecture. During task execution, its Test-Time Memory Extension component maintains an explicit repository containing current task state, general GUI rules, and successful trajectories from previous tasks. Its Memory-Augmented Self-Evolution pipeline subsequently uses the accumulated interaction data to improve the agent's policy through grounding supervision and reinforcement learning, with parameter-efficient LoRA updates used during self-evolution. The explicit repository makes recent and task-specific experience immediately retrievable, whereas the learned policy parameters absorb recurring interaction patterns so that they can influence future actions without replaying every source trajectory. SE-GA therefore exemplifies a slower form of parametric consolidation in which accumulated GUI experience is converted into a more general action policy. Fully parametric, episode-specific GUI memory remains uncommon because successful interaction often depends on exact screenshots, interface states, and action histories that are easier to inspect, revise, and reuse when retained explicitly.

Embodied control provides a clear example of parametric memory stored in separate, reusable adapters. Rather than merging all experience into the model's shared weights, VLA-Pro \citep{vlapro2026} maintains task-specific LoRA adapters that can be retrieved and combined according to the current situation. VLA-Pro first trains a shared LoRA adapter across previously seen manipulation tasks and then learns a separate task-specific adapter for each task. Each memory entry pairs its adapter with a sequence of structured procedural states describing when the stored behavior is relevant. These states specify the action type, object geometry, end-effector orientation, and target interaction point. Given the current visual observation, language instruction, and interaction history, the system extracts the agent's current procedural state and retrieves the most relevant task memories. It then combines their LoRA adapters using similarity-based weights, temporarily attaches the fused adapter to the vision-language-action model, and uses the adapted policy to generate the next action chunk. VLA-Pro demonstrates that parametric memory need not be incorporated permanently into a single set of shared weights. Parameterized experiences can instead remain modular, indexed by explicit descriptors, and selectively composed according to the current situation. This organization achieves a milestone improvement in cross-task transfer while limiting interference from unrelated manipulation behaviors.

Together, these systems show that parametric memory can operate at different granularities and persistence horizons. MemEIC \citep{memeic2025} accumulates modality-specific knowledge edits, Frames2LoRA \citep{video2lora2026} generates a separate adapter for each video, SE-GA \citep{sega2026} consolidates interaction experience into a shared GUI policy, and VLA-Pro \citep{vlapro2026} stores task-specific action adapters that can be retrieved and composed. Their common property is that past information influences later computation through learned parameters rather than solely through replayed observations or records. Parametric memory can therefore reduce repeated context processing and support direct behavioral adaptation, but it also makes the underlying evidence less accessible and introduces challenges in updating, attribution, interference control, and deletion. Consequently, many multimodal agents combine parametric internalization with explicit memory that preserves source evidence and provides more controllable access.

\subsection{Hybrid Representations}
\label{sec:hybrid_representations}

The preceding subsections describe three primary forms of multimodal memory: source- and modality-specific evidence, textual and structured records, and latent or parametric representations. Each form offers different advantages. Source evidence preserves perceptual detail, textual and structured records support efficient retrieval and explicit reasoning, and latent or parametric representations can contribute directly to model computation. In practice, however, no single form is sufficient for every memory operation. Hybrid memory is not an additional elementary encoding type; rather, it describes how different representations are organized and used together within a memory system. An important subtype is \emph{source-linked memory}, in which an abstract record retains a reference to the original observation or to selected evidence preserved from it. The agent can then search and reason over compact abstractions while recovering the corresponding image, frame, audio segment, or interface region when finer perceptual evidence is required.

For collections of images and visually rich documents, hybrid representations allow an agent to organize large amounts of retrieved evidence without reducing all visual content to text~\citep{wang2026cma}. VimRAG \citep{vimrag2026} represents iterative multimodal retrieval and reasoning as a directed acyclic graph. Each node records its dependencies on earlier nodes, the subquery that produced it, a concise textual summary, and a multimodal memory bank containing visual tokens from retrieved images or video frames. The graph exposes the temporal and logical structure of the reasoning process, while the visual tokens preserve information that may be lost in textual summarization. To operate under a bounded context budget, VimRAG dynamically adjusts the number of visual tokens retained for each memory item according to its semantic relevance, recency, and position in the reasoning graph. The method therefore demonstrates a combined hybrid representation: structured and textual components organize the agent's reasoning state, whereas continuous visual representations preserve perceptual information for later verification and answer generation.

Video streams create a stronger need to connect compact semantic records with recoverable visual evidence because relevant events may occur far apart, and details omitted during summarization may become important only after a query is received. FOLIO \citep{folio2026} combines a short-term buffer of recent frames, a long-term semantic memory, and a visual-evidence cache. As the video arrives, the system constructs structured records describing entities, attributes, locations, relations, actions, and state changes. These records contain timestamps and references to selected keyframes retained in the evidence cache. At query time, FOLIO retrieves relevant semantic records and can restore their associated keyframes when additional visual grounding is needed. The structured memory supports efficient long-horizon access, while the linked frames preserve visual details that the records may omit or describe incorrectly. FOLIO thus exemplifies source-linked hybrid memory: semantic abstraction reduces the cost of storing and searching a long video, but it does not permanently replace the observations from which the memory was derived.

In GUI interaction, source linkage is useful because a textual description may capture the meaning of an earlier interface event while omitting the precise visual target or interface state needed for a later action. MementoGUI \citep{mementogui2026} maintains both within-episode working memory and cross-episode episodic memory. For a selected interaction step, a working-memory entry contains a textual event summary, the bounding box of the relevant interface region, a crop of that region, and a visual embedding used only to organize the memory. When older entries are compressed, their textual content is summarized, but identifiers for selected visual regions are retained and later resolved back into the corresponding crops. Episodic-memory entries similarly combine trajectory summaries and action metadata with representative interface crops and retrieval embeddings. During action generation, the agent receives the retrieved textual summaries and region-of-interest images through the frozen GUI model's standard multimodal input interface. MementoGUI therefore represents interface experience through source-linked textual and visual components: text records task progress and the significance of earlier events, while linked crops recover localized details such as icons, labels, selected values, and interface states. Its retrieval embeddings serve only as indices and do not themselves constitute the memory records to be retrieved and used.

In interactive visual environments, hybrid memory can combine a continuously
revised semantic state with a complete record of visual interaction, since
details that appear unimportant at one step may become useful later.
VISTA~\citep{vista2026} maintains two forms of memory in parallel: revisable
textual notes that summarize reusable environment knowledge and the agent's
current state, and a lossless visual memory that stores every returned frame
in its original form. During reasoning, the agent can rely on the compact
textual state for routine planning while selectively revisiting arbitrary past
frames or spatial regions when finer-grained perceptual evidence is needed.
Unlike source-linked designs in which individual abstract records explicitly
reference particular observations, VISTA keeps its textual state and visual
history separately addressable and combines them on demand.

Embodied agents require hybrid memory to preserve perceptual evidence while organizing it according to the spatial structure of the environment~\citep{cmmrvln2026}. ObsGraph \citep{obsgraph2026} constructs an observation-centered hierarchical scene graph with room, view, and object layers. Room-level representations provide coarse semantic anchors, view-level memories preserve observations in which multiple objects appear together, and object-level memories retain fine-grained visual information about individual objects. Relations among these layers connect object evidence to the views and rooms in which it was observed. Given a task, the agent retrieves memory progressively from rooms to views and objects under a bounded budget. The level at which relevant evidence is missing also determines whether the agent should search another room, refine an existing viewpoint, or explore an unobserved frontier. ObsGraph therefore combines semantic hierarchy, spatial structure, and retained visual observations. The structured graph supports efficient retrieval and planning, while its linked visual evidence preserves appearance and co-visibility information that semantic labels alone cannot express. This combination is particularly important in embodied settings because memory must support not only retrospective question answering but also decisions about where to move and what additional evidence to acquire.
Beyond navigation, HiMe~\citep{hime2026} addresses long-horizon manipulation with a hierarchical hybrid-memory architecture that combines recent visual working memory, image--text contextual records, and a structured procedural plan to retain evolving object states and task progress across action sequences.

Across modalities, hybrid representations exploit the strengths of different memory forms. Textual and structured records provide compact semantic access, explicit relations, and controllable updates, whereas retained observations and modality-specific representations preserve perceptual detail and the provenance of abstracted information. Their combination enables a tiered access strategy: agents can rely on compact abstractions for routine retrieval and reasoning, while selectively recovering higher-fidelity evidence when a task requires verification, reinterpretation, or details lost during abstraction. This capability is likely to become increasingly important as multimodal agents operate over longer histories and in dynamic environments, where retaining all observations in the active context is impractical but irreversible compression is unreliable. Hybrid memory may therefore emerge as a prominent design pattern, particularly for applications in which both efficiency and evidential grounding are important. Its effectiveness, however, depends on more than storing multiple representations. Systems must preserve valid links as memories are revised, merged, or removed; maintain consistency between abstract records and their supporting observations; and learn when compact memory is sufficient and when source evidence should be restored. The central challenge is thus to coordinate representations of different fidelity, cost, and accessibility so that an agent can move between efficient abstraction and grounded perceptual recall as its information needs change.

\section{Axis II: Memory Management}
\label{sec:memory-management}

\begin{figure}[tbp]
    \centering
    \includegraphics[width=0.90\textwidth]
    {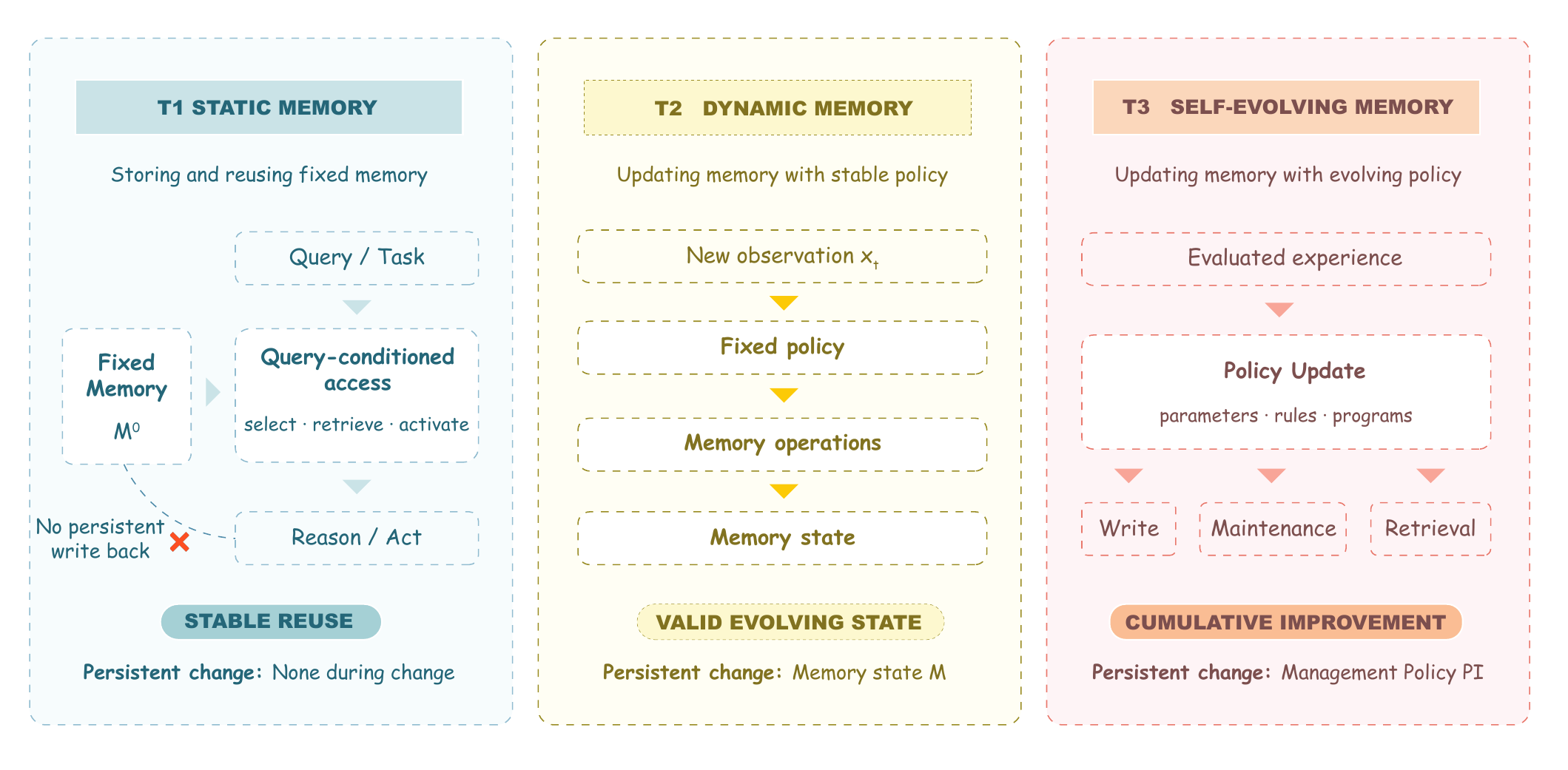}
    \caption{Progression across the three memory-management tiers. T1 Static Memory stores and reuses fixed memory. T2 Dynamic Memory updates memory through a stable policy. T3 Self-Evolving Memory updates memory through an evolving policy. The tiers therefore represent an increasing scope of change, from memory access to memory-state updates and, finally, policy evolution.}
    \label{fig:memory-management-progression}
\end{figure}

After forming memory records from raw messages, experience, and related metadata, the memory system must control how these representations are maintained, accessed, and changed as experience accumulates. \Cref{fig:memory-management-progression} summarizes this progression from T1 Static Memory to T2 Dynamic Memory and T3 Self-Evolving Memory. Specifically, memory management operations include writing, updating, consolidation, reorganization, retention, and forgetting, while retrieval selects relevant memory for ongoing reasoning and action. These operation mechanisms may be governed by heuristic rules~\citep{threedmem2024}, fixed prompts~\citep{graphpad2025}, frozen learned modules~\citep{li2026memento}, optimization procedures~\citep{dreamrunner2024}, or policies that themselves adapt from deployment feedback~\citep{zou2026taskmem}.

Multimodal memory introduces another representational challenge compared to textual inputs. Multimodal inputs are often high-dimensional and expensive to retain in their original form. Therefore, systems must balance storage and retrieval efficiency against the ability to recover source evidence, leading to the diverse representation choices discussed in Axis I. Hybrid systems may preserve multiple representations to balance information fidelity and computational efficiency. Consequently, multimodal memory management involves not only updating memory content but also maintaining consistency across these linked representations. It must determine which representations should change, how updates should propagate among related representations, and whether abstracted memories remain valid and grounded in their source evidence as experience accumulates. Conceptually, we can model these representations as nodes in a representation graph connected by transformation, alignment, fusion, and grounding relations. Memory operations then act on both the representation states and the relations connecting them. We distinguish three tiers according to what can change during inference.

\begin{keybox}{Three Memory-Management Tiers}
\begin{enumerate}[leftmargin=*,itemsep=3pt,topsep=2pt]
    \item \textbf{T1: Static Memory} (\Cref{sec:t1-fixed-memory}): Memory is
      constructed during the initial formation stage and remains fixed
      during subsequent use. The stored memory can be repeatedly accessed
      and reused for downstream reasoning or action.

    \item \textbf{T2: Dynamic Memory}
      (\Cref{sec:t2-fixed-policy}): New observations may update memory
      content, organization, access indices, or metadata through a stable
      management policy. The policy may be defined by hard-coded heuristics
      or learned before deployment, but it does not change during inference.

   \item \textbf{T3: Self-Evolving Memory}
      (\Cref{sec:t3-feedback-evolution}): Both memory and its management
      policy can be updated during deployment. Feedback from interactions
      or task outcomes produces persistent policy changes that guide
      subsequent memory operations.
\end{enumerate}
\end{keybox}

The distinction between memory-state updates and policy evolution depends on the function of the updated component. Memory state includes stored content, organization, access indices, and descriptive metadata, while the management policy governs how memory operations are selected and performed. Changes to retrieval keys, skill descriptions, or usage statistics can alter subsequent retrieval results without changing this policy. Such changes should remain within T2 Dynamic Memory when the selection and update mechanisms remain stable. T3 Self-Evolving Memory goes beyond state modification by using feedback to persistently revise reusable management rules, instructions, procedures, or learned policy components.

Accordingly, the three tiers distinguish whether the memory state and the policy that manages it change during inference. Axis~II, \emph{Memory Management}, organizes systems by the scope of persistent deployment-time change they support. Across these tiers, multimodal information may remain within one representation, be projected into another representation, be fused across multiple modalities, or be jointly maintained across linked representations. The tiers therefore progress from reusing fixed memory, to updating memory through a stable policy, and finally to updating memory through an evolving policy, while the transition patterns characterize how multimodal information moves through the representation system. Updating memory state and evolving management policies can help agents adapt to changing environments and mitigate training–deployment mismatch, but these changes may also introduce instability and make memory updates harder to control. The following subsections examine memory access and reuse in T1 Static Memory, five memory-state operations in T2 Dynamic Memory, and the evolution of writing, maintenance, and retrieval policies in T3 Self-Evolving Memory.

\subsection{T1: Static Memory}
\label{sec:t1-fixed-memory}


\begin{keybox}{Definition of T1: Static Memory}
We define T1 Static Memory as a system in which memory is constructed before downstream reasoning or action and subsequently stored and reused without modification. Its content, organization, access indices, and metadata remain fixed, although each query may retrieve or activate a different subset.
\end{keybox}

T1 Static Memory is well suited to large, static corpora~\citep{visrag2024}, stable knowledge bases \citep{visrag2024, hu2023reveal}, and invariant environment or task information that must be retrieved and reused across subsequent reasoning steps or tasks, such as long-video analysis \citep{fan2024videoagent} or navigation with a preconstructed environment map \citep{mslmaps2025}. The absence of an update policy preserves the memory state across tasks and prevents unverified outputs from modifying it. Representative systems across modalities follow this pattern by constructing memory before downstream task execution and leaving it unchanged during subsequent use~\citep{appagent2023, zhang2025dvd, li2026skill3d}.

Because T1 Static Memory undergoes a single formation cycle, its stored representations and their relationships remain unchanged after formation. Subsequent access and retrieval operations neither modify nor reorganize the persistent memory. Accordingly, the principal design variables in T1 Static Memory systems are the indexing and retrieval mechanisms, which may be optimized during system development or memory formation but remain fixed during inference. Fixed multimodal memories may be organized as flat sequential records~\citep{movieagent2025}, vector-indexed collections~\citep{museagent2026}, hierarchical or multi-tier~\citep{atlasva2026}, relational graphs~\citep{s3mem2026}, inference-time latent states~\citep{comem2025}, or parametric modules~\citep{vlapro2026}. Retrieval-augmented generation provides a common access framework for vector-indexed memory, using dense cross-modal similarity, late interaction, or reranking to retrieve relevant images, document pages, video segments, audio spans, or cross-represented textual records~\citep{visrag2024,pixelrag2026,yeo2026universalrag,fan2024videoagent}. Metadata-aware and spatiotemporal retrieval further exploits timestamps, locations, speaker identities, and temporal boundaries to localize relevant experiences, sometimes through hierarchical coarse-to-fine search \citep{memoryqa2025,someki2026planragaudio}. Relational memories support explicit traversal over connections among objects, states, events, or interface transitions~\citep{shen2025vgent,wen2024autodroid,graphpilot2026}. Planner- or tool-assisted retrieval decomposes a query and routes its subqueries to modality-specific indexes, temporal search tools, or structured databases \citep{fan2024videoagent,someki2026planragaudio}. Inference-time latent-state and parametric memory systems are less common in this setting because their stored information is less explicit and interpretable \citep{video2lora2026}.

In practice, static memory supports three main types of downstream applications: static knowledge bases, reusable evidence or environment stores, and reusable procedural knowledge. First, it can retain external knowledge or documentation for domain-specific applications, such as medicine, thereby improving grounding and reducing hallucinations \citep{xia2025mmedrag}. Second, it can store expensive-to-process but largely static content, such as long videos~\citep{fan2024videoagent} or GUI environments~\citep{wen2024autodroid}, through one-time ingestion and reuse this information across subsequent tasks \citep{fan2024videoagent,wen2024autodroid}. Finally, static memory can preserve procedural knowledge for repeated action planning and execution \citep{vlapro2026}.

T1 Static Memory is most effective when the underlying information changes slowly and the stored representation will be reused across tasks. Expensive operations such as document encoding, video analysis, environment exploration, and adapter training are performed during memory formation. Because the stored content remains unchanged, each task accesses the same evidence, environment model, or procedural library. This stability supports consistent behavior.

However, this stability limits the system's ability to respond to change. T1 Static Memory preserves both the useful information and any omissions or errors introduced during its formation. During normal task execution, it cannot incorporate corrections to stored evidence, changes in an interface or physical environment, or procedural knowledge gained from later successes and failures. The stored representation may therefore become incomplete or outdated as the source, environment, or task distribution changes.

\subsection{T2: Dynamic Memory}
\label{sec:t2-fixed-policy}


\begin{keybox}{Definition of T2: Dynamic Memory}
T2 Dynamic Memory updates memory during continued interaction through a stable management policy. New experiences can modify stored representations, metadata, organization, or access indices through operations such as updating, consolidation, revision, reorganization, and forgetting. The policy governing these operations may be implemented through hard-coded heuristics or learned before deployment, but it remains stable during inference. A stable policy may select different operations for different observations or memory states; its governing rules and policy parameters do not persistently evolve from deployment feedback during the phase classified as T2.
\end{keybox}

We organize T2 Dynamic Memory methods around five fundamental memory operations.

\begin{keybox}{Five Fundamental Memory Operations}
\begin{enumerate}[leftmargin=*,itemsep=2pt,topsep=2pt]
    \item \textbf{Update} (\Cref{sec:t2-memory-update}): adds new observations
          or integrates them into existing records.
    \item \textbf{Consolidation} (\Cref{sec:t2-memory-consolidation}): merges
          related records or abstracts repeated experiences into compact
          knowledge.
    \item \textbf{Revision} (\Cref{sec:t2-memory-revision}): corrects or
          replaces stored information that has become inconsistent or outdated.
    \item \textbf{Reorganization} (\Cref{sec:t2-memory-reorganization}): changes
          how records are grouped, linked, indexed, or transferred across memory
          levels.
    \item \textbf{Forgetting} (\Cref{sec:t2-memory-forgetting}): removes records
          or reduces their retention or retrieval priority.
\end{enumerate}
\end{keybox}

In multimodal memory systems, the central design challenge is to perform these operations across heterogeneous representations while preserving relevant information and cross-modal relationships. Therefore, to systematically examine T2 Dynamic Memory methods, we characterize each fundamental memory operation along two multimodal-specific dimensions. The representation-handling pattern indicates whether an operation acts locally within a single representation, coordinates distinct representations while keeping them separable, or transforms information across representations through projection or fusion. The computational mechanism specifies how the resulting memory change is produced: through hard-coded heuristics, a prompted LLM, a trained functional update module, or iterative optimization under a fixed objective.

\subsubsection{Memory Update}
\label{sec:t2-memory-update}

Memory update incorporates new observations, interactions, or feedback into persistent memory. Depending on the target representation, an update may append new evidence, revise an existing record, modify a structured state, integrate information into a latent state, or edit specific model parameters. The central multimodal challenge is to incorporate new information while preserving existing content in abstract representations, cross-modal alignment, and relationships among linked representations. 

Source-preserving memories incorporate new information by inserting observations, replacing outdated evidence, or modifying associated metadata. Source-preserving memories commonly incorporate new information by appending raw or minimally processed multimodal observations to an evidence log or feature buffer, thereby retaining prior perceptual evidence without rewriting it \citep{diko2025rewind,feng2026m2a}. The appended evidence may include metadata, source links, or derived retrieval indices that support efficient access while preserving the original observations \citep{feng2026m2a}.

Modality-specific feature representations are generally updated through within-modality matching followed by feature aggregation or insertion. Each incoming feature is compared with stored representations from the same modality. If a match is identified, the corresponding memory entry is augmented, averaged, or replaced; otherwise, a new entry is created. This mechanism preserves modality-specific evidence while using shared entity identifiers to coordinate related information across modalities \citep{long2025m3agent}.

Geometric memory representations are updated by first registering new observations within a common spatial reference frame and then associating the projected measurements or detected instances with existing spatial elements. Matched elements may be updated through feature fusion or attribute modification, whereas unmatched observations may create new map cells, objects, nodes, or relations. Invalid or obsolete elements may likewise be removed~\citep{liu2025dynamem,jiang2024roboexp,mitash2020taskdriven}. These predefined association and fusion procedures provide explicit control over spatial updates and help preserve geometric consistency, but their reliability depends on fixed matching criteria and their robustness to noisy, incomplete, or conflicting observations.

Structured memory representations organize information as entities, relations, events, or procedures. New information is incorporated by identifying the relevant structural unit and then modifying its attributes, inserting or removing nodes and edges, adjusting relation weights, or updating transition statistics. Target units may be selected through similarity-based gating, while their indices and relations are maintained through moving-average aggregation, count updates, and graph-editing operators \citep{jiang2026nsmem}. 

Textual semantic records derived from multimodal experience retain perceptual or interaction information as natural-language facts, attributes, or event descriptions. Unlike structured memories, whose updates modify explicit entities, relations, or topology, these records are maintained through content-level semantic edits. Updating them requires interpreting incoming evidence relative to existing content and selecting an operation such as addition, modification, confirmation, deletion, or no change. Prompted LLMs or VLMs are used to perform this comparison because the system must determine whether new evidence confirms, extends, contradicts, or supersedes an existing record from the multimodal
experience~\citep{feng2026m2a,hong2026tame,ma2026reflectworld,hime2026}. 

Latent memory incorporates new information through trained parameterized modules that either transform an existing latent state or construct new latent memory entries. Recurrent state-transition modules integrate the preceding latent state with the current multimodal representation to produce a successor state \citep{qian2024videostreaming}. Query-based write modules instead distill incoming features, instructions, and retrieved context into compact tokens that are added to a latent memory bank \citep{diko2025rewind}. Thus, the former modifies an existing latent representation, whereas the latter expands the memory with newly constructed latent entries. Because both forms distribute information across latent vectors, their updates are generally difficult to inspect, attribute to specific observations, or correct selectively.

Parametric memory incorporates new information by routing it to a designated subset of model parameters and optimizing those parameters under a fixed editing objective~\citep{memverse2025}. Modality-specific adapters can restrict visual and textual edits to separate parameter subsets, limiting interference across modalities and unrelated model behavior \citep{memeic2025}. Unlike a trained functional module, which computes the next memory state through a forward pass, a parametric editor produces the updated memory through inference-time optimization. The central challenge is to internalize new information while preserving previously stored knowledge and maintaining the model's behavior on unrelated inputs.

Hybrid memory systems coordinate updates across multiple representational forms, such as source evidence, semantic records, latent tokens, and model parameters. Common mechanisms include linking editable semantic records to immutable source evidence~\citep{feng2026m2a}, maintaining compact latent representations alongside detailed perceptual features~\citep{diko2025rewind}, and coupling parametric edits with corresponding external records~\citep{memeic2025}. These coordinated pathways preserve the complementary properties and provenance of different representations. However, they must keep linked memory components consistent as new information is incorporated, without collapsing representational distinctions or breaking associations.

\subsubsection{Memory Consolidation}
\label{sec:t2-memory-consolidation}

Memory consolidation transforms related stored information into a less redundant, more compact representation. Unlike update, which incorporates an incoming experience, consolidation operates across accumulated memories. Its central multimodal challenge is to reduce storage and retrieval costs without removing perceptual details, temporal distinctions, modality-specific evidence, or cross-representation links that may remain important for later reasoning. Existing mechanisms primarily perform feature-space compression, semantic abstraction, structured merging, or coordinated consolidation across representations~\citep{lian2026mmmem}.

Feature-space compression reduces redundancy among stored perceptual or latent encodings. Temporally adjacent or semantically similar visual features can be selected through fixed similarity criteria and then averaged~\citep{malmm2024}, clustered~\citep{flashvstream2025}, or merged~\citep{moviechatplus2026} into a smaller set of representative tokens. More adaptive mechanisms vary the degree of compression according to temporal change or semantic salience, preserving more features around distinctive events while compressing repetitive observations more aggressively~\citep{song2024moviechat,folio2026,videoem2025}.
FluxMem~\citep{xie2026fluxmemadaptivehierarchicalmemory}, for example, achieves competitive performance by retaining recent observations at full resolution, removing temporally redundant tokens from mid-term memory, and further consolidating spatially repetitive regions in long-term memory.
These mechanisms bound memory growth, but overly aggressive compression may remove brief actions, fine visual details, or temporal transitions that cannot be recovered from the consolidated features.

Semantic abstraction can likewise consolidate a growing video history in language space. ReMA~\citep{chen2026mmlifelong} first appends timestamp-aligned descriptions for successive video clips. Whenever a new passive or query-directed observation overlaps existing temporal nodes, a prompted summarizer merges the old and new content and replaces those nodes with a unified record; non-overlapping observations are added as new entries. The same procedure operates during initial stream ingestion and recursive query answering, so the belief state continues to change while the memory-management policy remains stable. ReMA therefore exemplifies consolidation in T2 Dynamic Memory rather than policy evolution in T3 Self-Evolving Memory: task-specific inspection can refine what is stored, but task outcomes do not update the rule governing later memory operations.

Structured merging consolidates memories whose internal organization must be preserved. Graph alignment and matching procedures identify equivalent entities, actions, or procedural steps before merging their nodes, edges, alternative paths, and associated statistics~\citep{worldmm2026}. Recurring action sequences may also be extracted from multiple episodes and converted into reusable procedural structures \citep{jiang2026nsmem}. Unlike feature averaging, structured merging must determine which elements are genuinely equivalent and which represent meaningful alternatives; incorrect alignment can collapse distinct entities, actions, or execution paths.

Coordinated cross-representation consolidation creates a compact representation while retaining links to complementary forms of evidence. A visual sequence may be compressed into representative features and paired with a textual clue~\citep{xiong2025streamchat}, an audiovisual episode may be summarized while preserving timestamps and representative frames~\citep{pyravid2026}, or an entity-level fact may remain linked to the observations from which it was derived~\citep{ma2026reflectworld}. This coordination allows the compact memory to support efficient retrieval while the linked source or lower-level representation preserves details omitted during abstraction. The central requirement is therefore not to preserve every representation at equal resolution, but to maintain sufficient links for the consolidated memory to be interpreted, verified, or expanded when needed.

\subsubsection{Memory Revision}
\label{sec:t2-memory-revision}

Memory revision corrects an existing memory when new evidence indicates that its content is inaccurate, inconsistent, incomplete, or outdated. The revision operation is narrower than update: adding an independent observation constitutes an update, whereas modifying or replacing an existing record in response to conflicting or superseding evidence constitutes revision. It also differs from forgetting, which removes information because of capacity, recency, or importance rather than because new experiences show the information to be incorrect. In multimodal systems, revision primarily involves spatial and object-state correction, semantic revision, and conflict reconciliation across linked representations.

Spatial and object-state correction revises geometric or structured environment memory when new observations indicate that an object has moved, disappeared, changed state, or been incorrectly associated with an existing entity. The mechanism registers each new observation within a common spatial reference frame and identifies the affected object or map region through geometric and semantic matching. It then replaces outdated attributes, removes invalid entities, updates object geometry, or recomputes spatial relations within the affected subgraph \citep{jiang2024roboexp,dovsg2025}. Restricting revision to the affected region avoids reconstructing the entire environment representation, but reliable correction requires maintaining object identity despite occlusion, viewpoint changes, and noisy observations.

Semantic revision operates on explicit facts, attributes, and event records derived from multimodal experience. Incoming evidence is compared with existing records to determine whether it confirms, extends, contradicts, or supersedes their content. Prompted LLMs or VLMs commonly perform this comparison and select among fixed operations such as modification, replacement, deletion, confirmation, or no change \citep{feng2026m2a,hong2026tame,ma2026reflectworld,worldmm2026}. Effective revision must correct outdated information without replacing compatible records merely because the new observation is incomplete or expressed differently.

Cross-representation conflict reconciliation is required when the same experience is retained as source evidence and as one or more derived semantic or structured records. Evidence identifiers or entity associations can be used to retrieve the observations supporting a disputed record before the record is revised. The correction is then applied to the derived representation while its links to the supporting visual, auditory, or interaction evidence are retained \citep{feng2026m2a,ma2026reflectworld}. This mechanism preserves provenance and supports later verification, but it must determine whether an apparent conflict reflects a genuine change, an unreliable observation, or information that is valid only under a different temporal or situational context.

\subsubsection{Memory Reorganization}
\label{sec:t2-memory-reorganization}

Memory reorganization changes the structural role, placement, or access path of information that remains in memory. It does not primarily add observations, correct content, or combine multiple memories. Instead, it determines where retained information resides and how it is connected or retrieved. Multimodal memory systems primarily perform reorganization through tier migration, graph relinking, and cross-representation reindexing or association.

Tier migration transfers retained memory records between storages with different temporal or functional roles, such as short- and long-term memory or dynamic and static memory. A fixed trigger first identifies information that meets a capacity, persistence, or stability criterion, then moves or copies it to the appropriate destination. Compact visual representations can be transferred from a bounded short-term buffer into long-term memory~\citep{song2024moviechat}, while attributes judged to be persistent can migrate from dynamic conversational state into static personalized memory~\citep{hong2026tame}. Here, tiers denote temporal or functional memory roles rather than hardware storage levels. Moving an unchanged artifact between accelerator memory, host memory, and storage is a physical residency decision, discussed in \Cref{subsec:scope-abstractions,subsec:storage-residency}; it does not by itself reorganize the logical memory. 

Graph relinking changes the relations through which retained entities or events are structurally connected. When an embodied scene changes, links involving affected objects can be invalidated and spatial relations within the local subgraph recomputed without rebuilding the complete scene representation \citep{dovsg2025}. Relinking can also establish or reinforce associations between separately retained modality-specific evidence, such as face and voice representations that are determined to refer to the same persistent entity \citep{long2025m3agent}. These mechanisms support entity continuity and cross-modal traversal, but incorrect associations can connect unrelated evidence or propagate a local error through the graph.

Cross-representation reindexing changes the retrieval representation associated with retained content without necessarily modifying the underlying content or the retrieval policy. Incoming observations can be compared with the goal- and step-level index vectors associated with an existing procedural graph, after which the matched vectors are adjusted through a fixed moving-average rule \citep{jiang2026nsmem}. The procedure itself remains unchanged, but its retrieval keys can be updated so that varied descriptions of the same task can reach a shared procedural memory. More generally, reindexing and association maintenance keep dense retrieval keys, modality-specific identifiers, and structured memory items aligned as experience accumulates~\citep{hymem2026,pyravid2026}. Excessive index drift or incorrect identity association, however, may make earlier access paths unreliable.

\subsubsection{Memory Forgetting}
\label{sec:t2-memory-forgetting}

Memory forgetting controls which accumulated information should remain available to the agent and at what level of detail. The forgetting operation may remove an item from an active store, reduce the detail preserved in one representation, or lower a memory's priority during retrieval. In multimodal systems, forgetting need not affect every representation of an experience equally: detailed perceptual evidence may leave active memory while a compact event record, textual summary, or linked long-term representation remains available. The major mechanisms in the current literatures are active evidence eviction, representation-detail decay, semantic accessibility decay, and coordinated selective forgetting across linked representations.

Active evidence eviction removes source-preserving evidence or their encodings from a bounded active memory. Selection may follow insertion order such as first-in, first-out (FIFO), event-level reservoir sampling, object identity, frequency and recency statistics, or recency-weighted retention probabilities~\citep{echovla2025,wen2026eventmemagent,wang2024karma,xiong2025streamchat}. Eviction bounds active-memory cost, and removal from the active buffer does not imply that all information derived from the observation has been erased from long-term memory. They can be reorganized or migrated to another memory store before being forgotten.

Representation-detail decay retains a memory in its source modality while progressively reducing the detail or size of its representation. Interaction histories rendered as memory images can be downsampled as they recede in time, therefore reducing their resolution and visual-token allocation while a more detailed textual record remains available \citep{li2026ocrmemory}. Optical-forgetting mechanisms similarly apply progressive resolution reduction to visually represented memory pages, preserving coarse semantic structure while discarding fine visual details \citep{scrapmem2026}. 

Semantic accessibility decay preserves the content of an episodic or semantic record while lowering its retrieval priority. A time-dependent decay factor can reduce a record's retrieval priority over time, while repeated recall or contextual relevance may restore its priority~\citep{memverse2025}. When retrieval priority decreases, the memory remains stored but becomes less likely to enter the agent's working context. This mechanism provides more gradual control than eviction, although an overly aggressive decay function may render stored experiences effectively inaccessible.

Coordinated selective forgetting operates on experiences stored in multiple representations and determines which representation should be removed, compressed, or deprioritized while other representations are retained. The goal is to exploit the complementary strengths of different representations while reducing redundant storage. Detailed visual information may be selectively discarded or compressed while a more compact long-term representation is preserved \citep{xiong2025streamchat,li2026ocrmemory,wang2024karma}. 

\subsection{T3: Self-Evolving Memory}
\label{sec:t3-feedback-evolution}


\begin{keybox}{Definition of T3: Self-Evolving Memory}
T3 Self-Evolving Memory uses feedback from continued interaction to update the policies that govern memory operations. Let $M_t$ denote the stored memory and $\pi_t$ the policy controlling its construction, maintenance, or retrieval. T2 Dynamic Memory may change $M_t$ while keeping $\pi_t$ stable, whereas T3 Self-Evolving Memory also updates the policy:
\[
M_{t+1}=U_{\pi_t}(M_t,x_t),
\qquad
\pi_{t+1}=E(\pi_t,f_t),
\]
where $x_t$ is the current experience and $f_t$ is a task outcome, verifier judgment, or user correction. The revised policy persists and influences memory operations in subsequent interactions or tasks.
\end{keybox}

T3 Self-Evolving Memory represents a memory-centered form of inference-time learning in which experience accumulated during inference drives improvements in memory management policies. Memory therefore supports inference-time learning at two levels: accumulated experience supports the agent's inference-time learning, while the memory system itself can also evolve through feedback-driven updates to its management policies.

The evolving policy may be realized through learned parameters~\citep{zou2026taskmem}, symbolic rules~\citep{barmann2026h2emv}, utility-based decision criteria~\citep{sun2026skemex}, or explicit procedures such as reusable retrieval strategies~\citep{fan2026rrm} and executable programs~\citep{elhattami2026skillnb}.

For multimodal memory, policy evolution must account for the different roles of heterogeneous representations. A write policy determines which source observations or features should be retained and how they should be transformed into persistent records. A maintenance policy determines which representations and levels of detail should remain available as memory accumulates. A retrieval policy determines which representation or store should be searched and how evidence across representations should be coordinated. We therefore organize T3 Self-Evolving Memory methods into writing-policy, maintenance-policy, and retrieval-policy evolution.

\subsubsection{Write-Policy Evolution}
\label{sec:t3-write-policy-evolution}

Write-policy evolution uses feedback to improve how raw experiences are selected and transformed into memory. Unlike a fixed write mechanism, which repeatedly applies the same admission and representation-transformation procedures, an evolving write policy adapts what information is retained and how that information is represented in response to feedback obtained during deployment.

The selection policy determines which parts of raw experience should be retained as memory records. Feedback from new interactions or completed tasks is used to adapt the policy toward retaining the most relevant parts of raw experience that are likely to support subsequent reasoning \citep{zou2026taskmem}. However, this adaptation introduces a key trade-off: when the task or environment changes substantially during deployment, a policy optimized for prior experience may discard important information in the new setting.

The transformation policy changes how selected multimodal evidence is converted into the target memory representation. This evolution currently operates through three main mechanisms. Representational abstraction determines the semantic granularity and level of detail preserved when raw experience is converted into an explicit record. Normalization and grounding improve temporal alignment, entity resolution, and contextual grounding across modality-specific observations before they are committed to memory. Structural construction changes how grounded evidence is organized into entities, events, attributes, relations, or graph structures. In source-linked memory, it may also adapt how derived semantic or structured records are connected to their supporting visual observations, interaction traces, timestamps, and entity identifiers. Failure traces can produce validated policy changes that repair graph construction and source-to-record mappings while preserving provenance and consistency across linked representations \citep{tian2026abot}. Across these mechanisms, the central tension is to improve task relevance and representational efficiency without losing source-grounded details, temporal consistency, or cross-modal relationships needed by future tasks.

\subsubsection{Maintenance-Policy Evolution}
\label{sec:t3-maintenance-policy-evolution}

Maintenance-policy evolution uses feedback to improve how existing memory representations are retained, revised, and removed over time. The maintenance policy determines which stored representations persist and how much information is preserved during operations such as consolidation, reorganization, and forgetting. Unlike a fixed maintenance policy, an evolving policy adapts these decisions based on user feedback or accumulated evidence of downstream utility. Two main mechanisms emerge for maintenance-policy evolution.

Feedback-driven retention uses user feedback to revise which memory representations should remain available over time. When a failed memory query reveals that relevant information was previously removed, the feedback updates the retention policy so that similar information encountered in the future is retained longer \citep{barmann2026h2emv}. This mechanism improves future retention decisions but cannot recover information already removed unless it remains available through another linked representation. Feedback-driven representation abstraction uses feedback to determine the information to be preserved when a memory representation is being compressed. Updated relevance criteria guide subsequent grouping and summarization toward preserving information that has proved useful to the user or downstream tasks. 

\subsubsection{Retrieval-Policy Evolution}
\label{sec:t3-retrieval-policy-evolution}

Retrieval-policy evolution uses feedback to improve the reusable strategies that govern memory access. Specifically, the retrieval policy determines how queries are formulated and how candidate memories are prioritized for subsequent reasoning. Existing methods primarily focus on two policy targets: search formulation and control, and memory selection and prioritization.

Search formulation and control adapt how queries are constructed and how incomplete retrieval is repaired. Successful and failed retrieval trajectories can be distilled into reusable guidance describing evidence requirements, query patterns, and corrective searches. In long-video reasoning, this guidance informs subsequent query construction over the current memory graph, while answer generation remains grounded in evidence retrieved from the current video \citep{fan2026rrm}. 

After searching, memory selection and prioritization determine which candidate records receive a higher probability of selection by the retrieval. In embodied memory, \citep{tian2026abot} utilizes validated rules to prioritize object records supported by direct observations, using location, recency, and links to source evidence to guide retrieval in later tasks. \citep{sun2026skemex} updates memory utility estimates from downstream feedback in clinical settings.

The evolution of memory-management policies benefits from feedback across the memory lifecycle, as errors introduced at one stage can affect later stages. Information omitted during writing or lost during maintenance cannot be recovered through improvements to the retrieval policy. A failed memory retrieval may therefore reflect earlier information loss, ineffective retrieval, or incorrect use of retrieved evidence. Distinguishing the causes of these failures motivates evaluating writing, maintenance, and retrieval jointly and targeting policy updates to the mechanisms that contribute to failure.

\section{Cross-Axis Analysis}
\label{sec:cross-axis-analysis}


\definecolor{TOneTableColor}{RGB}{202,222,229}
\definecolor{TTwoTableColor}{RGB}{253,248,219}
\definecolor{TThreeTableColor}{RGB}{246,221,199}

\begin{table*}[!t]
    \centering
    \footnotesize
    \setlength{\tabcolsep}{4pt}
    \renewcommand{\arraystretch}{1.18}
    \caption {Representational bottlenecks across memory-management tiers: T1 concerns memory formation and access, T2 adds controlled memory-state change,
   and T3 adds policy evaluation and evolution.}
    \label{tab:representation-bottlenecks}
    \begin{tabularx}{\textwidth}{
        @{}
        >{\raggedright\arraybackslash}p{0.13\textwidth}
        >{\raggedright\arraybackslash}p{0.17\textwidth}
        >{\raggedright\arraybackslash}X
        >{\raggedright\arraybackslash}X
        @{}
    }
        \toprule
        \makecell[c]{\textbf{Tier}} &
        \makecell[c]{\shortstack[c]{\textbf{Scope of}\\\textbf{persistent change}}} &
        \makecell[c]{\textbf{Core requirement}} &
        \makecell[c]{\textbf{Key representational bottlenecks}} \\
        \midrule

        \rowcolor{TOneTableColor!35}
        \textbf{T1 Static Memory} &
        None after formation &
        Preserve task-relevant information in a form that can be accessed and
        reused. &
        Failure to preserve task-relevant information during formation;
        ineffective access to retained information. \\

        \addlinespace

        \rowcolor{TTwoTableColor!35}
        \textbf{T2 Dynamic Memory} &
        Memory state &
        Apply a controlled change to the correct target while preserving
        unrelated content and dependent representations. &
        Limited editability or excessive operation scope; insufficient traces
        for verifiability; limited reversibility; incomplete update propagation
        and cross-representation inconsistency. \\

        \addlinespace

        \rowcolor{TThreeTableColor!35}
        \textbf{T3 Self-Evolving Memory} &
        Memory state and management policy &
        Attribute outcomes to policy decisions, revise the responsible policy,
        and validate the revision across subsequent use. &
        Limited policy addressability; insufficient policy-decision traceability
        and feedback attribution; nonpersistent policy revisions; insufficient
        policy-level validation. \\

        \bottomrule
    \end{tabularx}
\end{table*}

Axis I and Axis II describe different aspects of multimodal agent memory, but they interact in system design. Memory representation constrains how reliably the operations required at each management tier can be performed. T1 Static Memory requires fixed memory to be located, accessed, and reused in downstream reasoning or action. T2 Dynamic Memory further requires the correct memory units to be updated through a stable policy, verified, and synchronized across linked representations without affecting unrelated content. T3 Self-Evolving Memory adds a stronger requirement: feedback must be attributed to a memory-management decision, and the relevant writing, maintenance, or retrieval policy must evolve safely and persistently. 

As the scope of persistent change expands from memory state to management policy, each tier inherits the representational bottlenecks of the preceding tier and adds another level of uncertainty. As shown in \Cref{tab:representation-bottlenecks}, T1 depends on preserving and accessing task-relevant
information. T2 additionally depends on correctly changing and coordinating the memory state. T3 further depends on evaluating these management decisions and attributing feedback to a reusable policy component.

We therefore examine the representation requirements for T1 Static Memory, T2 Dynamic Memory, and T3 Self-Evolving Memory, followed by corpus-level patterns across the two axes.

\subsection{Representation Requirements for Static Memory}
\label{representation_for_static_memory}

T1 Static Memory imposes the simplest requirement on memory representation. Once memory has been formed, the system only needs to find and use it for downstream reasoning or action. The main representational requirement for T1 Static Memory systems is that memory formation preserves the information needed for later tasks in a form that can be effectively accessed and reused. Effective access mechanisms and retrieval indices must identify the stored context most relevant to the current query and make it available to the agent \citep{visrag2024, fan2024videoagent, yeo2026universalrag}. T1 Static Memory requires no persistent memory change: the fixed memory only needs to be accessed and reused.

\subsection{Representation Requirements for Dynamic Memory}
\label{representation_for_dynamic_memory}

T2 Dynamic Memory updates memory through a stable management policy. The memory state changes during use, while the policy that controls these changes remains stable. Finding and using memory is no longer sufficient. The system must change the correct memory unit, preserve unrelated information, and keep dependent representations consistent.
We unfold our discussion into the properties below:

\paragraph{Editability}
We use editability to describe whether a target memory unit can be directly modified, replaced, or removed, and at what granularity the change can be applied. We use operation scope to describe all memory units, relations, indices, and behaviors affected by the operation, including both intended and unintended effects. A controlled operation should have a minimal sufficient scope: it should change the target unit and its necessary dependencies while preserving unrelated memory.

The available editing scope depends on the representation, which we categorize them as source-preserving memory and cross-represented explicit memory. Source-preserving memory allows the system to append, replace, or remove individual observations. However, a semantic error may lie in an interpretation derived from the source evidence rather than in the retained observation itself. If an agent assigns the wrong identity to a person in a video, the stored frames still record the visual input available to the agent and need not be modified. The error lies in the annotation, entity association, or derived record. Correcting it therefore requires changing the derived representation rather than the source evidence. Cross-represented explicit memory exposes facts, events, entities, attributes, and relations as separate units. These units can be changed directly~\citep{feng2026m2a, worldmm2026}. However, an explicit memory unit may have several dependencies. Changing an entity identity may also require changes to its attributes, event records, summaries, relations, and retrieval indices. An operation that is too narrow leaves these dependent records outdated. An operation that is too broad may change unrelated information. Latent and parametric memory is harder to edit with a controlled scope. Information may be distributed across latent states, model weights, or adapters. A state transition or optimization step can change the representation, but the system may not know which numerical components encode the target information. The update may therefore affect unrelated knowledge or behavior.

\paragraph{Verifiability}
Verifiability asks whether the result of a memory operation can be checked for correctness.
It requires a trace that connects the previous memory state, the applied operation, the supporting evidence, and the resulting state. The representation used for verification may differ from the representation being changed. A revised textual event record may need to be checked against a video segment, GUI state, audio interval, or embodied interaction trace. Explicit and source-linked memory can support this comparison through evidence identifiers and provenance links. For latent or parametric memory, direct record-level verification is usually unavailable. The system must instead use behavioral tests to check whether the intended behavior changed while unrelated behavior remained stable~\citep{memeic2025}.

\paragraph{Reversibility}
Reversibility asks whether the memory state preceding that operation can be restored. Reversal is needed when later evidence, verification, or downstream outcomes reveal that a memory operation introduced an incorrect or harmful change, such as overwriting a valid record after an erroneous entity match. Preserving the original observation alone may not be sufficient for reversing a memory operation. Regenerating a summary or structured record from that observation may not reproduce the previous state when the formation process uses a stochastic model such as an LLM. Reliable reversal therefore requires a previous version, an operation log, or an inverse operation. Explicit memory can support item-level or version-level restoration. Latent and parametric memory often relies on model checkpoints, which may also remove valid updates made after the operation being reversed.

\paragraph{Update propagation}
Update propagation describes how the effects of a memory operation are carried from one representation to other dependent representations. This issue arises in hybrid and source-linked memory, where the same information may be stored in several forms. An operation applied to one representation may therefore require corresponding changes to the others. Cross-representation consistency describes whether these dependent representations remain aligned after the operation.

Propagation must be both complete and selective. Correcting a person's identity may require changes to textual records, entity links, summaries, and retrieval indices, while the original image remains unchanged as source evidence. A new observation of an object's location may update its current-state record and retrieval index while preserving earlier observations as historical evidence. Under-propagation leaves outdated or conflicting records. Over-propagation may remove valid evidence or change unrelated information. Reliable propagation, therefore, requires links between representations that refer to the same experience. These links may use timestamps, source identifiers, entity identifiers, version identifiers, or provenance relations. 

The representation requirements for T2 Dynamic Memory are therefore stronger than those for T1 Static Memory. Source-preserving memory retains evidence but often lacks direct targets for semantic correction. Cross-represented explicit memory supports precise edits but requires provenance, versioning, and dependency tracking. Latent and parametric memory can be updated, but its update scope is difficult to identify and verify. Hybrid and source-linked memory supports both explicit correction and source-based verification, but it must keep its representations synchronized. 

\subsection{Representation Requirements for Self-Evolving Memory}

\label{sec:cross-representation-consistency}

The representation requirements discussed for T2 Dynamic Memory remain relevant
to T3 Self-Evolving Memory. T3 introduces an additional meta-level requirement: the system must
represent its memory-management policies, the decisions produced by that policy,
and the feedback used to evaluate and revise it. Self-evolution therefore
depends on policy addressability, policy-level feedback attribution, and persistent, validated policy revision.

\paragraph{Policy Addressability}
For a memory-management policy to evolve, the part responsible for a recurring
behavior must provide an identifiable target for revision. It may be represented explicitly as a rule, instruction,
workflow, program, or retrieval specification, or implicitly in a learned module
or parameter set. Explicit policy representations allow more precise revisions, whereas distributed representations make the effects of individual changes harder to isolate.

An explicit memory representation and an explicit policy representation serve
different purposes. The former exposes the content being managed, whereas the
latter exposes the mechanism that determines how this content is formed,
maintained, and accessed. Identifiable memory items can make a state change
easier to inspect, but they do not by themselves identify which part of the
management policy should be revised.

\paragraph{Policy Evaluation and Feedback Attribution}
T3 must evaluate whether outcomes reveal a recurring weakness in the
memory-management policy rather than an isolated error in the memory state. This
requires connecting an outcome to the memory content involved, the management
decisions that shaped its availability and use, and the policy component and
version that produced those decisions. A policy-evaluation record may therefore
include the relevant task context, affected memory items, operations performed,
retrieved content, policy version, and subsequent feedback.

This requirement extends the operation-level traceability needed in T2. In T2,
a trace supports verification or reversal of a particular memory-state change.
In T3, traces across interactions support policy-level attribution: determining
whether repeated outcomes justify changing the reusable mechanism that selects
such operations. Feedback must therefore be sufficiently informative to
distinguish failures of writing, maintenance, retrieval, reasoning, and action.
A task-level reward alone may reveal that the agent failed without identifying
which part of memory management should change.

Axis~I constrains what evidence is available for this evaluation. Explicit
memory units, provenance, operation histories, and source links permit more
direct inspection of management decisions. Distributed latent or parametric
representations expose fewer identifiable units and therefore place greater
weight on behavioral evaluation across tasks. Thus, the memory
representation affects how directly the memory-management design can be evaluated.

\paragraph{Persistent and Validated Policy Revision}
A policy revision constitutes self-evolution only when it is retained as part
of the reusable memory-management process and affects subsequent memory
operations. The system must therefore represent the revised policy in a form
that can be stored, reapplied, and associated with later decisions. Maintaining
policy versions and their evaluation histories also supports comparison,
auditing, and restoration when a revision proves harmful.

Validation must assess the revised policy rather than only the memory-state
change that motivated it. Correcting one record demonstrates that a particular
error has been repaired, but it does not establish that the revised policy will manage future memories more effectively. Policy-level validation therefore requires subsequent, replayed, or held-out cases that test both the targeted behavior and possible regressions in other memory functions.

T3 Self-Evolving Memory requires a meta-level representation that closes
the loop from policy decisions to outcomes, from outcomes to evaluation, and
from evaluation to persistent and validated policy change.

\subsection{Corpus-Level Patterns Across the Two Axes}
\label{sec:cross-axis-corpus}

\begin{figure}[H]
\centering

\includegraphics[width=0.55\textwidth]{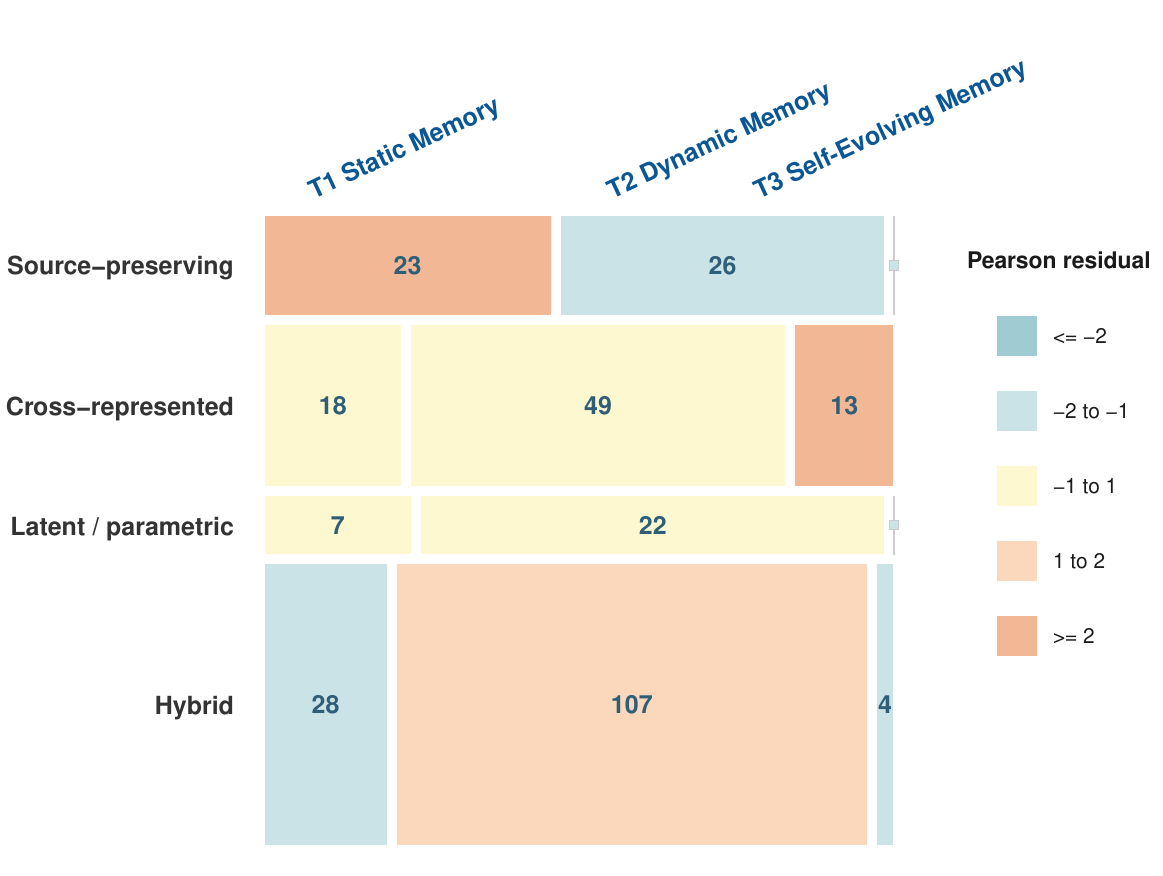}

\caption{Mosaic plot of representation types across management tiers for
\ArchitectureCount{} multimodal memory architectures. Rectangle area and
labels show counts. Colours encode Pearson residuals: blue indicates fewer
architectures than expected, yellow indicates near expectation, and peach
indicates more than expected.}

\label{fig:cross-matrix}
\end{figure}

\begin{figure}[!htb]
    \centering
    \includegraphics[width=0.78\textwidth]{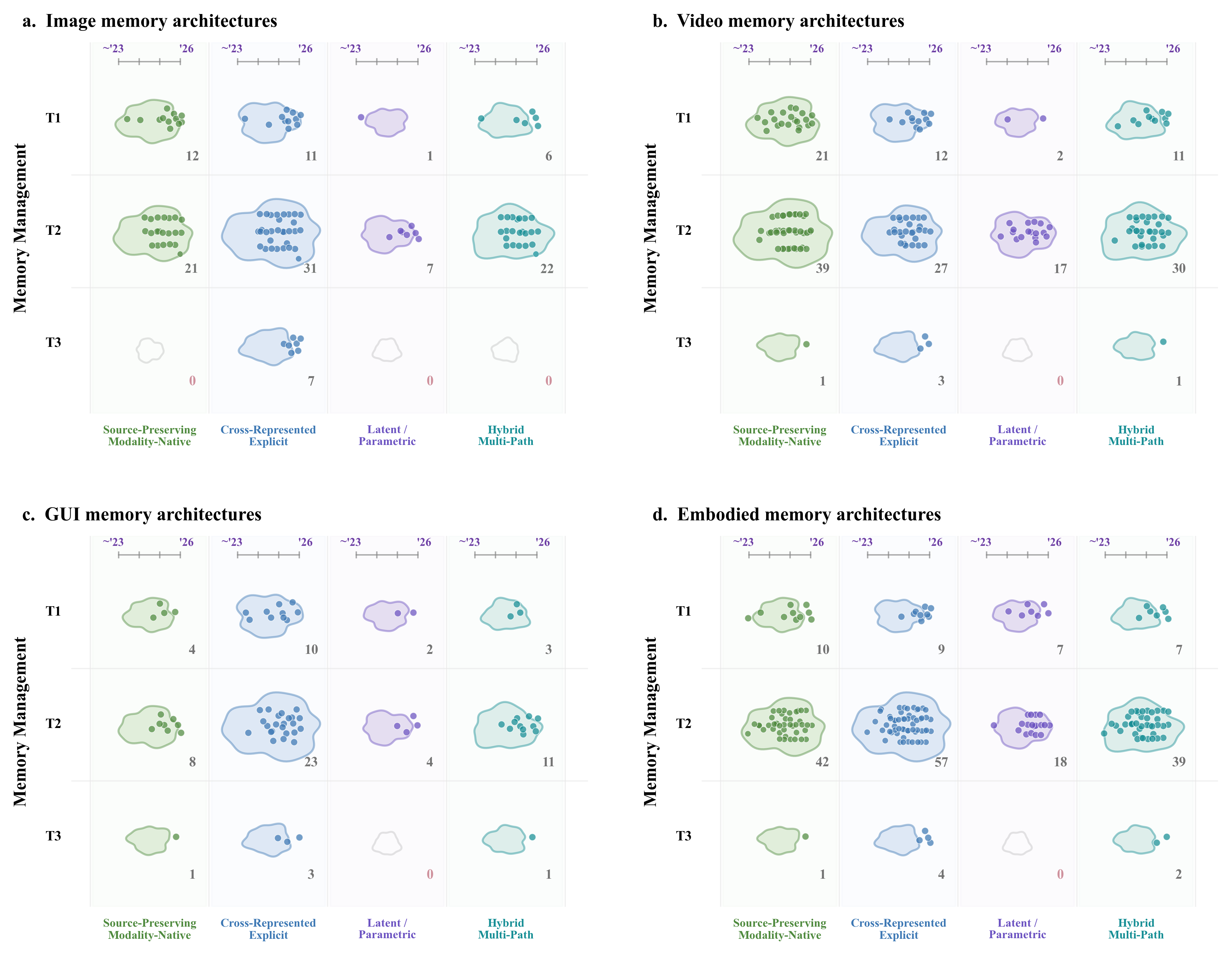}
    \caption{\textbf{T2 Dynamic Memory predominates across image, video, GUI, and embodied memory systems in our taxonomy.}
    Architecture-level incidence is shown for image, video, GUI, and embodied memory systems in the taxonomy-current corpus. Within each modality panel, rows correspond to memory-management tiers and columns correspond to representation families: source-preserving modality-native memory, cross-represented explicit memory, latent or parametric memory, and hybrid multi-path memory. Each dot denotes one architecture, positioned horizontally by publication year from approximately 2023 to 2026; architectures published before 2023 are grouped near the $\sim$'23 position. Irregular envelopes summarize the density of architectures in each cell, and cell labels report counts. Representation-family counts are non-exclusive, so an architecture may be counted in multiple columns when it combines multiple memory representations.}
    \label{fig:modality_matrix_taxonomy_current}
\end{figure}
\begin{figure}[!htbp]
    \centering
    \includegraphics[width=0.92\textwidth]{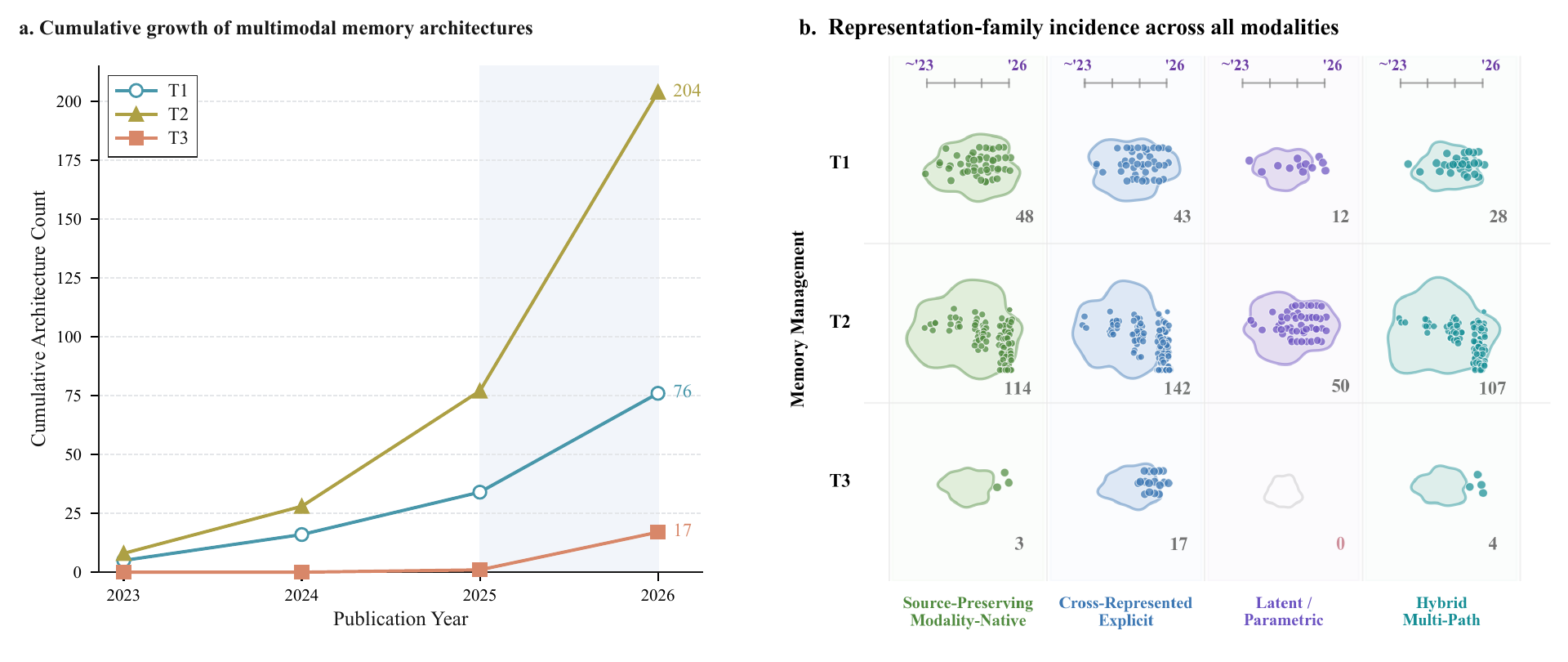}
    \caption{\textbf{Multimodal-agent memory architectures concentrate in T2 Dynamic Memory across modalities.}
    Cumulative counts of architectures in the taxonomy-current corpus show rapid growth after 2025, with most systems classified as T2 Dynamic Memory rather than T1 Static Memory or T3 Self-Evolving Memory. The matrix summarizes architecture-level incidence across memory-management tier and representation family for all modalities. Each dot denotes one architecture, positioned horizontally by publication year; architectures published before 2023 are grouped near the approximate 2023 position, marked as $\sim$'23. Irregular envelopes summarize the density of architectures in each cell, and cell labels report counts. Representation-family counts are non-exclusive, so an architecture may appear in multiple columns when it combines source-preserving evidence, cross-represented explicit records, latent or parametric state, or hybrid multi-path memory.}
    \label{fig:all_modality_mem_development}
\end{figure}

Figure~\ref{fig:cross-matrix} reports the distribution of architectures across the two taxonomy axes. Each architecture is assigned to its highest supported management tier and to one representation category. Hybrid architectures form a separate category in this plot; the remaining architectures are grouped by their representation family. The incidence plots in \Cref{fig:modality_matrix_taxonomy_current,fig:all_modality_mem_development} separately count each representation family present within an architecture, including those in hybrid systems.

\begin{keybox}{Three Corpus-Level Patterns}
\begin{enumerate}[leftmargin=*,itemsep=2pt,topsep=2pt]
    \item T2 Dynamic Memory predominates across various memory systems.
    \item Architectures frequently combine multiple representation types, with explicit and hybrid representations appearing particularly frequently.
    \item Deployment-time evolution of memory-management policies remains
          concentrated in systems with explicit memory artifacts.
\end{enumerate}
\end{keybox}

\Cref{fig:all_modality_mem_development,fig:modality_matrix_taxonomy_current} show that multimodal agent memory architectures concentrate in T2 Dynamic Memory across modalities and in the overall corpus, while T3 Self-Evolving Memory remains relatively rare. Of the \ArchitectureCount{} architectures, \AdaptiveCount{} (\AdaptivePercent{}\%) update memory state or its management policy during operation. T2 Dynamic Memory accounts for \TtwoCount{} architectures (\TtwoPercent{}\% of the corpus) that update memory through stable management policies, whereas T3 Self-Evolving Memory accounts for \TthreeCount{} architectures (\TthreePercent{}\%) that update their management policies based on deployment-time feedback. Stable management policies allow agents to incorporate new experience through predictable and controllable update procedures, making them the dominant approach to memory updating.

Hybrid representations help address the competing requirements of multimodal memory. Hybrid systems constitute \CellHybridTotal{} of the \ArchitectureCount{} architectures (\HybridPercent{}\%), making them the largest representation category. Their prevalence suggests that a single representation type is often insufficient to balance representational fidelity, preservation of modality-specific detail, retrieval efficiency, and storage cost for rich multimodal experiences. Hybrid architectures are therefore the most common representational choice, but they do not dominate the landscape: many systems address narrower memory problems using the single representation type best suited to their tasks.

Compact and explicit memory artifacts are the predominant target for deployment-time policy evolution. Of the \TthreeCount{} T3 Self-Evolving Memory architectures, \CellCrossTThree{} are cross-represented systems, while the other \CellHybridTThree{} are hybrid systems that include an explicit representation. By contrast, to the best of our knowledge, neither the source-preserving category nor the latent or parametric category contains a T3 Self-Evolving Memory system under our definition. Explicit rules, skills, workflows, retrieval descriptions, and programs occupy a useful middle ground for deployment-time evolution: they are more compact and directly editable than detailed source observations, while remaining more inspectable and controllable than information distributed across latent states or model parameters. These properties provide identifiable targets that can be revised from feedback, validated, and reused in subsequent tasks.

The analysis above characterizes multimodal agent memory at the logical level: how experience is represented, how memory state and management policies may change, and what requirements arise from their interaction. 
These logical requirements must ultimately be realized by concrete runtime mechanisms. 
Memory access requires stored representations to be located and materialized for computation, while memory updates require dependent artifacts to remain identifiable, synchronized, and valid as state changes. 
\Cref {sec:memory-infrastructure} therefore examines the systems mechanisms that realize these logical memory designs under finite compute, memory, storage, and communication resources.

\section{Systems Infrastructure for Multimodal Agent Memory}
\label{sec:memory-infrastructure}

\begin{figure*}[t]
  \centering
  \includegraphics[width=\textwidth]{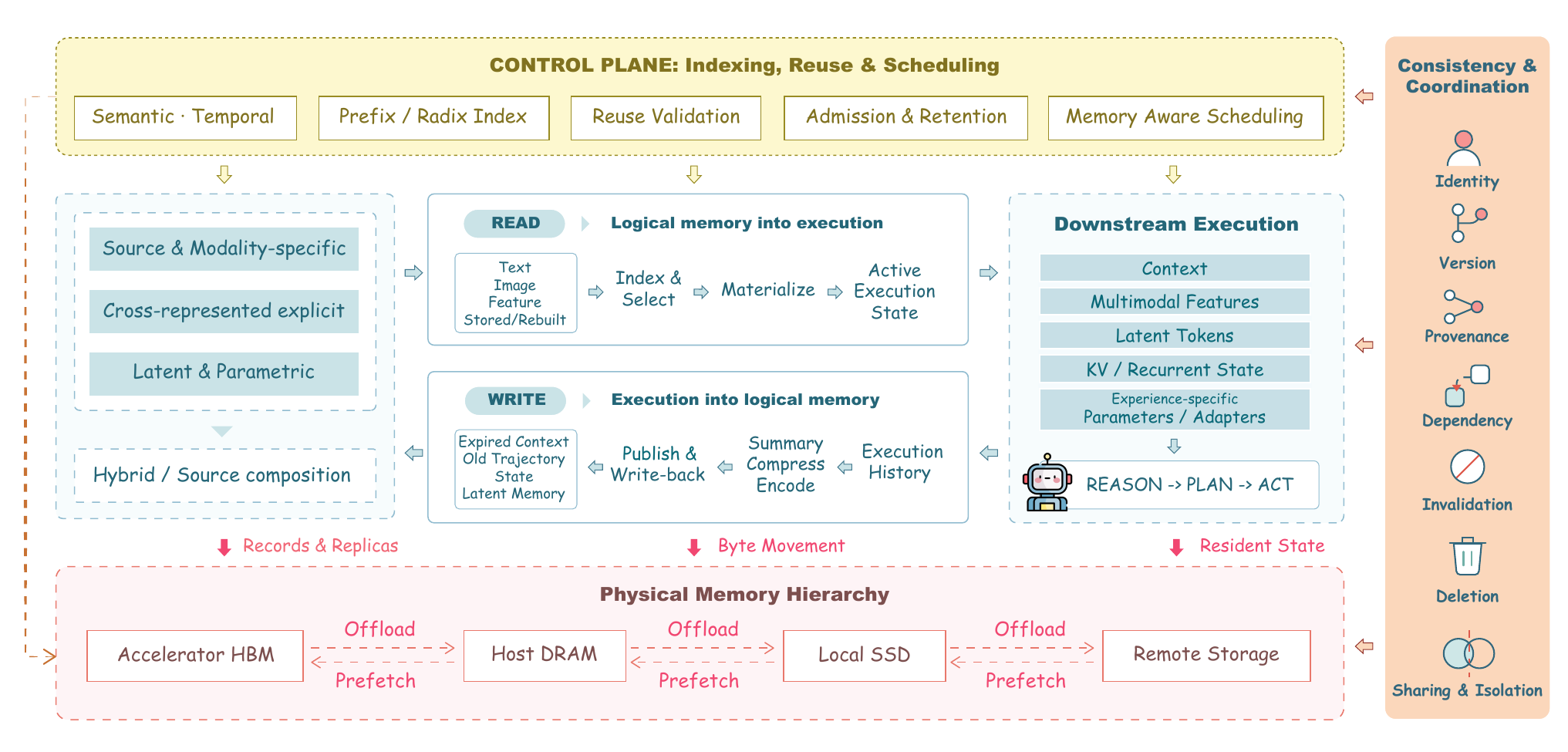}
  \caption{\textbf{Systems realization and lifecycle of multimodal agent memory.}
  A conceptual organization of the mechanisms that connect logical memory with runtime execution and physical storage, rather than an architecture implemented in full by any single analyzed system.
  The \textbf{control plane} (\S\ref{subsec:indexing-scheduling}) covers logical-memory and execution-state lookup, reuse validation, and resource-aware admission and scheduling.
  The \textbf{cross-layer runtime} (\S\ref{subsec:materialization-writeback}) connects memory to computation through materialization and connects execution history to retained memory through write-back.
  The \textbf{physical memory hierarchy} (\S\ref{subsec:storage-residency}, \S\ref{subsec:data-movement}) covers residency, replication, transfer, compression, and reconstruction across accelerator HBM, host DRAM, local SSDs, and remote resources.
  \textbf{Consistency and coordination} (\S\ref{subsec:consistency-coordination}) concern identity, visibility, provenance, invalidation, and isolation across these paths; the guarantees of individual mechanisms remain scoped to the artifacts they manage.}
  \label{fig:systems-memory-realization}
\end{figure*}

\subsection{Scope and System Abstractions}
\label{subsec:scope-abstractions}
\label{sec:logical-physical-memory}

\Cref{sec:formation-and-representation,sec:memory-management,sec:cross-axis-analysis} characterize multimodal agent memory at the logical level. 
Axis I determines what experience is retained and the operative representation through which it becomes available to downstream reasoning or action. 
Axis II determines whether the system reuses fixed memory, updates memory state through a stable policy, or evolves the memory-management policy itself. 
\Cref {sec:cross-axis-analysis} further shows that these choices impose requirements on editability, verifiability, reversibility, and update propagation. This section examines how those logical representations, operations, and requirements are realized through physical storage and runtime execution under finite compute, memory, storage, and communication resources.
These distinctions organize the infrastructure discussion without introducing an additional axis of the memory taxonomy.

\emph{Logical memory} denotes the experience-derived state defined in \Cref{sec:background}, whose operative representation determines its classification in \Cref{sec:formation-and-representation}. When the representation exposes an independently identifiable unit, such as an image, event record, graph node, or trajectory, we call that unit a \emph{logical memory object}. This does not assume that every latent state or parameter update exposes equally addressable semantic units. A \emph{physical realization} comprises the concrete records, tensors, parameter modules, replicas, and auxiliary indices that store this state or make it accessible. For example, PixelRAG indexes screenshot tiles using visual embeddings but supplies the retrieved screenshots to the VLM as source evidence~\citep{pixelrag2026}. An index or storage format therefore supports access to memory without necessarily defining its operative representation.

\emph{Materialized execution state} is the runtime form through which selected memory participates in current computation. Materialization may assemble textual or visual context, construct model-compatible features or latent tokens, restore KV states, or activate an experience-specific parameter module. These are overlapping roles rather than disjoint storage classes: a retained tensor can be both an operative memory representation and an artifact used during execution. Mem-W, for example, constructs latent tokens from retrieved trajectories and expired in-session history for subsequent agent decisions~\citep{memw2026}. Conversely, an ordinary request-local activation or cache does not become agent memory merely because it occupies storage or accelerates decoding. The relevant criterion is its deliberate role in retaining experience for later reasoning or action. Such reuse can occur within a trajectory: TRAM carries auxiliary recurrent states across reasoning steps while resetting them for each input~\citep{tram2026}. Accordingly, persistence here concerns the intended reuse horizon; it does not by itself imply cross-session storage or durability after a process or machine failure.

Logical and physical lifecycles must consequently be distinguished. The tier migration discussed in \Cref{sec:t2-memory-reorganization} changes a memory's temporal or functional role, whereas moving an unchanged artifact from accelerator HBM to host DRAM changes its residency. Likewise, evicting a reconstructible cache or redundant replica need not forget the underlying experience. This separation is conditional on preserving the required information: deleting the last recoverable copy or applying irreversible lossy transformations can affect what the agent can subsequently recall. Physical cache replacement alone does not establish T2 memory-state adaptation, nor does workload-responsive scheduling alone establish T3 evolution of a memory-management policy.

Our coverage therefore includes both systems that retain multimodal experience and transferable mechanisms from model serving, storage, and retrieval. InferCept's treatment of tool-interrupted KV state and SGLang's prefix reuse provide useful runtime mechanisms, but neither mechanism alone specifies how an agent forms or revises experience-derived memory~\citep{infercept2024,zheng2024sglang}. We use such work to explain infrastructure capabilities and their limits, rather than treating every cache optimization as a multimodal memory architecture.

Figure~\ref{fig:systems-memory-realization} organizes this systems view. \Cref{subsec:storage-residency,subsec:data-movement} examine where artifacts reside and the cost of moving or reconstructing them. \Cref{subsec:indexing-scheduling} separates evidence relevance, execution-state compatibility, and scheduling benefit. \Cref{subsec:materialization-writeback} considers transformations between retained experience and active computation, while \Cref{subsec:consistency-coordination} examines ownership, visibility, and compatibility across updates. The remaining integration challenges are discussed in \Cref{sec:open-memory-infrastructure}.

\subsection{Physical Storage and Residency}
\label{subsec:storage-residency}

Long-running agents can accumulate substantially more historical state than an individual request consumes, motivating a separation between retained memory and its active working set~\citep{agentmemorycharacterizationimplications}. 
ReKV provides a multimodal example: it retains video KV caches beyond the current processing window in RAM or on disk, then loads the caches of a query-relevant subset of frames for question answering~\citep{rekv2025}. 
Under a bounded retrieval budget, retained history can therefore grow without a proportional increase in the state loaded for each answer. 
Residency management determines which artifacts occupy each storage tier, how long physical copies remain there, and when to promote, demote, replicate, or evict them.

Storage tiers offer different capacity and access-cost trade-offs. Accelerator HBM accommodates state used by active inference; host DRAM can retain inactive state without occupying GPU capacity; and local SSDs extend retention beyond host memory. ReKV illustrates the latter two options for historical video KV~\citep{rekv2025}. Placement also extends beyond a single machine. Mooncake pools cluster-wide DRAM and SSD resources for shared KV retention, while ITME uses disaggregated, SSD-backed CXL-hybrid memory to expand prefix-cache capacity~\citep{mooncake2025,itmeinferencetieredmemory}. Network-attached pools and CXL-attached expansion provide different access paths, not interchangeable latency guarantees. Moreover, raw media, retrieval embeddings, model-consumed features, and KV states derived from the same experience need not be co-located: their sizes, reuse horizons, and reconstruction costs differ.

Tool calls make retention time an explicit decision. InferCept compares \emph{Preserve}, \emph{Swap}, and \emph{Discard}: keeping intercepted KV on the GPU, moving it to host memory, or reconstructing it after resumption~\citep{infercept2024}. Continuum adds a time-to-live (TTL) to GPU retention, balancing the cost of occupying capacity against avoided reload or recomputation and subsequent queueing delay~\citep{continuum2025}. TTL expiration ends protection under the scheduler's rules; it does not make the remembered evidence semantically obsolete. MORI instead ranks complete agent programs by recent relative idleness, retaining busier programs in HBM and more-idle programs in DRAM~\citep{mori2026}. Its sticky placement changes tiers when capacity pressure requires demotion or available HBM permits promotion, avoiding migration at every tool-call boundary. These mechanisms use different signals---interruption cost, retention time, and program activity---to balance occupancy against resumption overhead.

Distributed reuse introduces a second placement dimension: which nodes retain an artifact and how many copies are maintained. Mooncake tracks cache locations and supports replication so that frequently accessed KV blocks can be available on multiple workers~\citep{mooncake2025}. Local replicas reduce repeated remote access but increase aggregate storage; shared pools reduce the need to duplicate every artifact but expose remote-transfer costs. For multimodal memory, these decisions can be made separately for source media and expensive derived states. They must also respect the distinction between sharing an observation and reusing its computation: identical historical evidence does not imply that its KV state is compatible with every consuming context.

The central residency trade-off is therefore not simply hot versus cold storage. It combines expected reuse time, state size, reconstruction cost, concurrent demand, and access locality. Program admission and worker selection interact with these decisions, as discussed in \Cref{subsec:indexing-scheduling}. The cost of implementing a residency change depends on the transfer and restoration mechanisms considered next.

\subsection{Data Movement, Compression, and Reconstruction}
\label{subsec:data-movement}

A retained artifact is not necessarily ready for execution. Storage access, interconnect transfer, decompression, and layout conversion can delay its use even when lookup succeeds. These costs concern the restoration path; they are distinct from deciding where an artifact should reside or whether it is compatible with the current request.

Pipelining reduces the transfer latency exposed on the critical path. CachedAttention uses layer-wise preloading and asynchronous saving to overlap KV access with computation~\citep{cachedattention2024}. Across nodes, Mooncake supports asynchronous transfers over RDMA-capable infrastructure and streams newly produced KV toward decode workers during prefill~\citep{mooncake2025}. Buffering enables overlap but consumes additional memory, and concurrent transfers can compete with inference for bandwidth. Consequently, an asynchronous interface alone does not guarantee that restoration is hidden.

Compression and partial recomputation offer complementary ways to reduce transferred bytes. CacheGen combines quantization and entropy coding with chunked streaming, overlapping transmission with codec decoding~\citep{cachegen2024}. Cake concurrently recomputes from the beginning of a reusable prefix and loads cached KV from its end, adapting the division of work to available computation and I/O resources~\citep{cake2025}. The trade-off is therefore among transferred bytes, codec work, temporary storage, and recomputation. Lossy compression additionally affects approximation quality. In particular, a compact transport bitstream need not occupy proportionally less HBM after it is decoded into inference tensors.

Multimodal and agent workloads provide additional overlap windows. MPIC loads cached image state while computing missing image caches; this overlap is distinct from its selective recomputation for contextual reuse~\citep{mpic2025}. TokenCake exploits tool-call intervals for opportunistic asynchronous offload and predictive upload, and initiates immediate upload when a tool returns earlier than predicted~\citep{tokencake2025}. Workflow-informed prefetching also uses anticipated agent execution, as discussed for KVFlow in \Cref{subsec:indexing-scheduling}. These mechanisms reduce exposed restoration cost when sufficient independent work is available, but short intervals, prediction errors, and contention can leave transfers on the critical path. Completing a transfer establishes availability, not reuse validity.

\subsection{Indexing, Reuse, and Scheduling}
\label{subsec:indexing-scheduling}

The request-time control path separates three questions: which memory is relevant, whether a retained execution artifact is compatible, and whether reusing it is worthwhile. A \emph{logical-memory index} identifies accessible evidence or experience; an \emph{execution-state index} identifies potentially reusable computation associated with concrete inputs and execution dependencies. A hit in the first does not imply a hit in the second, and neither hit alone determines the best execution plan.

Logical-memory access reflects the organization of retained experience. M3-Agent combines semantic similarity with timestamp-based sequential retrieval and entity-based associative retrieval over text, image, and audio memory~\citep{long2025m3agent}. DynaMem couples language-based localization with a spatial-semantic map and can retrieve the latest image associated with a selected voxel~\citep{liu2025dynamem}. These access paths exploit temporal order, entity links, spatial structure, and modality-bearing evidence. Their results identify what the agent should inspect; they do not certify the reusability of tensors previously computed from those objects.

Execution-state indexing instead exploits computational dependencies. SGLang's RadixAttention organizes token prefixes and their KV states in a radix tree, with image hashes representing multimodal inputs within that indexing scheme~\citep{zheng2024sglang}. Exact-prefix reuse requires matching the relevant input prefix and computational configuration, including positions and model dependencies; an image hash by itself is insufficient. Prompt Cache enables modular reuse within an explicitly defined positional schema, whereas CacheBlend selectively recomputes cached states to restore cross-chunk contextualization~\citep{gim2024promptcache,yao2025cacheblend}. MPIC similarly addresses image-context reuse through position handling and selective recomputation~\citep{mpic2025}. These methods should not be treated as equivalent guarantees: schema-constrained reuse and quality-evaluated approximate reconstruction differ from exact reuse of an unchanged prefix.

Scheduling determines how compatible reuse opportunities interact with resource demand. SGLang orders requests to exploit common cached prefixes, while Preble jointly considers prefix reuse and load balancing when dispatching requests across workers~\citep{zheng2024sglang,srivatsa2025preble}. Reusing a prefix on an overloaded worker can be less attractive than recomputing it elsewhere. For retrieval-mediated requests, evidence relevance governs what enters computation, whereas cache locality concerns where that computation can run efficiently. Neither should silently substitute for the other.

Agent execution provides signals beyond individual requests. ThunderAgent treats a complete agent program as a scheduling unit, using execution phase and context footprint to pause or restore programs under memory pressure~\citep{kang2026thunderagent}. Its global waiting queue permits paused programs to resume on a worker with available capacity, rather than requiring their original placement. KVFlow exploits an exposed Agent Step Graph to estimate steps-to-execution, prioritize fixed-prefix KV nodes, and prefetch upcoming state from CPU to GPU~\citep{kvflow2025}. Its workflow-level reuse signal depends on an exposed graph, which may include conditional branches. Together with MORI's tier-level admission control~\citep{mori2026}, these designs connect execution readiness to cache retention and capacity allocation.

Here, admission concerns requests, programs, or physical cache entries, rather than the semantic decision to write an observation into logical memory. A useful reuse decision must account for validation, transfer, reconstruction, queueing, and displaced work, not just the amount of cached state. \Cref{subsec:materialization-writeback} examines the transformations selected by this control path, and \Cref{subsec:consistency-coordination} examines how subsequent updates affect their eligibility for reuse.

\subsection{Materialization and Write-back}
\label{subsec:materialization-writeback}

Materialization and write-back implement the two directions of the logical memory lifecycle described in \Cref{sec:formation-and-representation,sec:memory-management}.
On the read path, materialization converts a selected logical memory representation into model-consumable context, features, latent states, or parameter modules for current computation. 
On the write path, observations, execution history, or completed trajectories are transformed into persistent state that can subsequently participate in memory formation or management. 
The systems concern here is therefore not what should be remembered or updated, but when these transformations execute, what artifacts they produce, and what preparation and consumption costs they impose.

Textual and source-level inputs have different preparation costs. PixelRAG renders and indexes screenshot tiles during datastore construction, then supplies retrieved pixels to a VLM. Downsampling before reading reduces visual-token demand without rebuilding the datastore or retrieval index~\citep{pixelrag2026}. FOLIO supplies compact semantic records and can recover linked keyframes when its relevance criterion indicates insufficient grounding~\citep{folio2026}. Reopening images may recover details missing from an abstraction, but incurs visual encoding and additional model input. Storage compression, visual-token reduction, and downstream attention cost are therefore distinct efficiency measures.

Latent-token and KV-based paths move work across different boundaries. Mem-W uses a visual--textual encoder and a Q-Former-based compressor to produce a fixed number of soft memory tokens per selected trajectory segment~\citep{memw2026}. The limit bounds expansion per segment, not the total cost of encoding or selecting multiple segments. ReKV instead constructs video KV incrementally before a question arrives and binds retrieved blocks into question-answering attention~\citep{rekv2025}. Its retrieved tokens receive consecutive positions rather than their original video positions. This is part of a specific attention and positional scheme, not evidence that arbitrary cached tensors can be reused after context changes. Latent compression pays preparation work to reduce the model-facing input, whereas KV precomputation moves part of that work ahead of the query and retains the resulting tensors.

Parametric memory follows a further path that is not captured by context insertion alone. Frames2LoRA generates a video-specific LoRA adapter from video-conditioned representations through a trained hypernetwork; subsequent questions use the adapted model without replaying the original visual tokens~\citep{video2lora2026}. From a runtime perspective, the relevant artifacts are the generated parameter module and the backbone to which it is attached. The corresponding costs include adapter production, retention, loading, and activation, rather than only context length. This complements the parametric representation discussion in \Cref{sec:parametric_internalization} without treating ordinary pretrained model weights as experience-derived agent memory.

Write-back must likewise distinguish reuse horizons. Mem-W carries expired interaction history forward as compressed latent context, while MementoGUI both compacts older working-memory entries and uses a separate Episodic Writer to construct source-linked records for later episodes~\citep{memw2026,mementogui2026}. In-session compaction preserves task progress; episodic write-back creates a cross-episode retrieval artifact. Both can retain experience for later use, but only episodic write-back crosses that external episodic-storage boundary. Neither description alone specifies crash durability. Producing summaries, organizing visual crops, and constructing retrieval embeddings also consume resources before later reads can benefit.

Reasoning history can become either recurrent state or procedural experience. TRAM pools completed-step activations into fast and slow recurrent states that influence later reasoning within the same input~\citep{tram2026}. RRM instead distills completed retrieval trajectories and feedback into procedural records, excluding historical answers and video-specific facts; retrieved procedures guide subsequent evidence acquisition rather than being inserted as factual answer context~\citep{fan2026rrm}. These paths differ in both what persists and the model interface through which it is reused.

Write-back may produce linked artifacts asynchronously. FOLIO builds semantic records associated with retained keyframes while a short-term visual buffer supplies recent evidence; M$^{2}$A maintains raw message records alongside editable semantic observations linked by evidence identifiers~\citep{folio2026,feng2026m2a}. Retaining these links enables later source recovery but entails constructing and maintaining both abstractions and supporting artifacts. The cost of memory therefore spans construction, retrieval, and consumption, rather than stored size alone~\citep{agentmemorycharacterizationimplications}. Publication and update coordination across these artifacts are considered next.

\subsection{Consistency and Coordination}
\label{subsec:consistency-coordination}

Consistency and coordination provide the systems support for the T2 representation requirements identified in \Cref{representation_for_dynamic_memory}.
When memory is updated, later reads may encounter different record versions, physical replicas, or derived execution states. 
Editability requires the system to identify the affected logical object and its physical realizations; 
verifiability requires provenance and version information connecting the update to supporting evidence; reversibility requires an earlier version, operation log, checkpoint, or reconstructible source;
and update propagation requires dependent artifacts to be refreshed or invalidated. 
Coordination determines when these changes become visible and which versions or derivatives remain eligible for use. 
Deciding whether new evidence should revise a belief remains a memory-management question, whereas publishing the selected revision consistently across access paths is a systems question.

Update ownership and evidence identity provide a starting point. M$^{2}$A gives its MemoryManager exclusive read--write access to the memory bank and links semantic observations to supporting messages~\citep{feng2026m2a}. This identifies an application-level owner and evidence trail, not a distributed transaction protocol. When a dependency changes, the associated derivation must be refreshed or its compatibility re-established before it is used for a request requiring the revised state. Historical queries may still use an earlier version when its temporal scope and access permissions permit. This selective treatment avoids invalidating unchanged source evidence merely because an interpretation has been corrected.

Publication rules determine what readers see while writes proceed. FOLIO excludes segment records still being written and instead answers from completed memory and recent visual context~\citep{folio2026}. Manu coordinates vector-store subscribers through timestamped logs and time-tick progress markers, allowing queries to wait for a configured freshness level~\citep{guo2022manu}. These mechanisms operate at different boundaries: application records in the former and indexed data with row-level ACID support in the latter. Their guarantees do not automatically compose into atomic publication across separately maintained media, semantic records, retrieval indices, and model-side states.

Shared execution state also requires lifetime and aliasing control. Mooncake tracks prefix-dependent cache blocks and their replicas and protects blocks used by ongoing requests from eviction~\citep{mooncake2025}. Within an inference engine, vLLM uses reference-counted block sharing and copy-on-write when a sequence extends a shared KV block~\citep{kwon2023vllm}. These mechanisms prevent premature reclamation and interference between branches, but do not determine whether the underlying memory has become stale. Similarly, Manu's deletion bitmaps suppress deleted vectors in search before segment-index rebuilding~\citep{guo2022manu}. Read exclusion, storage reclamation, and invalidation of derived representations are separate operations.

Authorization imposes another visibility boundary. Collaborative Memory separates private and shared fragments and uses provenance and a current access graph to filter retrieval as permissions change~\citep{collabmem2025}. This controls exposure through its retrieval interface; it does not demonstrate retroactive removal of information already materialized in another agent's context or KV state. Authorization is consequently distinct from copy-on-write: one governs who may consume an artifact, while the other protects shared physical state against a branch's mutations.

Compatibility can change even when source observations do not. FastFill aligns embeddings across model updates and prioritizes partial backfilling using alignment uncertainty~\citep{jaeckle2023fastfill}. This addresses retrieval features, not arbitrary model-consumed latent memory. DroidSpeak selectively recomputes layer groups when reusing KV across profiled fine-tuned model variants~\citep{droidspeak2026}; its scope is model-pair-specific, and substantial distribution drift can degrade quality. Compatibility therefore depends on the producer, consumer, and validated transformation, not source identity alone.

The analyzed mechanisms coordinate particular records, retrieval views, or execution blocks. Their combination exposes trade-offs among freshness, read latency, sharing, bookkeeping, and reconstruction. A retained historical observation can remain valid after a current-state record changes, while an unchanged physical cache can become incompatible with a new consumer. Establishing explicit guarantees across these boundaries is the integration problem developed in \Cref{sec:open-memory-infrastructure}.  \Cref{subsec:storage-residency,subsec:consistency-coordination} do not introduce new memory types; they characterize the runtime conditions under which the representation and management choices in \Cref{sec:formation-and-representation,sec:memory-management,sec:cross-axis-analysis} can be served efficiently, consistently, and within finite resources.

\section{Evaluation and Applications}
\label{sec:evaluation}

\Cref{sec:cross-axis-analysis} analyzes how representation constrains memory management, while \Cref{sec:memory-infrastructure} examines how the resulting designs are realized at runtime. This section characterizes how multimodal memory systems are evaluated and what functions memory serves in their applications. \Cref{sec:eval-dimensions} defines the capabilities
that memory benchmarks attempt to measure, \Cref{sec:benchmark-landscape}
characterizes the \BenchmarkCount{} benchmark records by evaluation setting,
\Cref{sec:application-regimes} describes the functions memory serves in deployed systems,
and \Cref{sec:benchmark-coverage} reports what those benchmarks do and do not exercise.
What the field should do about the gaps recorded here is argued in \Cref{sec:agenda}.

\subsection{Evaluation Dimensions for Multimodal Agent Memory}
\label{sec:eval-dimensions}

End-task accuracy alone cannot establish whether an agent has used memory: a correct output may follow from the current context or parametric knowledge, while an error may arise during memory formation, access, maintenance, or utilization. 
Recent benchmarks therefore increasingly evaluate not only task outcomes, but also the evidence and state transitions supporting them~\citep{memlens2026,worldmemarena2026}. 
Accordingly, \Cref{tab:mm-memory-benchmarks} characterizes each benchmark through shared memory capabilities, modality-specific requirements, and evaluation settings. The six shared dimensions capture three stages of memory use.


\begin{keybox}{Six Shared Evaluation Dimensions}
\begin{enumerate}[leftmargin=*,itemsep=3pt,topsep=3pt]
    \item \textbf{Evidence access} comprises cross-modal evidence (CE), which
          requires complementary modalities to be jointly used, and long-range
          evidence (LE), which distributes relevant information across turns,
          sessions, frames, or events.

    \item \textbf{Memory validity} comprises state changes (SC), conflicting
          memories (CF), and unanswerable queries (UQ): these test whether an
          agent can distinguish current from obsolete information, resolve
          incompatible records using temporal or provenance cues, and recognize
          when available memory is insufficient.

    \item \textbf{Memory-grounded action} (MA) evaluates whether retrieved
          experience influences planning or interaction.
\end{enumerate}
\end{keybox}

These dimensions are modality-independent: they characterize whether memory preserves sufficient evidence, remains valid under change, and contributes to downstream decisions.

Shared dimensions alone, however, cannot express the internal structure that must survive within each modality. 
Image-memory benchmarks therefore distinguish fine-grained visual-detail retention (FG), entity tracking across observations (ET), and visual state-change tracking (VC). 
Video benchmarks add temporal localization (TL), temporal-relation reasoning over event order and duration (TR), and multi-event evidence integration (ME)~\citep{flashvstream2025,sember2026}. 
Audio and speech benchmarks require speaker tracking (SP), acoustic-cue retention (AC), and repair persistence (RP). SP tests whether a memory preserves who spoke across turns or sessions; AC tests whether it retains non-lexical evidence such as prosody, vocal style, and background sound; and RP tests whether spoken feedback or later evidence produces a persistent correction rather than a transient response~\citep{voicemem2026,agenticasr2026,statefulasr2026}. 
GUI evaluation emphasizes persistent interface state (GS), associations between actions and resulting states (AO), and reuse of experience across sessions (XR)~\citep{memguibench2026}. 
Embodied benchmarks instead require spatial memory (SM), object-state tracking under partial observability (OS), and task-state or progress tracking (TS)~\citep{threedllmmem2025,findingdory2025}. 
These dimensions expose representational failures that long-range or cross-modal labels cannot: an agent may retain an event while losing its precise time, remember an object while failing to preserve its location, or recall a GUI action without the state transition it produced.

The capability annotations should be interpreted together with the benchmark setup recorded in \Cref{tab:mm-memory-benchmarks}. The data source distinguishes real, synthetic, simulated, and hybrid settings, which differ in perceptual complexity, environmental control, and the availability of counterfactual conditions. 
The task field identifies whether memory supports question answering, retrieval, localization, generation, action, planning, learning, or mapping. 
The ground-truth field records what can be verified, including answers, supporting evidence, regions, timestamps, environment states, actions, provenance, and reliability labels. Answer-only supervision evaluates end-task correctness but provides limited evidence that the intended memory was used.
By contrast, evidence, region, timestamp, and provenance annotations support more direct evaluation of memory grounding, while state and action supervision connect memory to interactive behavior. These setup dimensions therefore determine how precisely a benchmark can attribute success or failure to memory, even when two benchmarks nominally cover the same capability.

These dimensions also connect benchmark coverage to the taxonomy introduced earlier. 
CE and the modality-specific dimensions primarily specify the evidence, attributes, and relations that memory formation must preserve and make accessible, and therefore connect most directly to Axis~I. LE, TL, ME, and XR probe the storage, access, and reuse of fixed memory and therefore align with T1 Static Memory, whereas SC, CF, VC, GS, AO, OS, and TS can expose demands for state maintenance or revision under changing observations and therefore inform evaluation of T2 Dynamic Memory. Offline reconstruction of a state history, however, does not by itself establish deployment-time memory updating.
UQ and MA primarily evaluate memory utilization through evidence sufficiency and action selection. 
None of these labels alone establishes T3 Self-Evolving Memory, which requires deployment feedback to produce persistent changes to a reusable memory-management policy. This mapping supports the lifecycle-based coverage analysis in \Cref{sec:benchmark-coverage}.


\subsection{Benchmark Landscape Across Modalities}
\label{sec:benchmark-landscape}

\Cref{tab:mm-memory-benchmarks} organizes \BenchmarkCount{} selected benchmark resources and evaluation protocols by their evaluation setting. The selection includes dedicated memory tests and broader tasks used to evaluate systems that retain and reuse multimodal experience. Such tasks can reveal the contribution of memory even when their scores also depend on perception, reasoning, planning, or execution. Selected text-only and semantic-state protocols provide comparisons for shared memory functions; they do not establish visual or acoustic retention. Inclusion therefore indicates evaluation relevance, while the capability annotations describe what each protocol directly or indirectly measures.
Beyond differences in sensory input, these groups vary in the experience that must persist, the horizon over which it is reused, and the supervision available to verify its contribution.

\begingroup
\setlength{\LTleft}{\fill}
\setlength{\LTright}{\fill}
\setlength{\LTcapwidth}{\textwidth}
\setlength{\LTpre}{0.35em}
\setlength{\LTpost}{0.55em}
\setlength{\tabcolsep}{1pt}
\renewcommand{\arraystretch}{1.06}
\definecolor{MMGreen}{HTML}{16833A}
\definecolor{MMOrange}{HTML}{C46A00}
\definecolor{MMRed}{HTML}{C62828}
\definecolor{MMBlue}{HTML}{EAF2F8}

\newcommand{\mmYes}{\textcolor{MMGreen}{\ding{51}}}
\newcommand{\mmPart}{\textcolor{MMOrange}{\ensuremath{\circ}}}
\newcommand{\mmNo}{\textcolor{MMRed}{\ding{55}}}
\newcommand{\mmCrossDomain}{\textsuperscript{\ensuremath{\ddagger}}}
\newcommand{\mmPlus}{+\allowbreak}
\newcommand{\mmSlash}{/\allowbreak}

\newcommand{\MMBenchHeader}[4]{%
  \toprule
  \multicolumn{1}{c}{} &
  \multicolumn{3}{c}{\textbf{Dataset setup}} &
  \multicolumn{6}{c}{\shortstack{\textbf{Memory capabilities}\\\textbf{evaluated}}} &
  \multicolumn{3}{c}{\shortstack{\textbf{Modality-}\\\textbf{specific}\\\textbf{capabilities}}} \\
  \cmidrule(lr){2-4}\cmidrule(lr){5-10}\cmidrule(lr){11-13}
  \textbf{Benchmark} & \textbf{Source} & \textbf{Task} & \textbf{Ground truth} &
  {\tiny\bfseries CE} & {\tiny\bfseries LE} & {\tiny\bfseries SC} &
  {\tiny\bfseries CF} & {\tiny\bfseries UQ} & {\tiny\bfseries MA} &
  {\tiny\bfseries #2} & {\tiny\bfseries #3} & {\tiny\bfseries #4} \\
  \midrule
}

\captionsetup{hypcap=false}
\captionof{table}{Selected benchmarks and evaluation protocols for multimodal agent-memory research, grouped by evaluation setting.}
\label{tab:mm-memory-benchmarks}

\edef\MMBenchSavedTableCounter{\number\value{table}}

\captionsetup[longtable]{labelformat=empty,justification=centering,singlelinecheck=true,font=normalsize,skip=2pt}

\scriptsize

\begin{longtable}{@{}>{\raggedright\arraybackslash}p{6.42cm}>{\raggedright\arraybackslash}p{1.02cm}>{\raggedright\arraybackslash}p{1.83cm}>{\raggedright\arraybackslash}p{2.21cm}*{9}{>{\centering\arraybackslash}p{0.52cm}}@{}}
\caption*{\textbf{(a) Image memory}}\\[-2pt]
\MMBenchHeader{Image}{FG}{ET}{VC}
\endfirsthead
\caption*{\textbf{(a) Image memory (continued)}}\\[-2pt]
\MMBenchHeader{Image}{FG}{ET}{VC}
\endhead
\midrule
\multicolumn{13}{r}{\footnotesize Continued on next page}\\
\endfoot
\bottomrule
\endlastfoot

LoCoMo~\citep{maharana2024evaluating} & Hyb. & QA/Gen. & Ans.+Ev. &
\mmPart&\mmYes&\mmPart&\mmNo&\mmYes&\mmNo&\mmNo&\mmNo&\mmNo\\

MMDU~\citep{liu2024mmdu} & Hyb.  & QA/Gen.  & Ans.  &
\mmYes&\mmPart&\mmNo&\mmNo&\mmNo&\mmNo&\mmPart&\mmPart&\mmNo\\

MMRC~\citep{xue2025mmrc} & Real  & QA/Gen.  & Ans.  &
\mmYes&\mmPart&\mmYes&\mmNo&\mmYes&\mmNo&\mmPart&\mmPart&\mmYes\\

Memory-QA~\citep{memoryqa2025} & Real  & QA/Ret.  & Ans.+Ev.  &
\mmPart&\mmYes&\mmNo&\mmNo&\mmNo&\mmNo&\mmYes&\mmNo&\mmPart\\

LCMP~\citep{hong2026tame} & Syn.  & QA/Gen.  & Ans.  &
\mmYes&\mmYes&\mmYes&\mmNo&\mmNo&\mmNo&\mmPart&\mmYes&\mmYes\\

DISBench~\citep{deepimagesearch2026} & Real  & Ret./Loc.  & Ev.  &
\mmPart&\mmYes&\mmNo&\mmNo&\mmNo&\mmNo&\mmPart&\mmYes&\mmPart\\

MMA-Bench~\citep{mma2026} & Syn.  & QA/Ref.  & Ans.+Rel.  &
\mmYes&\mmYes&\mmYes&\mmYes&\mmYes&\mmNo&\mmPart&\mmNo&\mmYes\\

Mem-Gallery~\citep{memgallery2026} & Syn.  & QA/Ret.  & Ans.+Ev.  &
\mmYes&\mmYes&\mmYes&\mmYes&\mmYes&\mmNo&\mmPart&\mmPart&\mmYes\\

ATM-Bench~\citep{atmbench2026} & Real  & QA/Ret.  & Ans.+Ev.  &
\mmYes&\mmYes&\mmYes&\mmYes&\mmYes&\mmNo&\mmYes&\mmYes&\mmYes\\

Persona-MME~\citep{personavlm2026} & Hyb.  & QA/Gen.  & Ans.  &
\mmYes&\mmYes&\mmYes&\mmNo&\mmNo&\mmNo&\mmPart&\mmYes&\mmYes\\

MemEye~\citep{memeye2026} & Hyb.  & QA  & Ans.+Ev.  &
\mmYes&\mmYes&\mmYes&\mmNo&\mmNo&\mmNo&\mmYes&\mmPart&\mmYes\\

MemLens~\citep{memlens2026} & Syn.  & QA/Ret.  & Ans.+Ev.  &
\mmYes&\mmYes&\mmYes&\mmNo&\mmYes&\mmNo&\mmYes&\mmPart&\mmYes\\

SMMBench~\citep{smmbench2026} & Hyb.  & QA\mmSlash Act.  & Ans.\mmPlus Ev.\mmPlus Prov.  &
\mmYes&\mmYes&\mmYes&\mmYes&\mmNo&\mmYes&\mmPart&\mmYes&\mmPart\\

VisualMem~\citep{visualmem2026} & Hyb.  & QA/Ret.  & Ans.+Ev.  &
\mmYes&\mmYes&\mmPart&\mmNo&\mmNo&\mmNo&\mmYes&\mmNo&\mmPart\\

CamRoll~\citep{camrollagent2026} & Real  & QA/Ret.  & Ans.+Ev.  &
\mmPart&\mmYes&\mmPart&\mmNo&\mmNo&\mmNo&\mmYes&\mmYes&\mmPart\\

RefMem-Bench~\citep{refmembench2026} & Syn.  & QA/Gen.  & Ans.+Ev.  &
\mmYes&\mmYes&\mmPart&\mmNo&\mmNo&\mmNo&\mmPart&\mmPart&\mmPart\\

M$^3$Exam~\citep{m3exam2026} & Hyb.  & QA/Ret.  & Ans.+Ev.  &
\mmYes&\mmYes&\mmPart&\mmNo&\mmNo&\mmNo&\mmYes&\mmYes&\mmPart\\

M2CA-Bench~\citep{wang2026cma} & Syn.  & Ret./Gen.  & Ev.+Ans.  &
\mmYes&\mmYes&\mmPart&\mmNo&\mmNo&\mmNo&\mmPart&\mmYes&\mmPart\\

MemLeak~\citep{wang2026memleakdiagnosinginformationleaks} & Hyb. & QA & Ans. &
\mmPart&\mmYes&\mmPart&\mmNo&\mmNo&\mmNo&\mmPart&\mmNo&\mmNo\\

\end{longtable}


\begin{longtable}{@{}>{\raggedright\arraybackslash}p{6.42cm}>{\raggedright\arraybackslash}p{1.02cm}>{\raggedright\arraybackslash}p{1.83cm}>{\raggedright\arraybackslash}p{2.21cm}*{9}{>{\centering\arraybackslash}p{0.52cm}}@{}}
\caption*{\textbf{(b) Video memory}}\\[-2pt]
\MMBenchHeader{Video}{TL}{TR}{ME}
\endfirsthead
\caption*{\textbf{(b) Video memory (continued)}}\\[-2pt]
\MMBenchHeader{Video}{TL}{TR}{ME}
\endhead
\midrule
\multicolumn{13}{r}{\footnotesize Continued on next page}\\
\endfoot
\bottomrule
\endlastfoot

VStream-QA~\citep{flashvstream2025} & Real  & QA  & Ans.+Time  &
\mmPart&\mmYes&\mmPart&\mmNo&\mmNo&\mmNo&\mmYes&\mmPart&\mmYes\\

StreamingBench~\citep{streamingbench2024} & Real  & QA  & Ans.+Time  &
\mmPart&\mmYes&\mmPart&\mmNo&\mmNo&\mmNo&\mmYes&\mmPart&\mmYes\\

StreamBench~\citep{xiong2025streamchat} & Hyb.  & QA/Gen.  & Ans.+Time  &
\mmYes&\mmYes&\mmPart&\mmNo&\mmNo&\mmNo&\mmYes&\mmPart&\mmYes\\

EgoLifeQA~\citep{egolife2025} & Real  & QA/Ret.  & Ans.+Time  &
\mmYes&\mmYes&\mmPart&\mmNo&\mmNo&\mmNo&\mmYes&\mmYes&\mmYes\\

HippoVlog~\citep{hippomm2025} & Real  & QA/Ret.  & Ans.+Time  &
\mmYes&\mmYes&\mmPart&\mmNo&\mmNo&\mmNo&\mmYes&\mmYes&\mmYes\\

M3-Bench~\citep{long2025m3agent} & Real  & QA  & Ans.  &
\mmYes&\mmYes&\mmPart&\mmNo&\mmNo&\mmNo&\mmYes&\mmYes&\mmYes\\

TeleEgo~\citep{teleego2025} & Real  & QA  & Ans.+Time  &
\mmYes&\mmYes&\mmYes&\mmNo&\mmNo&\mmNo&\mmYes&\mmYes&\mmYes\\

ELViM~\citep{sun2026videosalmonnsmemoryenhancedstreaming} & Real  & QA\mmSlash Learn  & Ans.+Time  &
\mmPart&\mmYes&\mmPart&\mmNo&\mmNo&\mmYes&\mmYes&\mmYes&\mmYes\\

EgoMem (VideoLucy)~\citep{videolucy2025} & Real  & QA/Gen.  & Ans.+Time  &
\mmYes&\mmYes&\mmPart&\mmNo&\mmNo&\mmNo&\mmYes&\mmYes&\mmYes\\

RIVER Bench~\citep{riverbench2026} & Hyb.  & QA\mmSlash Act.  & Ans.\mmPlus Time\mmPlus Act.  &
\mmYes&\mmYes&\mmYes&\mmNo&\mmNo&\mmYes&\mmYes&\mmYes&\mmYes\\

MA-EgoQA~\citep{kim2026maegoqa} & Real  & QA/Ret.  & Ans.+Time  &
\mmYes&\mmYes&\mmPart&\mmNo&\mmNo&\mmNo&\mmYes&\mmYes&\mmYes\\

LVOmniBench~\citep{tao2026lvomnibench} & Real  & QA  & Ans.  &
\mmYes&\mmYes&\mmPart&\mmNo&\mmNo&\mmNo&\mmPart&\mmYes&\mmYes\\

MM-Lifelong~\citep{chen2026mmlifelong} & Hyb.  & QA/Loc.  & Ans.+Time  &
\mmYes&\mmYes&\mmYes&\mmNo&\mmNo&\mmNo&\mmYes&\mmYes&\mmYes\\

EgoMemReason~\citep{egomemreason2026} & Real  & QA/Ret.  & Ans.\mmPlus Ev.\mmPlus Time  &
\mmPart&\mmYes&\mmPart&\mmNo&\mmNo&\mmNo&\mmYes&\mmYes&\mmYes\\

EgoExoMem~\citep{egoexomem2026} & Real  & QA  & Ans.+Time  &
\mmYes&\mmYes&\mmPart&\mmNo&\mmNo&\mmNo&\mmYes&\mmYes&\mmYes\\

M$^3$Eval~\citep{m3eval2026} & Hyb.  & QA  & Ans.+Time  &
\mmPart&\mmYes&\mmPart&\mmNo&\mmNo&\mmNo&\mmPart&\mmYes&\mmYes\\

SuperMemory-VQA~\citep{supermemoryvqa2026} & Real  & QA/Ret.  & Ans.\mmPlus Ev.\mmPlus Time  &
\mmPart&\mmYes&\mmPart&\mmNo&\mmNo&\mmNo&\mmYes&\mmYes&\mmYes\\

StreamMemBench~\citep{streammembench2026} & Real  & QA\mmSlash Act.\mmSlash Learn  & Ans.\mmPlus Act.  &
\mmYes&\mmYes&\mmYes&\mmNo&\mmNo&\mmYes&\mmYes&\mmYes&\mmYes\\

S-EMBER~\citep{sember2026} & Real  & QA/Ret.  & Ans.\mmPlus Ev.\mmPlus Time  &
\mmPart&\mmYes&\mmPart&\mmNo&\mmNo&\mmNo&\mmYes&\mmYes&\mmYes\\

SLVMBench~\citep{slvmbench2026} & Real  & QA\mmSlash Learn  & Ans.+Time  &
\mmPart&\mmYes&\mmPart&\mmNo&\mmNo&\mmYes&\mmYes&\mmYes&\mmYes\\

StreamArena~\citep{streamarena2026} & Real  & QA\mmSlash Act.  & Ans.\mmPlus Time\mmPlus Act.  &
\mmYes&\mmYes&\mmPart&\mmNo&\mmNo&\mmYes&\mmYes&\mmYes&\mmYes\\

EgoMonth~\citep{egomonth2026} & Real  & QA/Ret.  & Ans.+Time  &
\mmPart&\mmYes&\mmYes&\mmNo&\mmNo&\mmNo&\mmYes&\mmYes&\mmYes\\

VideoWebArena\mmCrossDomain~\citep{videowebarena2025} & Hyb. & QA/Act. & Ans.\mmPlus State &
\mmPart&\mmYes&\mmPart&\mmNo&\mmNo&\mmYes&\mmPart&\mmYes&\mmYes\\

\end{longtable}


\begin{longtable}{@{}>{\raggedright\arraybackslash}p{6.42cm}>{\raggedright\arraybackslash}p{1.02cm}>{\raggedright\arraybackslash}p{1.83cm}>{\raggedright\arraybackslash}p{2.21cm}*{9}{>{\centering\arraybackslash}p{0.52cm}}@{}}
\caption*{\textbf{(c) Audio and speech memory}}\\[-2pt]
\MMBenchHeader{Audio}{SP}{AC}{RP}
\endfirsthead
\caption*{\textbf{(c) Audio and speech memory (continued)}}\\[-2pt]
\MMBenchHeader{Audio}{SP}{AC}{RP}
\endhead
\midrule
\multicolumn{13}{r}{\footnotesize Continued on next page}\\
\endfoot
\bottomrule
\endlastfoot

H2HMem~\citep{h2hmem2026} & Hyb.  & QA\mmSlash Learn  & Ans.\mmPlus Ev.  &
\mmYes&\mmYes&\mmYes&\mmYes&\mmYes&\mmNo&\mmPart&\mmNo&\mmPart\\

ChatMem-Bench~\citep{voicemem2026} & Syn. & QA/Gen. & Ans. &
\mmYes&\mmYes&\mmYes&\mmPart&\mmYes&\mmNo&\mmYes&\mmYes&\mmPart\\

PlanRAG-Audio~\citep{someki2026planragaudio} & Real & QA/Ret. & Ans.+Ev.+Time &
\mmYes&\mmYes&\mmNo&\mmNo&\mmNo&\mmNo&\mmYes&\mmYes&\mmNo\\

LongAudioQA~\citep{tang2026longaudioqa} & Real & QA/Ret. & Ans.+Ev.+Time &
\mmYes&\mmYes&\mmNo&\mmNo&\mmNo&\mmNo&\mmYes&\mmYes&\mmNo\\

ChronosAudio~\citep{chronosaudio2026} & Hyb. & QA/Class. & Ans. &
\mmPart&\mmYes&\mmNo&\mmNo&\mmNo&\mmNo&\mmPart&\mmYes&\mmNo\\

AudioMarathon~\citep{audiomarathon2025} & Hyb. & QA/Reason. & Ans.+Eff. &
\mmPart&\mmYes&\mmNo&\mmNo&\mmNo&\mmNo&\mmPart&\mmYes&\mmNo\\

Agentic ASR evaluation~\citep{agenticasr2026} & Hyb. & Gen./Rev. & Text+Sem. &
\mmYes&\mmPart&\mmYes&\mmYes&\mmNo&\mmNo&\mmNo&\mmPart&\mmYes\\

Stateful streaming ASR~\citep{statefulasr2026} & Real & Gen./State & Text+WER &
\mmNo&\mmPart&\mmPart&\mmNo&\mmNo&\mmNo&\mmPart&\mmPart&\mmNo\\

EgoMem evaluation~\citep{egomem2025} & Hyb. & Ret.\mmSlash Gen.\mmSlash Learn & ID+Fact+Resp. &
\mmYes&\mmYes&\mmYes&\mmNo&\mmNo&\mmPart&\mmYes&\mmPart&\mmPart\\

CarMem (text-only)~\citep{kirmayr2025carmem} & Syn. & Ret./Learn & Pref.+Ret. &
\mmNo&\mmYes&\mmYes&\mmYes&\mmNo&\mmNo&\mmNo&\mmNo&\mmPart\\

\end{longtable}


\begin{longtable}{@{}>{\raggedright\arraybackslash}p{6.42cm}>{\raggedright\arraybackslash}p{1.02cm}>{\raggedright\arraybackslash}p{1.83cm}>{\raggedright\arraybackslash}p{2.21cm}*{9}{>{\centering\arraybackslash}p{0.52cm}}@{}}
\caption*{\textbf{(d) GUI memory}}\\[-2pt]
\MMBenchHeader{GUI}{GS}{AO}{XR}
\endfirsthead
\caption*{\textbf{(d) GUI memory (continued)}}\\[-2pt]
\MMBenchHeader{GUI}{GS}{AO}{XR}
\endhead
\midrule
\multicolumn{13}{r}{\footnotesize Continued on next page}\\
\endfoot
\bottomrule
\endlastfoot

EMemBench~\citep{emembench2026} & Sim.  & QA/Act.  & Ans.+State  &
\mmPart&\mmYes&\mmYes&\mmNo&\mmYes&\mmYes&\mmPart&\mmYes&\mmNo\\

MemGUI-Bench~\citep{memguibench2026} & Real  & Act.  & State+Act.  &
\mmPart&\mmYes&\mmYes&\mmNo&\mmNo&\mmYes&\mmYes&\mmYes&\mmYes\\

KnowU-Bench~\citep{knowubench2026} & Hyb.  & Act./Plan  & State+Act.  &
\mmYes&\mmYes&\mmYes&\mmNo&\mmNo&\mmYes&\mmYes&\mmPart&\mmYes\\

MementoGUI-Bench~\citep{mementogui2026} & Real  & Act.\mmSlash Learn  & Reg.+Act.  &
\mmYes&\mmYes&\mmPart&\mmNo&\mmNo&\mmYes&\mmYes&\mmYes&\mmYes\\

Memory-World~\citep{wang2026stamp} & Sim.  & Act.  & State+Act.  &
\mmPart&\mmYes&\mmYes&\mmNo&\mmNo&\mmYes&\mmYes&\mmYes&\mmPart\\

DMV-Bench~\citep{dmvbench2026} & Sim.  & Ret.\mmSlash Act.  & State\mmPlus Act.\mmPlus Prov.  &
\mmPart&\mmYes&\mmPart&\mmNo&\mmNo&\mmYes&\mmYes&\mmPart&\mmYes\\

DataScope~\citep{atmem2026} & Hyb. & Act. & State &
\mmPart&\mmYes&\mmYes&\mmNo&\mmNo&\mmYes&\mmYes&\mmYes&\mmNo\\

\end{longtable}


\begin{longtable}{@{}>{\raggedright\arraybackslash}p{6.42cm}>{\raggedright\arraybackslash}p{1.02cm}>{\raggedright\arraybackslash}p{1.83cm}>{\raggedright\arraybackslash}p{2.21cm}*{9}{>{\centering\arraybackslash}p{0.52cm}}@{}}
\caption*{\textbf{(e) Embodied memory}}\\[-2pt]
\MMBenchHeader{Embodied}{SM}{OS}{TS}
\endfirsthead
\caption*{\textbf{(e) Embodied memory (continued)}}\\[-2pt]
\MMBenchHeader{Embodied}{SM}{OS}{TS}
\endhead
\midrule
\multicolumn{13}{r}{\footnotesize Continued on next page}\\
\endfoot
\bottomrule
\endlastfoot

LongSpace-Bench~\citep{longspace2026} & Real  & QA  & Ans.+Time  &
\mmPart&\mmYes&\mmYes&\mmNo&\mmNo&\mmNo&\mmYes&\mmYes&\mmPart\\

FindingDory~\citep{findingdory2025} & Sim.  & QA\mmSlash Loc.\mmSlash Act.  & Ans.\mmPlus State\mmPlus Act.  &
\mmPart&\mmYes&\mmPart&\mmNo&\mmNo&\mmYes&\mmYes&\mmYes&\mmPart\\

3DMem-Bench~\citep{threedllmmem2025} & Sim.  & QA\mmSlash Plan\mmSlash Act.  & Ans.\mmPlus State\mmPlus Act.  &
\mmYes&\mmYes&\mmPart&\mmNo&\mmNo&\mmYes&\mmYes&\mmYes&\mmPart\\

SEER-Bench~\citep{seerbench2025} & Sim.  & QA\mmSlash Plan\mmSlash Act.  & Ans.\mmPlus State\mmPlus Act.  &
\mmYes&\mmYes&\mmYes&\mmNo&\mmNo&\mmYes&\mmYes&\mmYes&\mmYes\\

LMEE-Bench~\citep{lmeebench2026} & Sim.  & QA\mmSlash Act.  & Ans.\mmPlus State\mmPlus Act.  &
\mmYes&\mmYes&\mmPart&\mmNo&\mmNo&\mmYes&\mmYes&\mmYes&\mmYes\\

RMBench~\citep{rmbench2026} & Sim.  & Act.  & State+Act.  &
\mmPart&\mmYes&\mmYes&\mmNo&\mmNo&\mmYes&\mmPart&\mmYes&\mmYes\\

RoboMME~\citep{robomme2026} & Hyb.  & Act.  & State+Act.  &
\mmYes&\mmYes&\mmYes&\mmNo&\mmNo&\mmYes&\mmYes&\mmYes&\mmYes\\

SpaMEM~\citep{spamem2026} & Sim.  & QA\mmSlash Act.  & Ans.\mmPlus State\mmPlus Act.  &
\mmPart&\mmYes&\mmYes&\mmNo&\mmNo&\mmYes&\mmYes&\mmYes&\mmPart\\

RoboMemArena~\citep{robomemarena2026} & Hyb.  & Act.\mmSlash Learn  & State+Act.  &
\mmYes&\mmYes&\mmYes&\mmNo&\mmNo&\mmYes&\mmYes&\mmYes&\mmYes\\

WorldMemArena\mmCrossDomain~\citep{worldmemarena2026} & Hyb.  & QA\mmSlash Act.\mmSlash Learn  & Ev.\mmPlus State\mmPlus Act.  &
\mmYes&\mmYes&\mmYes&\mmNo&\mmNo&\mmYes&\mmYes&\mmYes&\mmYes\\

eMEM-Bench v1~\citep{emembench_embodied2026} & Sim.  & QA/Ret.  & Ans.+State  &
\mmPart&\mmYes&\mmPart&\mmNo&\mmNo&\mmNo&\mmYes&\mmPart&\mmPart\\

ReMemBench~\citep{remembench2026} & Sim.  & Act.  & State+Act.  &
\mmPart&\mmYes&\mmYes&\mmNo&\mmNo&\mmYes&\mmPart&\mmYes&\mmYes\\

WorldLines (semantic states)~\citep{worldlines2026} & Sim.  & QA\mmSlash Plan  & Ans.\mmPlus State\mmPlus Act.  &
\mmNo&\mmYes&\mmYes&\mmNo&\mmNo&\mmYes&\mmYes&\mmYes&\mmYes\\

RoboMME-Interference~\citep{robommeinterference2026} & Sim.  & Act.  & State+Act.  &
\mmPart&\mmYes&\mmYes&\mmNo&\mmNo&\mmYes&\mmPart&\mmYes&\mmYes\\

MEMORA-Bench~\citep{memora2026} & Real & QA\mmSlash Plan & Ans.\mmPlus Act. &
\mmPart&\mmYes&\mmPart&\mmNo&\mmYes&\mmYes&\mmPart&\mmPart&\mmPart\\

APRS~\citep{panoseeker2026} & Hyb. & Loc.\mmSlash Act.\mmSlash Seg. & Reg. &
\mmPart&\mmYes&\mmPart&\mmNo&\mmNo&\mmYes&\mmYes&\mmPart&\mmPart\\

SIS-Bench~\citep{sisbench2026} & Real & QA\mmSlash Plan & Ans. &
\mmPart&\mmYes&\mmPart&\mmNo&\mmNo&\mmPart&\mmYes&\mmPart&\mmPart\\

EmbodiedWorldBench~\citep{tian2026abot} & Sim.  & Act.\mmSlash Learn  & Ev.\mmPlus State\mmPlus Act.  &
\mmYes&\mmYes&\mmYes&\mmNo&\mmNo&\mmYes&\mmYes&\mmYes&\mmYes\\

RoboGraph benchmark~\citep{robograph2026} & Sim.  & Plan/Act.  & State+Act.  &
\mmYes&\mmYes&\mmYes&\mmNo&\mmNo&\mmYes&\mmYes&\mmYes&\mmYes\\

DunphyBench~\citep{li2026memento} & Sim.  & Plan/Act.  & Ev.+Act.  &
\mmYes&\mmYes&\mmPart&\mmNo&\mmNo&\mmYes&\mmYes&\mmPart&\mmYes\\

HIOcc~\citep{hiocc2026} & Real & Map & Reg.\mmPlus State &
\mmPart&\mmYes&\mmPart&\mmNo&\mmNo&\mmNo&\mmYes&\mmPart&\mmNo\\

LIBERO-Mem~\citep{slotssm2026} & Sim. & Act. & Reg.\mmPlus State\mmPlus Act. &
\mmPart&\mmYes&\mmYes&\mmNo&\mmNo&\mmYes&\mmYes&\mmYes&\mmYes\\

MIKASA-Robo (RGB+joints)~\citep{cherepanov2025mikasa} & Sim. & Act. & State\mmPlus Act. &
\mmPart&\mmYes&\mmYes&\mmNo&\mmNo&\mmYes&\mmYes&\mmYes&\mmYes\\*

\multicolumn{13}{@{}p{\dimexpr16.16cm+24\tabcolsep\relax}@{}}{%
\footnotesize
\setlength{\parskip}{0.25em}
\rule{\linewidth}{0.4pt}\par
\textbf{Table notes}\par
\textbf{Legend.} \mmYes~= directly evaluated; \mmPart~= partial or indirect coverage; \mmNo~= not evaluated. A check indicates that the benchmark contains a scored task for which the corresponding memory demand is necessary to produce the correct answer or action. A double dagger (\mmCrossDomain) marks a cross-domain benchmark listed under its dominant setting.

\textbf{Dataset setup.} Source: real, synthetic (Syn.), simulated (Sim.), or hybrid (Hyb.). Task: question answering (QA), retrieval (Ret.), localization (Loc.), generation (Gen.), refusal or calibration (Ref.), action (Act.), planning (Plan), learning or transfer (Learn), and mapping (Map). Ground truth: answers (Ans.), evidence (Ev.), regions (Reg.), timestamps (Time), state (State), actions (Act.), provenance (Prov.), and reliability labels (Rel.).

\textbf{Memory capabilities evaluated.} CE: cross-modal evidence; LE: long-range evidence distributed across turns, sessions, frames, or events; SC: state changes; CF: conflicting memories; UQ: unanswerable queries; MA: memory-grounded action.

\textbf{Modality-specific capabilities.} Image: FG, fine-grained visual-detail retention; ET, entity identity tracking across observations; VC, visual state-change tracking. Video: TL, temporal evidence localization; TR, temporal-relation reasoning, including event order and duration; ME, multi-event evidence integration. Audio: SP, speaker tracking; AC, acoustic-cue retention; RP, repair persistence. GUI: GS, GUI-state tracking; AO, action-outcome tracking; XR, cross-session experience reuse. Embodied: SM, spatial memory; OS, object-state tracking under partial observability; TS, task-state and progress tracking.
\par}\\
\end{longtable}

\setcounter{table}{\MMBenchSavedTableCounter}

\endgroup

\begingroup
\setlength{\LTleft}{\fill}
\setlength{\LTright}{\fill}
\setlength{\LTcapwidth}{\textwidth}
\setlength{\LTpre}{0.35em}
\setlength{\LTpost}{0.55em}
\setlength{\tabcolsep}{1pt}
\renewcommand{\arraystretch}{1.06}

\newcommand{\MMRelatedWorkHeader}{%
\toprule
\multicolumn{1}{>{\centering\arraybackslash}m{2.5cm}}{\textbf{Work}} &
\multicolumn{1}{>{\centering\arraybackslash}p{2.5cm}}{\textbf{Scope}} & 
\multicolumn{1}{>{\centering\arraybackslash}p{2.5cm}}{\textbf{Modalities}} & 
\multicolumn{1}{>{\centering\arraybackslash}p{5cm}}{\textbf{Applications}} &
\multicolumn{1}{>{\centering\arraybackslash}m{1.8cm}}{\textbf{Mem.\& Rep.}} &
\multicolumn{1}{>{\centering\arraybackslash}m{1.5cm}}{\textbf{Mem.\& Mgt.}} \\
\midrule
}

\captionsetup{hypcap=false}
\Needspace{10\baselineskip}
\captionof{table}{Comparison of our framework with related work on agent and multimodal memory.}\label{tab:related-work-comparison}
\captionsetup[longtable]{labelformat=empty,justification=centering,singlelinecheck=true,font=normalsize,skip=2pt}
\scriptsize

\begin{longtable}{@{}
>{\raggedright\arraybackslash}m{2.5cm}
>{\raggedright\arraybackslash}p{2.5cm}
>{\raggedright\arraybackslash}p{2.5cm}
>{\raggedright\arraybackslash}p{5cm}
>{\raggedright\arraybackslash}m{1.8cm}
>{\raggedright\arraybackslash}m{1.5cm}
@{}}
\MMRelatedWorkHeader
\endfirsthead

\MMRelatedWorkHeader
\endhead

\midrule
\multicolumn{6}{r}{\footnotesize Continued on next page}\\ 
\endfoot

\bottomrule
\endlastfoot

Xie et al.\\~\citep{xie2024largemultimodalagents} & Multimodal agents & Text/\allowbreak Image/\allowbreak Video/\allowbreak Audio & Multimodal Understanding \& Generation; GUI/Web; Embodied/Robotics & S+E & T1--T2 \\
Zhang et al.\\~\citep{zhang2025memorymechanisms} & LLM-agent memory & Text & Dialogue \& Personalization; Social/Game Agents; Task-oriented Agents & E+L/P & T1--T2 \\
Du et al.\\~\citep{du2025rethinkingmemory} & LLM-agent memory & Text/\allowbreak Image/\allowbreak Video/\allowbreak Audio & QA/RAG; Dialogue \& Personalization; Tool/Embodied Agents & S+E+L/P & T1--T3 \\
Hu et al.\\~\citep{hu2026memoryageaiagents} & Foundation-model agent memory & Text/\allowbreak Image/\allowbreak Video/\allowbreak Audio/\allowbreak Embodied & QA; Personalization; Social/Embodied Agents; Domain-specific Agents & S+E+L/P (+H) & T1--T3 \\
Jia et al.\\~\citep{jia2025aihippocampus} & LLM/MLLM memory & Text/\allowbreak Image/\allowbreak Video/\allowbreak Audio & Multimodal Understanding \& Generation; Dialogue \& Personalization; GUI/Embodied Agents & S+E+L/P (+H) & T1--T2 \\
Xia et al.\\~\citep{xia2026memoryvlm} & VLM memory & Text/\allowbreak Image/\allowbreak Video & Multimodal Understanding \& Generation; Dialogue \& Personalization; Embodied/Robotics & S+E+L/P (+H) & T1--T2 \\
Luo et al.\\~\citep{luo2026storageexperience} & Agent memory evolution & Text/\allowbreak Image/\allowbreak Video/\allowbreak Audio & Dialogue; Planning \& Tool Use; Web/Embodied Agents & S+E+L/P (+H) & T1--T3 \\
Mokaria et al.\\~\citep{mokaria2026multimodalagentic} & Multimodal-agent frameworks & Text/\allowbreak Image/\allowbreak Video/\allowbreak Audio & Multimodal Understanding \& Generation; GUI/Web; Embodied/Robotics & S+E+L/P (+H) & T1--T2 \\
Ours & Multimodal-agent memory & Text/\allowbreak Image/\allowbreak Video/\allowbreak Audio/\allowbreak Embodied & Reasoning \& Recall; Personalization; GUI/Web; Embodied/Robotics; Generation & S+E+L/P (+H) & T1--T3 \\
\end{longtable}

\vspace{-0.4em}
\begin{minipage}{\linewidth}
\footnotesize
\textbf{Legend.} Modalities and applications denote areas receiving substantive discussion rather than isolated citations. Representation families and memory-management tiers are mapped to our taxonomy for consistent comparison. S: source- and modality-specific representations; E: cross-represented explicit memory; L/P: internalized latent or parametric memory; H: hybrid or source-linked composition.

\end{minipage}

\endgroup

\subsubsection{Image Memory}

Image benchmarks typically treat memory as a collection of visual episodes associated with dialogue, time, location, or personal context.
Their horizons range from multi-turn image sequences to persistent personal archives, with question answering and retrieval as the dominant tasks.
These resources differ in whether evaluation uses images directly or substitutes textual descriptions, as well as in the stages of the memory lifecycle they exercise.

LoCoMo~\citep{maharana2024evaluating} evaluates long-term conversational memory through question answering, event summarization, and multimodal dialogue generation. Its conversations include shared images, but the original QA and summarization experiments replace images with captions; images are used directly only in dialogue generation.

Evidence-centered benchmarks provide the clearest tests of visual persistence. 
Memory-QA~\citep{memoryqa2025} evaluates recall over real-world multimodal records, requiring temporal or location cues and evidence distributed across multiple memories. 
DISBench~\citep{deepimagesearch2026} instead evaluates context-aware retrieval and localization within interconnected visual histories, while MemLens~\citep{memlens2026} stresses whether fine-grained visual evidence survives long-term storage and retrieval. 

Personalized and conversational benchmarks extend evaluation from evidence access to memory validity. 
LCMP~\citep{hong2026tame} and ATM-Bench~\citep{atmbench2026} distribute user-specific attributes and references across multiple sources and sessions, requiring the agent to track both persistent and changing
information. 
MemEye~\citep{memeye2026} evaluates image-grounded histories and visual state changes, while Mem-Gallery~\citep{memgallery2026} and H2HMem~\citep{h2hmem2026} additionally test conflicting records and insufficient evidence.
These benchmarks test whether retrieved memories remain current and mutually consistent, although updates are usually defined by curated histories rather than by an evolving memory-management policy. MemLeak~\citep{wang2026memleakdiagnosinginformationleaks} examines a complementary requirement: whether facts requested for deletion remain recoverable from retained text or images.

A group connects image memory to action or source verification.
SMMBench~\citep{smmbench2026} distributes evidence across conversations, screenshots, images, tables, and documents, and provides answer, evidence, and provenance supervision for memory-grounded decisions. 
EMemBench~\citep{emembench2026} similarly introduces state or action consequences, but such supervision remains uncommon. 
Conversational benchmarks span broader dialogue evaluation and explicit memory tasks: MMDU~\citep{liu2024mmdu} evaluates multi-image, multi-turn response quality, whereas MMRC~\citep{xue2025mmrc} includes memory recall, image management, and information updates. Their aggregate scores nevertheless combine memory with other dialogue capabilities. Across these settings, supervision remains predominantly answer- and evidence-based, while provenance, state, reliability, and action labels appear only in a smaller subset. 
Current image evaluation therefore covers fine-grained evidence retention, entity continuity, and visual state change more thoroughly than extended interaction or adaptive memory use.
Provenance and behavioral verification also remain comparatively underdeveloped.

\subsubsection{Video Memory}

Video-memory evaluation has expanded from offline retrieval over completed recordings to causally constrained streaming and interactive assistance, but it remains dominated by readout-style question answering. We group current benchmarks into three evaluation regimes: fixed-recording or lifelog recall, causal streaming evaluation, and future-oriented interactive assistance. Although all three require access to past video, they differ in whether evidence is available as a completed record, arrives incrementally, or must influence subsequent interaction.

Fixed-recording and lifelog benchmarks provide the clearest tests of long-range temporal access. M3-Bench~\citep{long2025m3agent} and HippoVlog~\citep{hippomm2025} cover extended audiovisual streams, while EgoLifeQA~\citep{egolife2025} uses week-long multimodal egocentric records to support event recall, personalized question answering, and recommendation. EgoMonth~\citep{egomonth2026} extends the horizon to month-scale experience, and EgoMem~\citep{videolucy2025} emphasizes backtracking through long videos. EgoMemReason~\citep{egomemreason2026} further separates entity, event, and behavioral memory. LVOmniBench~\citep{tao2026lvomnibench} broadens this setting beyond egocentric data by requiring joint reasoning over long-form audio and video. These benchmarks are strong tests of temporal localization and long-range evidence retrieval, with answers established primarily through timestamps or supporting evidence, but they usually assume that the relevant video history is already available as a completed record.

Streaming evaluation introduces causal evidence availability. VStream-QA~\citep{flashvstream2025} and StreamingBench~\citep{streamingbench2024} ask questions while the video continues to arrive, whereas StreamBench~\citep{xiong2025streamchat} adds multi-round interaction over hierarchically maintained short- and long-term records. Causal access makes these benchmarks closer to deployment, but does not by itself establish that a model maintains or revises a persistent memory state. Multi-view benchmarks add a distinct challenge: memory must align temporally related evidence across agents, cameras, or viewpoints rather than retrieve from a single chronological stream. MA-EgoQA~\citep{kim2026maegoqa} constructs a shared history from synchronized egocentric streams and requires the model to identify and combine agent-specific evidence, while EgoExoMem~\citep{egoexomem2026} introduces cross-view temporal and spatial dependencies. These settings therefore test distributed perspective and evidence fusion, not merely access to a longer video, although they still treat memory chiefly as a retained record to be queried rather than as a state that must be actively maintained.

A smaller group evaluates whether video memory supports future behavior. VideoWebArena~\citep{videowebarena2025} tests factual and skill retention from tutorials during subsequent web tasks, scoring intermediate answers and final environment states. Because video-derived information can remain in context, these scores do not isolate memory-management quality. RIVER Bench~\citep{riverbench2026} and StreamArena~\citep{streamarena2026} combine historical recall with live or proactive responses under real-time constraints, while StreamMemBench~\citep{streammembench2026} tests whether observed evidence, feedback, and prior interaction are reused in a later task. SLVMBench~\citep{slvmbench2026} further asks agents to extract procedural knowledge from recorded experience and apply it to a live stream. These tasks move beyond retrospective recall toward stateful assistance, but still mainly test whether prior experience is reused, not whether deployment feedback changes the memory-management policy itself.

\begin{figure}[H]
    \centering
    \includegraphics[width=\textwidth]{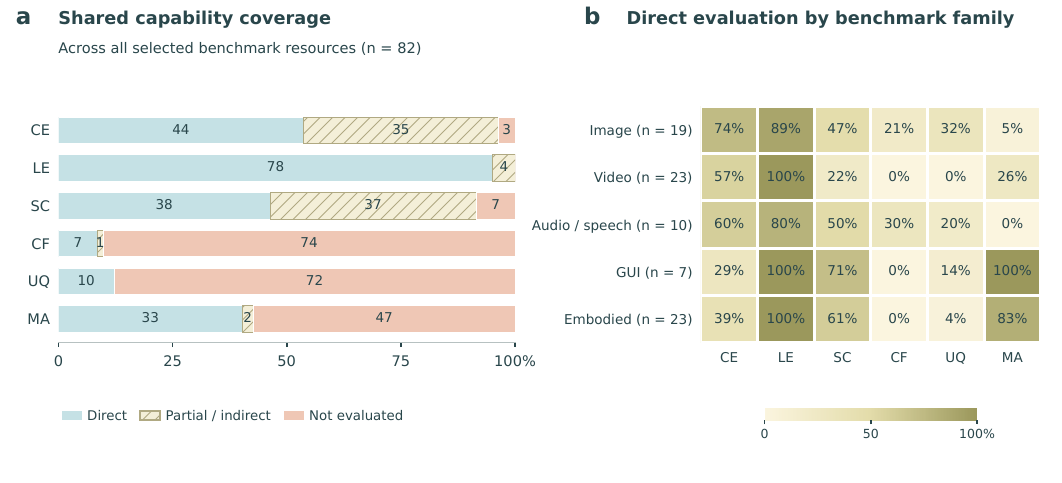}
    \caption{\textbf{Coverage of shared memory capabilities across selected evaluation resources.}
    \textbf{a,} Distribution of direct, partial, and not-evaluated coverage across all \BenchmarkCount{} benchmarks in \Cref{tab:mm-memory-benchmarks}; labels within the bars report benchmark counts.
    \textbf{b,} Percentage of benchmarks in each family that directly evaluate each capability; family sizes are shown in parentheses.
    CE, cross-modal evidence; LE, long-range evidence distributed across turns, sessions, frames, or events; SC, state changes; CF, conflicting memories; UQ, unanswerable queries; MA, memory-grounded action.}
    \label{fig:benchmark-capability-coverage}
\end{figure}

Across these settings, supervision remains predominantly answer-, evidence-, and timestamp-based. Overall, video benchmarks evaluate temporal localization, long-range event relations, and multi-event integration more thoroughly than most other modalities. However, conflict resolution, unanswerable queries, and memory-grounded action remain less common in \Cref{fig:benchmark-capability-coverage}; source-level provenance and feedback-driven policy change are also weakly evaluated. This gap contrasts with the architectural trend in \Cref{fig:cross-matrix}: memory systems commonly update state, yet video benchmarks rarely verify whether deployment feedback produces persistent improvements in the memory-management policy.

\subsubsection{Audio and Speech Memory}

Audio and speech occupy a small but differentiated part of the benchmark landscape. Existing evaluations span persistent voice-grounded conversational memory, memory over streaming audiovisual experience, retrieval from a fixed long-form recording, and short-horizon speech state carried or revised across turns. 
These audio-based settings exercise different parts of the memory lifecycle; text-only preference protocols provide a separate comparison.

ChatMem-Bench, introduced with VoiceMem~\citep{voicemem2026}, evaluates long-horizon speech memory using 316 questions over 53 hours of dialogue. Human-curated synthetic histories support tasks covering information updates, abstention, persona, affective attribution, speaker association, and non-lexical acoustic evidence. Its published category scores support direct SC, UQ, SP, and AC coverage, but only indirect CF and RP coverage: the construction pipeline includes contradictions, while a dedicated persistent spoken-repair protocol is not specified. The paper defers detailed annotation methodology to a follow-up report.
EgoMem~\citep{egomem2025} instead combines face and voice identity with asynchronously managed facts, preferences, and social relations for later personalized dialogue.

Streaming audiovisual benchmarks broaden the horizon but make audio attribution less clean. 
M3-Bench, TeleEgo, and Video-SALMONN~S retain or evaluate audiovisual experience over long streams, yet their video-centred tasks rarely isolate speaker, prosodic, or non-speech acoustic retention~\citep{long2025m3agent,teleego2025,sun2026videosalmonnsmemoryenhancedstreaming}. 
Fixed-recording benchmarks provide more controlled audio evidence: PlanRAG-Audio and LongAudioQA evaluate planned retrieval from long recordings, while ChronosAudio and AudioMarathon stress length generalization and temporal reasoning~\citep{someki2026planragaudio,tang2026longaudioqa,chronosaudio2026,audiomarathon2025, gong25c_interspeech}. 
Because their memory remains fixed after formation, these settings primarily evaluate access and reuse in T1 Static Memory rather than cross-session memory updating.

Interactive ASR supplies a narrower form of state change. 
Agentic ASR evaluates whether spoken feedback repairs a transcription state, whereas Beyond Fresh Starts carries or rolls back acoustic and linguistic state across utterance boundaries~\citep{agenticasr2026,statefulasr2026}. 
CarMem~\citep{kirmayr2025carmem} provides a text-only comparison for preference extraction, retrieval, and maintenance. Its voice-assistant application does not make these experiments tests of acoustic memory. 
Current benchmarks therefore cover acoustic retention, long-range retrieval, speaker association, and repair separately, but rarely combine cross-session acoustic memory, conflicting spoken evidence, explicit revision, and memory-grounded action.

\subsubsection{GUI Memory}

GUI benchmarks occupy a compact but functionally differentiated part of the evaluation landscape. Existing settings span cross-step state retention, cross-session experience reuse, personalized and proactive interaction, and controlled diagnosis of visual memory. 
All reuse prior interface experience, but differ in whether memory must preserve a transient state, recover an action outcome, transfer a workflow, or retain visual evidence unavailable from text alone.

MemGUI-Bench~\citep{memguibench2026} provides the broadest direct evaluation of mobile-agent memory. Its tasks span multiple applications and use mirrored interactions to separate within-session retention from cross-session improvement, while hierarchical evaluation measures memory fidelity, learning effectiveness, and execution efficiency. 
MementoGUI-Bench~\citep{mementogui2026} instead formulates long-horizon GUI execution as a memory-control problem: working memory preserves task-relevant interface events and localized visual evidence, whereas episodic memory retrieves reusable trajectories. 
Together, these benchmarks evaluate GUI-state tracking, action--outcome associations, and experience reuse through regions, task progress, and action supervision.

Personalized and controlled settings test complementary aspects of state maintenance. KnowU-Bench~\citep{knowubench2026} hides the user profile and requires agents to infer preferences from behavioral histories, request missing information, and calibrate proactive intervention in an interactive mobile environment. 
Its task scores evaluate the use of history in personalized assistance, while also depending on preference inference, interaction, and execution.
Memory-World~\citep{wang2026stamp}, introduced with STAMP, programmatically controls what transient interface information must be stored and later reused, enabling state- and action-verified evaluation in a simulated environment. 
DataScope~\citep{atmem2026} combines synthetic task data with real mobile applications in 96 cross-page and cross-application workflows. Terminal success, app-level progress, and Scope-F1 verify which required operations were completed and which distractors were modified from the resulting application states; they do not provide expert action-sequence ground truth or evaluate cross-session reuse. Such benchmarks improve experimental control, although their synthetic states do not fully reproduce the variability of real applications.

DMV-Bench~\citep{dmvbench2026} isolates visual memory more directly by injecting pixel-specific incidental cues into products visited across autonomous shopping sessions. Textual descriptions cannot recover the relevant cue, and successful navigation therefore depends on retaining and retrieving visual evidence with its source provenance. 
Current GUI evaluation consequently provides strong coverage of long-range state tracking, action outcomes, and memory-grounded execution. 
It remains weaker on conflicting histories, unanswerable or unsafe requests, robustness to interface change, and persistent evolution of memory-management policies. 
Moreover, endpoint task success alone may still fail to identify whether an error arose during memory formation, retrieval, or action selection.

\subsubsection{Embodied Memory}

Embodied-memory benchmarks assess whether agents can retain and use experience while navigating, mapping, and manipulating partially observable environments. Because an agent's actions affect both the environment and the observations available later, memory must preserve spatial structure, maintain object and task states, and associate actions with their consequences. Existing benchmarks can be organized into three complementary regimes: navigation and mapping, memory-dependent manipulation, and long-horizon stateful interaction.

Navigation and mapping benchmarks primarily assess spatial persistence. FindingDory~\citep{findingdory2025} requires agents to retrieve prior visual experience for subsequent navigation and action in tasks that can be procedurally extended to longer horizons, whereas 3DMem-Bench~\citep{threedllmmem2025} evaluates long-term spatiotemporal question answering, scene captioning, and task execution in dynamic multi-room environments. SpaMEM~\citep{spamem2026} isolates action-conditioned revision of spatial beliefs, while HIOcc~\citep{hiocc2026} evaluates local occupancy prediction and causal room- and building-level mapping. HIOcc measures map accuracy, online stability, and revisit consistency in static scenes; these support direct spatial-memory coverage but only partial state-change and object-state coverage, without task-progress or downstream-action evaluation. FindingDory, 3DMem-Bench, and SpaMEM additionally assess object histories or updates. Spatial regions, maps, and environment states provide supervision beyond terminal task success.

Other embodied settings assess experience reuse through planning or active perception. MEMORA-Bench~\citep{memora2026} pairs recall over participant-specific egocentric histories with generated plans for observed and novel goals. Its explicit information-not-available controls directly evaluate UQ, while plan-order, object-coverage, and preference scores evaluate prospective memory use; spatial, object-state, and task-progress tracking are only indirectly covered by these aggregate tasks. APRS~\citep{panoseeker2026} evaluates sequential viewpoint selection and target segmentation within static panoramas, scoring search success, efficiency, and mask quality. Expert action trajectories support training rather than serving as benchmark action ground truth. SIS-Bench~\citep{sisbench2026} combines landmark and action recall with spatial reasoning and multiple-choice path planning over UAV videos. Its planning answers provide indirect MA coverage, not evidence of executed control or online task-progress tracking.

Manipulation benchmarks assess whether policies preserve action-relevant history. LIBERO-Mem~\citep{slotssm2026} introduces non-Markovian tasks that require object histories beyond the current observation, with object masks, instance identities, and subgoal annotations. MIKASA-Robo~\citep{cherepanov2025mikasa}, evaluated here in its simulated RGB+joints setting, tests delayed object recall, spatial tracking, and sequential action. RMBench~\citep{rmbench2026} and ReMemBench~\citep{remembench2026} provide controlled memory-dependent manipulation tasks. RoboMME~\citep{robomme2026} evaluates temporal, spatial, object, and procedural memory in vision--language--action policies, while RoboMemArena~\citep{robomemarena2026} extends evaluation to longer tasks with subtask and keyframe annotations and paired real-world tasks. State- and action-level supervision provides direct evidence for memory-grounded action (MA), although endpoint success cannot distinguish memory failures from errors in perception, planning, or control.

Long-horizon and cross-session benchmarks evaluate whether memory remains valid as world state and interaction history evolve. WorldLines~\citep{worldlines2026} pairs evidence-linked memory questions with embodied planning over semantic household traces; this protocol evaluates state maintenance without establishing raw visual or acoustic retention. WorldMemArena~\citep{worldmemarena2026} evaluates memory formation, maintenance, retrieval, and use across multi-session action--world interaction. RoboMME-Interference~\citep{robommeinterference2026} measures retrieval robustness as unrelated prior sessions accumulate. WorldLines and WorldMemArena therefore extend evaluation from access to T1 Static Memory toward maintenance and revision under T2 Dynamic Memory, whereas RoboMME-Interference isolates interference during access. Most current benchmarks nevertheless rely on simulated or hybrid environments, and few systematically test conflicting memories (CF), unanswerable queries (UQ), lifelong updating in physical environments, or failure attribution across memory-lifecycle stages.

\subsection{Application Regimes}
\label{sec:application-regimes}


Application regimes are distinguished here by the function memory serves rather than by input modality. 
They consolidate concrete downstream tasks---including multimodal recall, long-form video and audio understanding, personalized interaction, GUI automation, embodied control, and iterative generation---according to the role memory plays in the resulting system.
This view is orthogonal to Axis~I: the same images, audio, documents, or interaction traces may be retained to support evidence-based decisions, continuity with a user, execution in an environment, or consistency in generated content. 
Each regime consequently stresses a different property of memory.

\subsubsection{Recall and Evidence-Based Reasoning}

This regime encompasses multimodal question answering, episodic retrieval, and long-form video or audio understanding. 
Although these tasks operate over different evidence---from visual details and temporally distributed events to speakers and acoustic cues---they share the requirement that later outputs remain grounded in recoverable 
observations~\citep{memoryqa2025,egolife2025,someki2026planragaudio}.
Personal archives and long-document assistants require cross-modal retrieval and source grounding, whereas enterprise analysis additionally requires evidence to remain attributable across files, reporting periods, and tool outputs. 
MARDoc, for example, distils long-document interaction traces into structured evidence and reasoning memories and checks whether the recovered evidence is sufficient~\citep{mardoc2026}. 
Financial reasoning makes the temporal and numerical demands especially explicit: FinAcumen selectively reuses successful strategies and failure-derived cautions while coordinating retrieval, visual decoding, calculation, and verification over text, tables, charts, and time-indexed records~\citep{guo2026finacumen}. 
The central memory property is therefore not recall volume alone, but recoverable provenance and temporal validity, since persistence without these safeguards can retain stale or weakly grounded evidence.

\subsubsection{Personalization and Cross-Session Continuity}

Personalization uses memory to maintain a user-specific state across otherwise disconnected encounters. 
Relevant experience may include stated preferences, prior dialogue, viewed media, corrections, and recurring activity patterns. 
M$^2$A combines complementary memory layers for long-term multimodal interaction, TAME maintains a state-aware personalized context without additional training, and Persode grounds a visual-journaling assistant in episodic personal history~\citep{feng2026m2a,hong2026tame,jin2025persode}. 
In consumer settings, the resulting continuity supports recommendations and assistance; analogous requirements arise in enterprise workflows that carry user, team, or account context across sessions. 
More recently, LLM-based individual simulation has attracted growing interest. 
The pioneering Individual Turing Test highlights the value of memory in capturing personal opinions and preferences~\citep{guo2026individual}, which could be more faithful through multimodal personal histories beyond text.
This regime stresses identity scoping, selective retention, and revision: useful continuity requires the agent to associate evidence with the correct subject and to update preferences without allowing obsolete or cross-user information to persist.

\subsubsection{Control and Task Execution}

Control applications require memory to change what an agent does, not merely what it can report. 
They include GUI workflow automation, spatial navigation and mapping, embodied manipulation and planning, and the transfer of procedures learned from demonstrations or prior experience~\citep{wen2022youonlydemonstrate,lee2025delta,liang2022sensorimotor,mandlekar2023mimicgen,garrett2025skillmimicgen,hsu2025spot,li2026hydra0}.
MAGNET, for instance, evolves visual-semantic and procedural memories so that GUI skills transfer across changing interfaces~\citep{magnet2026}. 
In embodied settings, remembered layouts, object states, and task progress support navigation and manipulation under partial observability, while environment states and action outcomes provide direct evidence of memory utility~\citep{findingdory2025,robomme2026,worldmemarena2026,bjorck2026vesta,ma-etal-2025-pragmatics}.
The same principle extends to other domains.
FinAgent combines numerical, textual, and visual market evidence with diversified memory retrieval and dual-level reflection to adapt its decisions as market conditions change~\citep{zhang2024finagent}. 
Enterprise automation adds operational constraints and heterogeneous artifacts: Matrix refines experience-derived memory to specialize agents for recurring business-document extraction, while MemoPhishAgent retrieves episodic reasoning trajectories to orchestrate tools for recurring and novel phishing threats~\citep{liu2024matrix,chen2026memophish}. 
The defining property is action relevance under change: stored experience must improve the present decision while remaining revisable when interfaces, environments, document formats, market conditions, or threat patterns evolve.

\subsubsection{Generation and World Consistency}

Generative applications use memory to preserve constraints, entities, style, and causal state across iterative production. 
Unlike recall and control, however, this regime remains weakly represented by dedicated memory benchmarks: existing generation tasks commonly evaluate open-ended responses rather than image, video, or speech synthesis with persistent identity and world state. Application systems nevertheless demonstrate the underlying demand.
MemSlides separates profile, working, and tool memories to support personalized slide creation and local revision, while agentic video systems retain executable event structures to keep generated scenes aligned with an evolving plan~\citep{memslides2026,gest2026}. 
Financial reporting combines this consistency requirement with evidence grounding. FinSight's variable memory unifies financial data, tools, and intermediate analyses before synthesizing citation-aware narratives, tables, and charts~\citep{jin2026finsight}. 
Here memory is valuable because generation is not a single decoding pass: evidence is collected, transformed, assessed, visualized, and revised over multiple stages.
The regime therefore stresses versioned state and recoverable dependencies, so that a change to an underlying fact or analytical result can propagate coherently through the final multimodal artifact.

\subsection{Application–Evaluation Alignment and Gaps}
\label{sec:benchmark-coverage}

The benchmark landscape is broader in modality than in lifecycle coverage.
This pattern is partly determined by the downstream tasks used for evaluation.
Image, video, and audio benchmarks predominantly use question answering, retrieval, or localization, which expose recall and grounding but provide limited evidence of subsequent memory use.
GUI and embodied benchmarks instead evaluate action or planning through interface states, environment transitions, and task success
\citep{memguibench2026,findingdory2025,worldmemarena2026}.
The task therefore determines which aspects of memory can be observed, whereas the capability dimensions characterize the memory demands being evaluated.

This task dependence leads to uneven coverage across the application regimes in \Cref{sec:application-regimes}.
Long-range and cross-modal evidence access supports many important applications and, correspondingly, constitutes the dominant evaluation target across \Cref{tab:mm-memory-benchmarks}.
State tracking is increasingly considered in personalized, interactive, and embodied settings.
Conflict resolution and evidence insufficiency remain sparsely evaluated, however, and memory-grounded action is concentrated in GUI and embodied benchmarks.
Consequently, retrieval-centric applications are better represented in current benchmarks than applications that require persistent state maintenance, conflict handling, or memory-grounded action.
Current evaluation therefore establishes whether experience can be recovered more consistently than whether it remains valid, sufficient, and actionable.

Modality-specific evaluation is strongest for structure native to each input stream. 
Image benchmarks emphasize visual details, entities, and state changes \citep{memoryqa2025,memlens2026}; 
video benchmarks emphasize temporal localization and multi-event reasoning \citep{egolife2025,streammembench2026};
audio and speech benchmarks examine long-form retrieval, speaker- and acoustic-level retention, and stateful interaction \citep{someki2026planragaudio,egomem2025}.
GUI and embodied benchmarks instead focus on interface, spatial, object, and task states \citep{findingdory2025,worldmemarena2026}.
Audio evaluation nevertheless remains predominantly recall- or transcription-oriented, with limited coverage of persistent correction, personalized response, speaker conflict, abstention, and memory-grounded action.

Lifecycle coverage remains concentrated on persistence and retrieval in T1 Static Memory; state-update and experience-reuse tasks probe T2 Dynamic Memory, but usually under stable policies.
A Learn label therefore does not establish T3 Self-Evolving Memory unless deployment feedback produces persistent changes to a reusable memory-management policy
\citep{mementogui2026,streammembench2026}.
Thus, although current benchmarks cover important application-facing retrieval capabilities, their lifecycle coverage remains incomplete.
The main gaps lie in conflict and uncertainty handling, source provenance, end-to-end memory attribution, policy adaptation, and memory-grounded visual generation.
\section{Challenges and Outlook}
\label{sec:agenda}

The preceding sections show that long-term multimodal agent memory depends on three coupled concerns: what experience is retained and how it is represented, how that memory is maintained as new evidence arrives, and how its logical operations are realized under finite system resources. Their interaction becomes more consequential over long trajectories, changing environments, and shared deployments. We therefore distinguish challenges in formation and representation, challenges in memory management, and tensions between the two, followed by deployment-oriented evaluation and open infrastructure problems. The infrastructure discussion builds on \Cref{sec:memory-infrastructure} and focuses on how the assumptions and guarantees of the analyzed mechanisms can be integrated under sustained deployment.

\subsection{Multimodal Memory Formation and Representation}
\label{sec:open-formation}

Memory formation determines not only what information is retained, but also what
can still be recovered after the original experience has been transformed, what
can later be located and changed inside the resulting representation, and how
far any of this has been worked out beyond the modalities that dominate current
practice. Current systems use a wide range of source, explicit, latent, and
parametric representations, yet important questions remain about what each of
them should preserve, what structure each should expose, and whether the answers
established for visual experience carry over to other modalities.

\subsubsection{Recovering Supporting Evidence}
\label{sec:open-recoverability}

Abstraction should move from lossy compression toward recoverable
representation. As previous sections show,
modality-specific encodings, textual, structured, latent, and parametric
memories can all remove information that remains available in source-level
evidence. Source-linked designs introduced in
\Cref{sec:hybrid_representations} partly address this problem by preserving
links from derived memories back to supporting observations. The remaining
question is what information should be preserved during formation so that this
recoverability is reliable.

Existing systems use different combinations of timestamps~\citep{videoem2025}, source identifiers~\citep{feng2026m2a},
entity links~\citep{long2025m3agent}, spatial references~\citep{liu2025dynamem}, and pointers to supporting observations~\citep{hippomm2025}. However,
a substantial portion of the analyzed architectures do not retain a usable route
back to the evidence from which their memories were formed. Among those that do,
the route is often simply that the source was never abstracted away, rather than
a link deliberately carried by a derived memory, and where such links do exist
they differ in what they record. Provenance is therefore seldom a designed
property of the representation. A useful
direction is to define a minimal provenance bundle for formed memories,
containing enough information to identify where a memory came from, how it was
produced, and what evidence supports it. The required fields may differ across
modalities and representation types, but provenance should be treated as part of
the representation itself rather than as optional metadata.

Recoverable representation does not require every source observation to be
retained indefinitely. Instead, memory formation should make explicit which
information may be discarded and which supporting evidence must remain reachable
for later inspection or reinterpretation. This would make abstraction more
controlled without requiring all downstream memories to preserve the full
original experience. What remains open is how to make this decision. Current systems use different rules~\citep{visor2026, hamvln2026}, but a better criterion should balance retention cost against the need to verify or correct information later.

\subsubsection{Finding and Updating Specific Memory}
\label{sec:open-addressability}

Formed memory should expose the units that memory management acts on. While recoverability asks whether a memory can be traced back to its supporting evidence, semantic addressability asks whether specific content can be identified and changed within the representation itself. Explicit textual and structured records provide such units by
construction, which is why they support revision and conflict resolution most
directly~\citep{vilomem2025}. Semantic addressability remains rare outside these representations. The difficulty is sharpest in latent and parametric memory, which is
compact and integrates directly with the model's computation but does not expose
content as separately identifiable units by construction. An update to such
memory may therefore reach information it was not intended to change, and the system may
not be able to identify which part of memory is responsible for a particular
behaviour.

This limitation appears in the corpus as a pattern rather than only as an
argument. \Cref{sec:cross-axis-corpus} shows that every architecture evolving a
memory-management policy at deployment time maintains an explicit
cross-represented component, and that none whose memory is only latent or only
parametric does so. The absence is not a matter of these systems being static:
most latent-only architectures do update their memory state under a fixed
policy. What they lack is not dynamism, but clearly identifiable units over
which memory-management decisions can be attributed, inspected, and revised.
Source-preserving memory shows the same absence despite being addressable in
another sense, and the contrast is instructive: its units can be identified
perceptually, by time or position, yet still not at the semantic level at which
a write or retention decision is made. The obstacle is therefore not distributed
representation as such, but the absence of units that correspond to what a
policy decides about.

Parametric memory can already be divided into separable modules associated with
different experiences, modalities, or tasks, as in adapter banks that hold
per-task procedures~\citep{vlapro2026}, per-video parameters~\citep{video2lora2026}, or modality-specific edits~\citep{memeic2025}. This makes parts of a
distributed representation easier to locate and manipulate, but it has not so
far produced semantic addressability: among the analyzed systems that organize
memory this way, the few that evolve a management policy during deployment do so
through an explicit cross-represented component rather than through the
parameter store itself. Structured latent memory that binds persistent state to identifiable slots or
entities~\citep{slotssm2026} would be a further step in the same direction. What a fuller answer
requires is nonetheless clear: a distributed memory is addressable in the sense
that management needs only when the same unit can be identified, revised on its
own, and afterwards verified to have changed while the rest of memory did not.
The open question is whether that combination can be achieved without losing the
compactness and direct model integration that motivate latent and parametric
memory in the first place.




\subsection{Multimodal Memory Management}
\label{sec:open-management}

\Cref{sec:memory-management} describes how memory changes after formation, from
fixed retrieval to persistent maintenance and feedback-driven policy evolution.
The challenges collected here concern two questions that arise once memory is
allowed to change over time: how the system should keep its memory up to date as
new experience arrives, and how the policies that manage memory can themselves
improve from feedback.

\subsubsection{Maintaining Up-to-Date Memory}
\label{sec:open-up-to-date-memory}

Memory maintenance should move from recording change toward keeping track of
which state is still valid. New observations can be appended, existing memories
can be revised, and repeated evidence can be merged, but these operations do not
by themselves tell the system whether an older state still holds, has been
replaced, or should be kept only as history. This becomes especially important
in multimodal settings, where evidence about the same object, event, or user may
arrive at different times and through different modalities. The difficulty here
concerns observations of the same state that arrive separately, not disagreement
among the stored representations derived from them.

A timestamp records when an observation was made, not how long what it asserts remains true. In
multimodal settings the two can come apart in a particular way. A visual
observation may show where an object was seen, while a later instruction may
report that it has moved. A new image may correct an earlier description, while
the older observation remains useful as historical evidence. Where staleness is
handled at all, it is often handled by removing or replacing the outdated
record~\citep{worldmm2026}, or by checking a cached state against a fresh observation at the moment
of use~\citep{mobimem2025}, rather than by representing how long a belief remains valid. 

A promising direction is to explicitly model this status within the memory management process. A
memory may be marked as current, historical, superseded, uncertain, or still
valid alongside another state. Such decisions may depend on the source and
reliability of new evidence rather than on recency alone. They may also need to
distinguish when an observation is received from when the event it describes
actually happened. What remains open is how an agent should decide when new
evidence replaces an older state, when both should coexist, and when uncertainty
should be preserved rather than resolved.

\subsubsection{Self-Improving Memory Management}
\label{sec:open-self-improving-management}

Keeping memory up to date assumes that the rules for doing so are already good
enough. The next challenge is to improve those rules from experience. As
\Cref{sec:t3-feedback-evolution} describes, feedback-driven memory management
changes how memories are written, maintained, or retrieved during deployment.
This is different from optimizing a controller during training and then freezing
it: once deployment begins, a frozen controller continues to apply a stable policy even as
the memory contents change.

Current policy evolution is concentrated on retrieval, while changes to write
and maintenance policies remain much less common. 
One likely reason is that retrieval receives relatively direct feedback: the
system can often observe whether retrieved memory helped the next answer or
action. The effect of an earlier write, merge, revision, or forgetting decision
may appear much later, and in a multimodal agent it may not appear in the same
modality at all. A memory written from a visual observation may be revealed as
wrong only by a later instruction, by an observation in another modality, or by
the failure of an action taken on its basis. Credit therefore has to travel
across both time and modality before it can reach the decision that should
change.

A useful direction is to connect later success and failure back to the memory
decision that most likely caused it. This would allow feedback-driven
improvement to extend beyond retrieval without treating every downstream error
as evidence that the entire memory policy should change. What remains open is
how to assign delayed and cross-modal feedback to earlier memory decisions while
separating their effects from the many other operations that occur in between.

\subsection{Representation--Management Tensions in Multimodal Memory}
\label{sec:open-cross-axis}

\Cref{sec:cross-axis-analysis} shows that representation and memory management
cannot be designed independently. What a memory preserves limits what later
operations can inspect, change, verify, or remove. At the same time, the way
memory is managed determines which parts of a representation remain useful over
long interactions. The challenges here arise from this coupling. They are not
problems of representation or management alone, but cases in which one limits
what the other can do.

\subsubsection{Editable Representations without Loss of Detail}
\label{sec:open-fidelity-maintainability}

Multimodal memory should preserve perceptual detail without making later
correction difficult. Source observations and modality-specific representations
retain information that may disappear after an image, video, trajectory, or
other observation is converted into text or structure. Yet the same abstraction
that loses detail often makes memory easier to revise. A textual fact or graph
entry can be changed directly, while correcting one semantic detail inside a
long video, a distributed latent state, or a parameter update is much less
straightforward. What a representation preserves therefore limits how
precisely memory can later be maintained, both in how narrowly an update can be
scoped and in how directly its result can be checked.

Hybrid and source-linked designs reduce this tension by keeping an abstract
memory together with a route back to supporting evidence~\citep{vimrag2026, folio2026}. 
But coexistence alone does not settle the problem. An abstract record can become
outdated while the source observation remains valid as historical evidence. A
new observation may also contradict the current abstraction without making the
older evidence useless. The system must then decide which layer should guide
current behavior and when the source should be reopened rather than trusting
the abstraction.

A useful direction is to combine editable memory state with recoverable
modality-specific evidence and an explicit rule for moving between them. Routine
updates can operate on compact textual or structured state, while uncertain or
high-impact changes can return to the supporting image, video, audio segment, or
trajectory for verification. What remains open is how much source detail must
stay reachable for later correction without making long-term memory expensive
to maintain.

\subsubsection{Selective Evidence Retention for Verifiable Memory}
\label{sec:open-selective-verification}

Verifiable memory should not require every past observation to remain available
forever. Retained source evidence, provenance, version history, and
checkpoints make memory changes easier to inspect and reverse. In multimodal memory, however, the cost of keeping this
support depends strongly on what form the evidence takes. A short textual record,
selected image frames, a long video, an audio segment, a 3D trajectory, and a
latent or parametric checkpoint impose very different storage and processing
costs. The representation therefore limits how much evidence a management
system can realistically keep for later verification.

Current systems usually make this choice implicitly through what their memory
architecture happens to retain, rather than through an explicit policy for
verification. 
Some retain source observations~\citep{pixelrag2026}, some retain only derived records~\citep{chainofmemory2025}, and others
keep checkpoints or structured state~\citep{conceptgraphs2024}. What is often missing is a rule for which
evidence deserves long-term retention, which evidence can be compressed, and
which updates are important enough to justify preserving an earlier version.

A useful direction is selective verification. Lightweight provenance can remain
available for most memories, while expensive source evidence or checkpoints are
kept only when later inspection is likely to matter. High-impact revisions,
uncertain memories, or information supported by conflicting modalities may
justify stronger evidence retention than routine observations. The open problem
is how to predict this need before the future correction or dispute actually
occurs. A system that keeps too little loses the ability to verify its memory,
while one that keeps everything gives up much of the benefit of abstraction.

\subsubsection{Policy Adaptation with Preserved Evidence Grounding}
\label{sec:open-evidence-use}

Self-improving memory management should not silently change what evidence the
agent relies on. As \Cref{sec:t3-feedback-evolution} describes, a policy may
evolve how memory is written, maintained, or retrieved. In a multimodal system,
such a change can also alter the path through which information reaches the
agent. A retrieval policy may gradually favor textual summaries over visual
evidence, a write policy may compress one modality more aggressively, or a
maintenance policy may preserve some forms of memory while discarding others.
Management evolution can therefore change not only performance, but also which
representations continue to support that performance.

This creates a stability problem that task accuracy alone may not reveal. A
revised policy can maintain or even improve short-term answers while relying less
on the modality that originally grounded them. For example, a system may answer
correctly from a cached textual summary while rarely reopening the visual
evidence from which that summary was formed. If the summary later becomes wrong,
the loss of that evidence path may only become visible much later. Current
policy-evolution methods rarely evaluate whether adaptation changes the balance
of evidence used across representations. 

A useful direction is to evaluate policy changes together with the evidence
paths they alter. Before a new policy is accepted, the system should check not
only whether downstream performance improves, but also whether access to
supporting modalities and representations has been weakened. This may require
tracking which memory paths support important decisions and comparing that use
before and after adaptation. What remains open is how much change in evidence
use is healthy adaptation and when it becomes a loss of grounding that should
block or reverse the policy update.

\subsubsection{Sustaining Memory without Losing the Ability to Forget}
\label{sec:open-sustaining-forgetting}

Long-term memory should remain bounded without losing control over what can be
removed. The difficulty depends on representation. Retained images, videos,
audio, and trajectories grow with experience; structured memory grows with
entities, events, and relations; a bank of adapters can grow with tasks or
episodes; and a fixed-size latent state faces a different problem, because new
information must compete with what is already encoded. Representation therefore
determines what grows, while memory management determines how that growth can be
contained.

Existing systems use operations such as consolidation~\citep{emembench_embodied2026}, compression~\citep{li2026memento}, eviction~\citep{echovla2025},
decay~\citep{sun2026skemex}, and forgetting~\citep{memverse2025}, but sustained memory growth under a fixed budget remains
weakly addressed. 
A useful direction is representation-aware budgeting. The system should not
apply one retention rule to every memory form. Source evidence may be sampled or
compressed, structured records may be merged, latent capacity may need to be
reallocated, and parametric modules may need consolidation or removal. The hard
part is deciding what can be reduced without destroying information whose value
will only become clear later.

Capacity is only one reason to remove memory. Some information may need to
disappear because it is outdated, corrected, or explicitly deleted. In a
multimodal system, that information may already have spread from a source
observation into a caption, embedding, graph node, summary, latent state, or
parameter update. Removing one copy therefore does not establish that the
information is gone. This also creates a direct tension with
\Cref{sec:open-selective-verification}: verification benefits from retaining
evidence and version history, while deletion may require the same evidence and
history to disappear.

A useful direction is to specify forgetting as a cross-representation outcome rather than as the removal of a single record. Provenance and derivation links can identify the captions, embeddings, structured entries, summaries, or parameter modules that may carry the targeted information, but locating a derivative does not guarantee that the information can be selectively erased from it. The memory policy must determine which information should become unavailable, which unrelated knowledge should remain, and how the outcome can be verified. This also makes the tension with reversibility explicit: retaining an earlier version may support correction but conflict with a removal requirement. The open representational problem is especially difficult for latent states and shared parameters, where the targeted information is not separately addressable. \Cref{subsec:consistency-coordination} examines relevant coordination mechanisms, while \Cref{sec:open-memory-infrastructure} addresses how a declared removal reaches replicas and active derivatives and when that operation can be considered complete.

\subsection{Deployment-Oriented Memory Evaluation}
\label{sec:open-deployment-evaluation}

Evaluation determines which forms of progress become visible. Current
benchmarks establish increasingly broad coverage across modalities and tasks,
yet long-term memory is ultimately valuable as a process that preserves,
updates, accesses, and applies experience under continuing interaction. Future
evaluation should expose this process, stress it over sustained horizons and
bounded resources, and test it under the operating conditions faced by deployed
agents.

\subsubsection{Lifecycle-Aware Evaluation}

Multimodal memory evaluation should move from outcome-level scoring toward
stage-resolved diagnosis. Many existing benchmarks summarize performance with
final answer accuracy or task success, including long-video reasoning and
interactive visual-recall settings~\citep{long2025m3agent,dmvbench2026}. These
scores establish whether the complete system succeeded, while leaving the
origin of failure unresolved. Recent benchmarks demonstrate the value of richer
supervision through evidence annotations, update points, retrieval judgments,
feedback-reuse tests, and memory-specific execution
metrics~\citep{worldmemarena2026,streammembench2026,smmbench2026,memguibench2026}.
A lifecycle-aware protocol can organize such signals around four observable
stages.

\paragraph{Formation.}
Evaluation should determine whether the agent identifies the parts of an
experience that will matter later and forms faithful memory from them. Useful
process metrics include coverage of gold memory points, correctness and
supportedness of written records, rates of hallucinated or irrelevant entries,
preservation of source links, and retention of task-relevant visual, acoustic,
spatial, and temporal detail. These signals evaluate selective writing and
representation quality before later retrieval can conceal their failures.

\paragraph{Maintenance.}
Evaluation should track whether persistent memory remains valid as new evidence
arrives. Intermediate metrics can measure successful incorporation of current
state, rejection or supersession of stale records, conflict resolution,
faithfulness of consolidation, and consistency across source, textual,
structured, latent, or parametric representations. Versioned memory snapshots
and annotated state transitions can further reveal when an initially correct
memory becomes corrupted during update, compression, or reorganization.

\paragraph{Access.}
Evaluation should assess whether the evidence required by the current decision
is recovered from the appropriate memory path. Evidence recall, precision,
ranking quality, temporal or spatial localization, cross-modal coverage, and
retrieval latency expose distinct access failures. Comparing ordinary retrieval
with oracle evidence also separates limitations of memory access from those of
the downstream reasoner.

\paragraph{Utilization.}
Evaluation should verify that accessed memory actually guides the response,
plan, or action. Evidence-conditioned correctness, source attribution,
adherence to the current state under conflict, action--outcome consistency, and
recovery after feedback provide process-level measures of memory use.
Counterfactual removal or replacement of retrieved evidence can further test
whether an apparently correct output causally depends on memory.
For generative agents, utilization should additionally test whether retrieved constraints preserve entity, style, temporal state, and world consistency, and whether subsequent memory updates propagate coherently through later revisions of the generated artifact.

Few benchmarks expose all four stages, and complete lifecycle supervision is
especially scarce for native multimodal evidence. Establishing shared
trajectory annotations and stage-level metrics would make memory failures
localizable, allow process rewards to target the operation that failed, and
support meaningful comparison among architectures whose final accuracy may be
similar for very different reasons.

\subsubsection{Long-Horizon and Resource-Aware Evaluation}

Long-horizon evaluation should measure sustained memory quality under the
resource pressure created by accumulating multimodal experience. The field has
begun to extend temporal scale: M3-Bench evaluates videos averaging roughly
half an hour~\citep{long2025m3agent}, StreamArena reaches hour-scale causal
streams~\citep{streamarena2026}, EgoLife records a week of shared daily
experience~\citep{egolife2025}, and EgoMonth spans 20--120 days of egocentric
observations~\citep{egomonth2026}. This progression brings evaluation closer to
the timescale of persistent assistants. Continuous deployment protocols with
interleaved writing, maintenance, querying, and feedback remain much less
developed; even long multi-session histories are often frozen and queried
offline~\citep{memlens2026}.

\paragraph{Resource pressure.}
As the observation stream grows, a memory system consumes persistent storage,
write-time computation, retrieval bandwidth, model context, and response time.
Evaluation should therefore impose explicit budgets and report memory size and
growth rate, ingestion throughput and lag, retrieval latency, token or compute
cost per unit of experience, and the fraction of source evidence retained at
each fidelity level. Early studies already expose large differences in
write--read latency and execution cost across memory
designs~\citep{worldmemarena2026,memguibench2026}; standardized budgets would turn
these isolated measurements into comparable operating constraints.
Systems accounting should also distinguish the size of logical memory from the total physical footprint of its replicas, indices, cached execution states, and temporary buffers. In addition to end-to-end latency and memory quality, experiments should report peak accelerator occupancy, transferred bytes, and the costs of materialization, decompression, and recomputation. Warm-cache and cold-cache conditions should be identified separately, and construction or background write-back should not be treated as free when their costs are shifted outside the query interval. These measurements connect the quality--resource frontier to the mechanisms in \Cref{sec:memory-infrastructure} without reducing memory quality to cache hit rate.

\paragraph{Longitudinal degradation.}
A bounded memory must continually decide what to preserve, consolidate, or
forget. Its evaluation should trace accuracy, evidence recall, update
correctness, grounding, and action quality as a function of elapsed time,
accumulated experience, and available memory. Context-length analyses and
time-stratified forgetting curves provide useful starting
points~\citep{memlens2026,riverbench2026}, while deployment-oriented protocols should
also reveal delayed corruption, retrieval slowdown, representation drift, and
the compounding effect of early writing errors on later decisions. The central
comparison is the quality--resource frontier: how reliably a system preserves
useful, verifiable memory as its operating horizon grows under a fixed budget.

\subsubsection{Real-World Deployment}

Deployment evaluation should reproduce the conditions under which multimodal
memory must operate continuously. Real environments combine strict response
deadlines, imperfect sensing, asynchronous feedback, open-ended tasks, and
changes in users and worlds. Benchmarks increasingly incorporate real
egocentric recordings, interactive interfaces, and evolving
state~\citep{egolife2025,egomonth2026,memguibench2026,worldmemarena2026}, while most
still isolate these factors within curated data, controlled state transitions,
or resettable environments.

\paragraph{Real-time operation.}
An agent that recalls the correct evidence after the decision window has passed
cannot provide effective assistance. Evaluation should measure end-to-end
response latency, deadline-miss rate, proactive trigger timing, and the delay
between observation and query-visible memory availability. RIVER couples memory
accuracy with temporal distance and response timing, while StreamArena measures
query-to-answer latency across routing, recall, external search, and
inference~\citep{riverbench2026,streamarena2026}. Extending these protocols to continuous
memory writing and maintenance would expose whether background processing can
keep pace with the incoming multimodal stream.
For deployments requiring failure recovery, the time to a recoverable commit and behavior after interrupted writes should be measured separately under an explicit failure model; query visibility alone does not establish durability.

\paragraph{Perceptual and environmental robustness.}
Deployment streams contain motion blur, occlusion, poor lighting, overlapping
speech, sensor dropout, uncertain timestamps, and incomplete or contradictory
feedback. Systems also encounter devices, interfaces, locations, speakers, and
object appearances outside the distributions represented in curated
benchmarks. Evaluation should introduce controlled corruptions alongside
naturally noisy trajectories, vary which modalities remain available, and test
whether memory preserves uncertainty, requests clarification, or abstains when
the evidence is insufficient. Robustness should be measured across both memory
formation and later reuse, since an unnoticed perceptual error can persist long
after the original observation has disappeared.

\paragraph{Continual user and world change.}
Long-running agents must follow changing preferences, identities, routines,
goals, interface states, and physical environments. Evaluation should present
irregular state changes, delayed corrections, cross-modal disagreement, and
multiple users whose memories must remain separated. The resulting metrics
should capture adaptation delay, stale-memory use, recovery after correction,
cross-user contamination, and retention of valid historical evidence. Such
protocols would connect benchmark success to the defining requirement of
deployed memory: maintaining reliable continuity while the world that memory
describes continues to change.

\subsection{Memory Infrastructure and Efficiency}
\label{sec:open-memory-infrastructure}

\Cref{sec:memory-infrastructure} examines mechanisms for residency, restoration, reuse, materialization, and coordination. The open systems problem is how their assumptions and guarantees can compose over the lifetime of changing multimodal memory under bounded resources. The directions below concern runtime decisions and guarantees. They complement the representational and management requirements in \Cref{sec:open-cross-axis} and the evaluation protocols in \Cref{sec:open-deployment-evaluation}.

\subsubsection{Workflow-Responsive Residency and Data Movement}
\label{sec:memory-residency-movement}

Program- and workflow-aware systems already exploit execution phase, relative idleness, and anticipated reuse~\citep{kang2026thunderagent,mori2026,kvflow2025}. The next question is how placement and transfer execution should respond when those signals change while data is in flight. TokenCake, for example, already triggers an immediate upload when a tool returns earlier than predicted~\citep{tokencake2025}. A broader runtime contract should address branch cancellation, overlapping readers, and reprioritization across storage tiers: which transfers can be interrupted, which must drain, when staging buffers can be reused, and whether completed chunks remain useful. Research should connect transfer granularity and queue depth to bounded wasted traffic and resumption delay, rather than treating cancellation as an instantaneous reversal of offload.

Compression should also be optimized over repeated use rather than a single transfer. Building on the transport and compute--load mechanisms in \Cref{subsec:data-movement}, future designs should jointly account for encoding, decompression, temporary storage, layout conversion, and the number of expected restorations. Incremental restoration may reduce startup delay, but finer transfer units add metadata and scheduling overhead. Quality analysis should distinguish repeatedly decoding one unchanged compressed artifact from repeatedly re-encoding or approximately reconstructing its outputs; only the latter introduces an additional chain of potentially accumulating approximation. The objective is a lifetime cost--quality trade-off, not network-byte reduction alone.

\subsubsection{Dependency-Aware Reuse and Scheduling}
\label{sec:memory-index-reuse-schedule}

The dependency assumptions of prefix caching and selective recomputation in \Cref{subsec:indexing-scheduling} become harder to track when memory is revised, reordered, cropped, or materialized through a different model interface. Future work should investigate compact dependency descriptions that connect execution artifacts to their source versions and relevant transformations, such as preprocessing, encoder or adapter versions, prompt composition, positions, and attention configuration. These descriptions should distinguish exact reuse from empirically validated approximation and permit selective invalidation or reconstruction. An unresolved trade-off is granularity: coarse dependencies cause unnecessary recomputation, while fine-grained tracking can cost more than the computation it saves.

Scheduling should use this information without collapsing relevance into cache locality. Preble establishes the value of balancing prefix reuse with worker load~\citep{srivatsa2025preble}; multimodal memory introduces additional choices among available evidence representations and materialization paths. Research should optimize worker selection, cache admission, prefetching, and reconstruction jointly, subject to the evidence, freshness, and quality requirements of the task. Reuse value must include validation overhead, expected remaining uses before invalidation, and interference with other programs. A high hit rate is not a sufficient objective when the hits are expensive to restore, likely to be invalidated before further use, or available only for evidence that is less appropriate to the current decision.

\subsubsection{Coordinating Materialization and Write-back}
\label{sec:memory-read-write-orchestration}

Read-side acceleration can shift cost into memory construction rather than eliminate it. Existing workload characterization separates construction, retrieval, and generation costs, while FOLIO illustrates asynchronous construction with a recent-evidence path~\citep{agentmemorycharacterizationimplications,folio2026}. Future systems should jointly schedule foreground inference and background encoding, summarization, indexing, or adapter production. The key decisions are when to materialize expensive artifacts, when to defer work until demand is known, and how to prevent persistent write backlogs from making new experience unavailable. These decisions should consider expected reuse and invalidation, not just instantaneous GPU utilization.

Such orchestration also needs representation-specific fallbacks. When a compact record is not ready, a runtime may expose retained source evidence; when a model-specific artifact is incompatible, it may reconstruct a compatible form from an authorized source. These alternatives are not semantically interchangeable and should operate within the memory policy's requirements, rather than silently replacing evidence to meet a latency target. Research should determine when eager materialization, lazy reconstruction, or incremental publication provides the best balance of responsiveness, groundedness, and recoverability. Systems evaluation in \Cref{sec:open-deployment-evaluation} should account for the costs incurred on both sides of this read--write boundary.

\subsubsection{Cross-Layer Consistency, Recovery, and Revocation}
\label{sec:memory-orchestration}

The scoped guarantees in \Cref{subsec:consistency-coordination} motivate explicit publication and recovery contracts across artifacts. Future work should specify whether a read requires a consistent snapshot, read-your-writes visibility, or bounded staleness, and how these requirements span source media, semantic records, retrieval indices, and derived execution state. The coordination granularity need not be the entire memory bank; a versioned set of dependent artifacts may be sufficient. The challenge is to expose a coherent read path while expensive derivatives are being rebuilt, without globally blocking unrelated requests.

Durability should be specified separately from logical retention and query visibility. For deployments that require recovery from failures, research should identify which source records, versions, and transformation results must be committed, which caches can be reconstructed, and how interrupted write-back can resume without duplicating or exposing incomplete records. Preserving source media alone may not reproduce a prior stochastic summary or parameter update. The required guarantee must therefore distinguish recovery of the exact previous artifact from reconstruction of a newly generated alternative.

Deletion and permission changes extend coordination to already materialized state. Whereas \Cref{sec:open-sustaining-forgetting} asks what must be forgotten and how to verify that outcome, the infrastructure question is how a declared removal propagates through replicas, indices, summaries, contexts, caches, and retained versions. Future work should define a completion boundary and distinguish preventing new retrieval from retiring active derivatives and reclaiming storage. Collaborative Memory's retrieval-time access control provides one boundary, not a demonstrated solution for all downstream artifacts~\citep{collabmem2025}. Dependency metadata can identify retirement targets, but cannot by itself establish semantic erasure from a distributed latent state or shared model parameters.

MemOS proposes common abstractions for heterogeneous memory resources, while MemoryOS organizes hierarchical memory for agent interaction~\citep{memos2025,memoryos2025}. Building on such efforts, the objective should be interoperable contracts for identity, version, provenance, compatibility, and lifetime---not a universal representation that erases modality-specific semantics. The central open problem is to make their guarantees compose under finite resources and partial failure while preserving the evidence needed for future reasoning and correction.

\section{Conclusion}
\label{sec:conclusion}

The central question of multimodal agent memory is what evidence to retain from heterogeneous multimodal experience, how to represent it for subsequent reasoning and action, and how to keep it valid over time. Our framework characterizes multimodal agent memory along two axes: Formation and Representation, and Memory Management. Axis~I classifies the operative representation through which retained experience becomes available to downstream computation; Axis~II distinguishes T1 Static Memory, T2 Dynamic Memory and T3 Self-Evolving Memory. Joint analysis of the two axes shows that representation constrains the accessibility, revisability, verifiability, and recoverability of memory. The coexistence of multiple representations further requires maintaining provenance links, temporal validity, and cross-representation consistency, and updating memory state is not equivalent to evolving the management policy.
Current systems remain dominated by fixed memory and memory updating through stable policies. Online policy evolution, bounded lifelong memory, governed forgetting, and lifecycle-aware evaluation remain underdeveloped. At the systems level, current infrastructure lacks unified mechanisms for efficiently and consistently storing, restoring, and reusing memory under finite resource constraints; while existing benchmarks remain insufficient to establish whether memories remain valid, sufficient, and actionable throughout the full memory lifecycle. Together, these challenges point toward a multimodal memory system for long-term operation that preserves verifiable and recoverable multimodal evidence under bounded resources, maintains temporally evolving and cross-representation-consistent memory states, and uses continual feedback to improve its memory-management policies.

\bibliographystyle{plainnat}
\bibliography{references}
\end{document}